\documentclass[10pt,twocolumn,letterpaper]{article}

\usepackage{overpic}

\usepackage[pagenumbers]{cvpr} % To force page numbers, e.g. for an arXiv version

\usepackage{graphicx}
\usepackage{amsmath}
\usepackage{amssymb}
\usepackage{booktabs}
\usepackage{enumitem}
\usepackage{colortbl}
\usepackage{multirow}
\usepackage{subcaption}
\usepackage[pagebackref,breaklinks,colorlinks]{hyperref}

\usepackage{xcolor}

\newcommand\blfootnote[1]{%
  \begingroup
  \renewcommand\thefootnote{}\footnote{#1}%
  \addtocounter{footnote}{-1}%
  \endgroup
}

\newcommand{\CC}[1]{\cellcolor{#1}}
\definecolor{novelcolor}{rgb}{.9, .9, .9}

\begin{document}

\title{The {\color{teal}MONET} dataset: {\color{teal}M}ultimodal dr{\color{teal}one} {\color{teal}t}hermal dataset\\ recorded in rural scenarios}

\author{
Luigi Riz$^1$ \quad 
Andrea Caraffa$^1$ \quad 
Matteo Bortolon$^1$ \quad 
Mohamed Lamine Mekhalfi$^1$ \quad 
Davide Boscaini$^1$ \vspace{5px}\\
Andr\'e Moura$^2$ \quad
Jos\'e Antunes$^2$ \quad
Andr\'e Dias$^2$ \quad
Hugo Silva$^2$ \quad
Andreas Leonidou$^3$ \vspace{5px}\\
Christos Constantinides$^3$ \quad
Christos Keleshis$^3$ \quad
Dante Abate$^3$ \quad
Fabio Poiesi$^1$ \vspace{5px}\vspace{2mm}\\
\normalsize{
$^1$Fondazione Bruno Kessler \quad
$^2$INESC TEC \quad
$^3$The Cyprus Institute
}}

\maketitle

\begin{abstract}
We present MONET, a new multimodal dataset captured using a thermal camera mounted on a drone that flew over rural areas, and recorded human and vehicle activities. We captured MONET to study the problem of object localisation and behaviour understanding of targets undergoing large-scale variations and being recorded from different and moving viewpoints. Target activities occur in two different land sites, each with unique scene structures and cluttered backgrounds. MONET consists of approximately 53K images featuring 162K manually annotated bounding boxes. Each image is timestamp-aligned with drone metadata that includes information about attitudes, speed, altitude, and GPS coordinates. MONET is different from previous thermal drone datasets because it features multimodal data, including rural scenes captured with thermal cameras containing both person and vehicle targets, along with trajectory information and metadata. We assessed the difficulty of the dataset in terms of transfer learning between the two sites and evaluated nine object detection algorithms to identify the open challenges associated with this type of data. Project page: \url{https://github.com/fabiopoiesi/monet_dataset}.
\end{abstract}

%%%%%%%%%%%%%%%%%%%%%%%%%%%%%%%%%%%%%%%%%%%%%%%%%%%%%%%%%%%%%%%%%%%%%%%%%%%%%%%%
%%%%%%%%%%%%%%%%%%%%%%%%%%%%%%%%%%%%%%%%%%%%%%%%%%%%%%%%%%%%%%%%%%%%%%%%%%%%%%%%
%%%%%%%%%%%%%%%%%%%%%%%%%%%%%%%%%%%%%%%%%%%%%%%%%%%%%%%%%%%%%%%%%%%%%%%%%%%%%%%%
\section{Introduction}

Thermal image understanding enables localisation of objects that may not be visible through traditional RGB cameras.
\blfootnote{This work was carried out within the scope of the SHIELD project that received funding from the European Union’s Joint Programming Initiative – Cultural Heritage, Conservation, Protection and Use joint call.}
This can be useful in a variety of applications, such as in surveillance and security, where illicit activities typically occur overnight~\cite{Wu2014,Ma2016}, or in search and rescue \cite{Burke2019}, and military operations, where targets can be easier to locate based on their emitted heat rather than their cloth textures.

%-----------------------------------
\begin{figure}[t]
    \centering
    \includegraphics[width=\columnwidth]{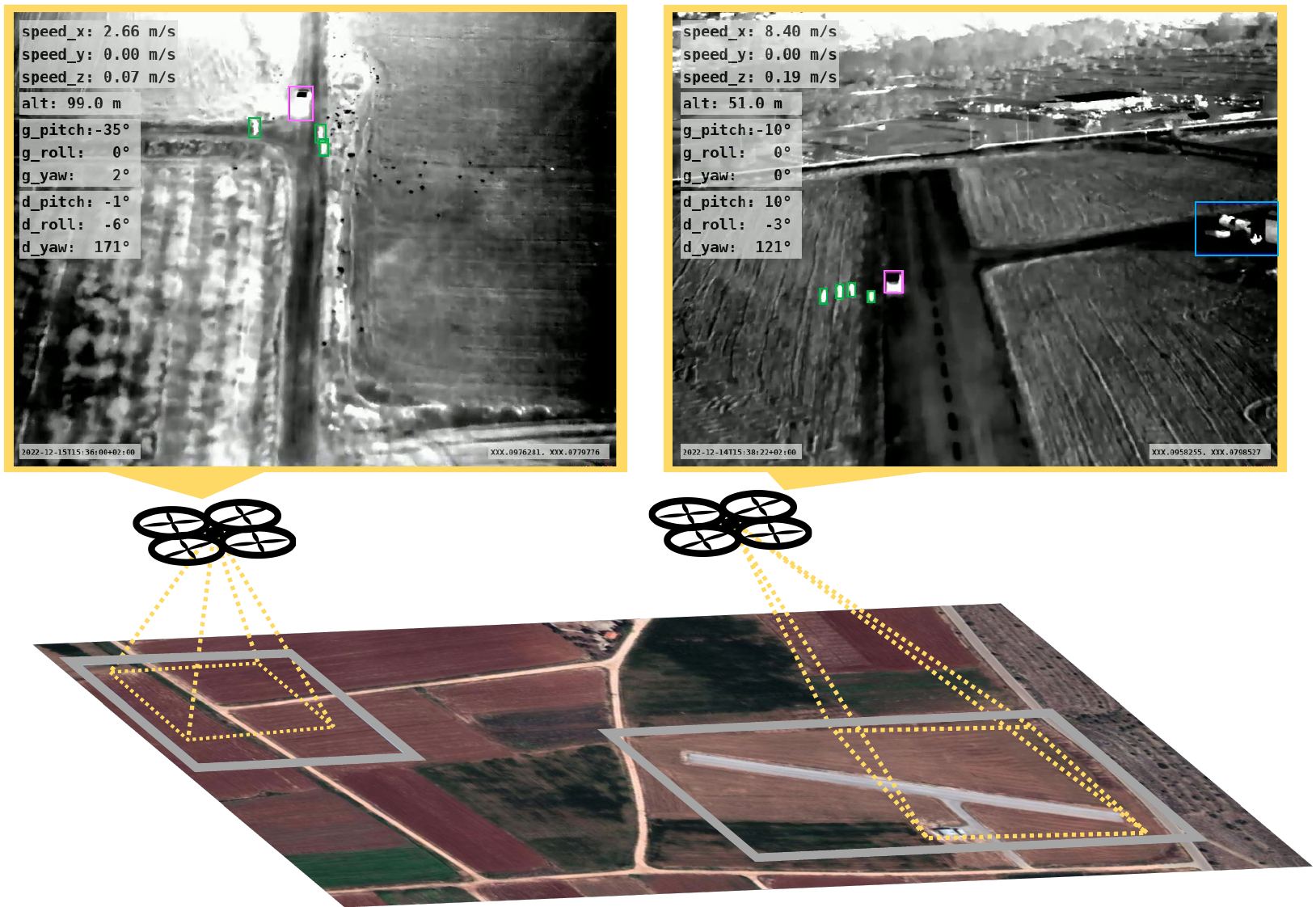}
    \vspace{-.5cm}
    \caption{The MONET dataset is captured with a thermal camera mounted on a multirotor drone.
    Three types of objects are annotated with bounding boxes: green is \texttt{person}, magenta is \texttt{vehicle}, and blue is \texttt{ignore}.
    Outlined in grey are the two recording sites: left-hand site is \textit{dirt-road} and the right-hand site is \textit{runway}.
    MONET provides drone metadata (e.g.~attitude, GPS, altitude) timestamp aligned to each image.
    The map below is a screenshot from Google Maps.
    Outlined in yellow are the two fields of view associated to the example thermal images.}
    \label{fig:location}
\end{figure}
%-----------------------------------

Object localisation in thermal images presents several challenges.
Typically, thermal imaging cameras have lower resolution than traditional RGB cameras, hence distinguishing fine-scale details in images is difficult.
Thermal imaging cameras are susceptible to noise and ghost effects, such as interference from other sources of heat or surface reflections.
The appearance of objects can vary depending on their temperature and emissivity, as well as the ambient temperature and humidity, hence algorithms should be robust to different environment conditions.
Thermal images can be cluttered, thus foreground objects may be indistinguishable from their background.
Similarly to RGB images, objects can also occlude each other, thus hindering multiple object localisation.
Objects of the same size and with the same temperature (e.g.~a boulder and a person), if captured from a distance, may appear with a similar silhouette.
In order to develop algorithms for object detection in thermal images, it is necessary to have a large and diverse dataset of annotated images.
However, such datasets may be difficult to obtain, because collecting large amounts of thermal data from drones and annotating them is often costly.

\begin{table*}[t]
\centering
\tabcolsep 5pt
\caption{Relevant publicly available real-world datasets recorded from drones outdoors.
Keys. 
V: Visible. 
NIR: Near Infrared (841$nm$). 
RE: Red Edge (717$nm$). 
LWIR: Long-Wave Infrared (7.5-13.5$\mu m$).
The dataset name is clickable and links to the dataset webpage.}
\vspace{-.2cm}
\label{tab:methods_comparison}
\resizebox{1\textwidth}{!}{
\begin{tabular}{llccccccccc}
\toprule
\multicolumn{2}{l}{\textbf{Dataset attribute}}
&
\textbf{\href{https://cvgl.stanford.edu/projects/uav_data/}{Campus}} \cite{Robicquet2016}
&
\textbf{\href{https://lafi.github.io/LPN/}{CARPK}} \cite{Hsieh2017}
&
\textbf{\href{http://aiskyeye.com/}{VisDrone}} \cite{Zhu2022}
& 
\textbf{\href{https://sites.google.com/view/grli-uavdt/}{UAVDT}} \cite{Du2018} 
&
\textbf{\href{https://sites.google.com/view/elizabethbondi/dataset}{BIRDSAI}} \cite{Bondi2020}
&
\textbf{\href{https://bozcani.github.io/auairdataset}{AU-AIR}} \cite{Bozcan2020}
&
\textbf{\href{https://seadronessee.cs.uni-tuebingen.de/home}{SeaDronesSee}} \cite{Varga2022}
&
\textbf{\href{https://github.com/suojiashun/HIT-UAV-Infrared-Thermal-Dataset}{HIT-UAV}} \cite{Suo2022}
&
\textbf{\href{https://github.com/fabiopoiesi/monet_dataset}{MONET}}
\\
\midrule
\# images & & 930K & 1.5K & 40K & 80K & 62K & 33K & 54K & 2.9K & 53K\\
\midrule
\multicolumn{2}{l}{\# bounding boxes} & 11M & 90K & 2.5M & 841K & 154K & 132K & 400K & 25K & 162K\\
\midrule
\multicolumn{2}{l}{\# object trajectories} & \checkmark & & \checkmark & \checkmark & \checkmark & & \checkmark & & \checkmark \\
\midrule
\multicolumn{2}{l}{Metadata} & & & & \checkmark & & \checkmark & \checkmark & \checkmark & \checkmark \\
\midrule
% \multicolumn{2}{l}{Type} & real & real & real & real & real+synth & real+synth & real &  real & real \\
% \midrule
%
\multirow{4}{*}{Categories} & People & \checkmark & & \checkmark & & \checkmark & \checkmark & \checkmark & \checkmark & \checkmark \\ 
\cmidrule{2-11}
& Vehicles & \checkmark & \checkmark & \checkmark & \checkmark & & \checkmark & \checkmark & \checkmark & \checkmark \\
\cmidrule{2-11}
& Animals & & & & & \checkmark & & & & \\
\midrule
\multicolumn{2}{l}{Sensor modality} & V & V & V & V & LWIR & V & V+NIR+RE & LWIR & LWIR \\
\midrule
\multirow{4}{*}{Time/Weather} & Day/Clear & \checkmark & \checkmark & \checkmark & \checkmark & & \checkmark & \checkmark & \checkmark & \checkmark \\
\cmidrule{2-11}
& Day/Foggy & & & & \checkmark & & & & & \\
\cmidrule{2-11}
& Night/Clear & & & \checkmark & \checkmark & \checkmark & & & \checkmark & \checkmark \\ 
\midrule
\multirow{4}{*}{Scenario} & Urban & \checkmark & \checkmark & \checkmark & \checkmark & & \checkmark & & \checkmark & \\
\cmidrule{2-11}
& Rural & & & \checkmark & & \checkmark & & & & \checkmark \\
\cmidrule{2-11}
& Maritime & & & & & & & \checkmark & & \\
\midrule
\multirow{2}{*}{Camera view} & Fixed & \checkmark & \checkmark & & & & \checkmark & & & \\
\cmidrule{2-11}
& Direction & top-down & top-down & varying & varying & varying & varying & varying & varying & varying \\ 
\midrule
\multirow{2}{*}{Drone attitude} & Static & \checkmark & & & & & & & & \\
\cmidrule{2-11}
& Moving & & \checkmark & \checkmark & \checkmark & \checkmark & \checkmark & \checkmark & \checkmark & \checkmark \\
\midrule
\multicolumn{2}{l}{Altitude [m]}
& 80 & 40 & n.a. & 10-70 & 60-120 & 5-30 & 5-260 & 60-130 & 20-130 \\
\midrule
\multicolumn{2}{l}{Year}
& 2016 & 2017 & 2018 & 2018 & 2020 & 2020 & 2022 & 2022 & 2023 \\
\bottomrule
\end{tabular}
 }
\end{table*}

Publicly, we can find several drone datasets recorded with visible spectrum cameras \cite{Robicquet2016,Hsieh2017,Du2018,Zhu2022,Bozcan2020}, instead thermal datasets are less popular.
There are some thermal datasets that are either annotated for single-object tracking applications \cite{Berg2015,Liu2019} (one target max per frame is annotated), or they are captured from static cameras resembling images captured from aerial vehicles \cite{Portmann2014,Wu2014}.
Our focus is understanding scenes that can potentially contain multiple objects and that are recorded from moving drones, but only few datasets are available with such desired properties \cite{Bondi2020,Suo2022}.
One is BIRDSAI \cite{Bondi2020}, a long-wave thermal infrared (LWIR) dataset that contains nighttime images of animals and humans in Southern Africa.
Another one is HIT-UAV \cite{Suo2022}, a LWIR dataset that contains both nighttime and daytime images of humans, bicycles, and vehicles, captured by a drone in urban scenarios (schools, parking lots, roads, playgrounds) flying between 60m to 130m altitude.
The SeaDroneSee dataset \cite{Varga2022} also provides thermal images captured from drones, but it includes scenes with humans and vehicles in water.
Although water is a challenging scenario, its challenges are different from terrain scenarios, i.e.~the structure of the environment is different, and background materials can make targets indistinguishable if they emit the same heat.
Moreover, drone metadata (e.g.~speed, altitude, drone and gimbal attitudes) is an important piece of information because one can use it to retrieve the expected scale of the targets to detect \cite{Messmer2022}, or to calibrate the motion models of tracking algorithms \cite{Li2017}.
To this end, SeaDroneSee provides a comprehensive list of metadata.
HIT-UAV only provides information about altitude, camera perspective, and a day/night flag.
BIRDSAI does not provide metadata.

In this paper, we present MONET, a multimodal drone thermal dataset recorded in rural scenarios that provides timestamp-aligned images and drone metadata (see Fig.~\ref{fig:location}). 
MONET comprises approximately 53K frames, with about 162K manually-annotated bounding boxes. 
The dataset includes two main target categories, i.e.~\texttt{person} and \texttt{vehicle}, plus a third category for a region to \texttt{ignore}.
Frames with targets contain 2.96 people and 1.33 vehicles on average.
Target bounding boxes are annotated with identities for multi-object tracking applications.
MONET's metadata includes drone and gimbal attitude (pitch, roll, and yaw), GPS, altitude, and speed (in the x, y, and z axes). 
The dataset presents challenges such as sudden and fast camera motion, background heat, large-scale variations, and different environmental structures.
We define two dataset scenarios, i.e.~\textit{runway} and \textit{dirt-road}, which include scenes recorded near a runway and in an agricultural land, respectively. 
Using the Faster R-CNN detector~\cite{Shaoqing2015}, we analyse MONET's challenges when training and evaluating on the same scenario (e.g.~runway to runway) and on different scenarios (runway to dirt-road). 
Although the sensor is the same, experiments show that training on one scenario and evaluating on the other leads to a significant drop in performance.
We evaluate nine object detectors and discuss and analyse MONET's challenges through qualitative results.

%%%%%%%%%%%%%%%%%%%%%%%%%%%%%%%%%%%%%%%%%%%%%%%%%%%%%%%%%%%%
%%%%%%%%%%%%%%%%%%%%%%%%%%%%%%%%%%%%%%%%%%%%%%%%%%%%%%%%%%%%
%%%%%%%%%%%%%%%%%%%%%%%%%%%%%%%%%%%%%%%%%%%%%%%%%%%%%%%%%%%%

\section{Related real-world datasets}\label{sec:related}

We conduct a survey of related real-world datasets by analysing various factors, including 
(i) object categories such as people, vehicles, and animals, 
(ii) different sensor modalities, such as visible and infrared, 
(iii) different scenarios, such as urban, rural, and maritime,
(iv) different camera views, such as fixed top-down and varying in different directions, and 
(v) different drone attitudes, such as static and moving. 
The datasets were recorded at various times of the day, with different weather conditions and at different altitudes, and all of them provide annotations in the form of bounding boxes (see Tab.~\ref{tab:methods_comparison}).

The Campus dataset \cite{Robicquet2016} comprises approximately 930K images with about 11M bounding box instances of pedestrians, bicyclists, and vehicles. 
These targets interact with each other within the Stanford University campus. 
The Campus dataset was designed to facilitate multi-object tracking, activity understanding, and trajectory forecasting. Object trajectories, along with their IDs, were annotated. The images were captured in the visible spectrum using a top-down camera during daytime from a multirotor drone hovering at an altitude of about 80m.

The Car Parking Lot Dataset (CARPK) \cite{Hsieh2017} comprises approximately 1.5K images with about 90K bounding box instances of cars from four different parking lots in urban scenarios. CARPK was designed for car counting, and no target IDs were annotated. 
The images were captured in the visible spectrum using a top-down camera during daytime from a multirotor drone flying at an altitude of 40m.

The VisDrone dataset \cite{Zhu2022} comprises approximately 40K images with about 2.5M bounding box instances of pedestrians, vehicles, and bicycles, captured from 14 cities in China, between urban and rural scenarios. The images were captured in the visible spectrum with arbitrary camera viewpoints during both daytime and nighttime from a multirotor drone flying at different altitudes. VisDrone provides object trajectory annotations, but no altitude information.

The UAVDT dataset \cite{Du2018} comprises 80K images with about 2.7K vehicles and 841K bounding box vehicle instances, such as cars, trucks, and buses, from different urban scenarios. The images were captured in the visible spectrum from arbitrary viewpoints during both daytime and nighttime using a multirotor drone at different altitudes. UAVDT provides object trajectory annotations and sequence-level metadata information, including i) time/weather conditions (daytime, nighttime, and fog), ii) flying altitude (low: 10-30m, medium: 30-70m, and high: above 70m), and iii) camera views (front, side, and bird).

The BIRDSAI dataset \cite{Bondi2020} comprises 62K images with about 120K animal and 34K human bounding box instances from different national parks in Southern Africa. 
BIRDSAI was designed for protected area monitoring to curb illegal activities like poaching and animal trafficking. BIRDSAI provides object trajectory annotations. The images were captured in the LWIR spectrum from arbitrary viewpoints during nighttime by using a fixed-wing drone flying at altitudes between 60m to 120m.

The AU-AIR dataset \cite{Bozcan2020} comprises 33K images with about 132K bounding box instances of people and vehicles in an urban scenario. This dataset was designed for object detection tasks, hence target IDs are unavailable. The images were captured in the visible spectrum with arbitrary camera viewpoints during daytime by using a multirotor drone flying at altitudes between 5m to 30m. Unlike UAVDT, AU-AIR provides image-level metadata, which includes drone speed, roll, pitch, yaw, altitude, latitude, and longitude. The camera is fixed on the drone and it points in different directions during the flight.

The SeaDroneSee dataset \cite{Varga2022} comprises 54K images with about 400K bounding box instances of people and vehicles (boats) in different maritime scenarios.
SeaDroneSee was designed for search and rescue applications, specifically for benchmarking multi-object tracking algorithms, hence object trajectories are provided.
Images were captured in both the visible and infrared spectrum by using fixed-wing and multirotor drones flying at altitudes between 5m to 260m.
Similarly to AU-AIR, SeaDroneSee provides metadata logged at 10Hz, which include drone speed, attitude, altitude, GPS, and gimbal pitch.

The HIT-UAV dataset \cite{Suo2022} comprises 2.9K images with about 25K bounding box instances of people and vehicles in urban scenarios.
This dataset was designed for object detection tasks, hence target IDs are unavailable. 
Images were captured in the LWIR spectrum with arbitrary camera viewpoints during both daytime and nighttime by using a multirotor drone flying at altitudes between 60m to 130m.

Unlike BIRDSAI, MONET includes metadata and the vehicle category instead of the animal category, and is recorded from a multirotor drone as opposed to a fixed-wing drone.
Unlike HIT-UAV, MONET includes more metadata, and has several more annotations that can also be used for multiple object-tracking applications.

%%%%%%%%%%%%%%%%%%%%%%%%%%%%%%%%%%%%%%%%%%%%%%%%%%%%%%%%%%%%
%%%%%%%%%%%%%%%%%%%%%%%%%%%%%%%%%%%%%%%%%%%%%%%%%%%%%%%%%%%%
%%%%%%%%%%%%%%%%%%%%%%%%%%%%%%%%%%%%%%%%%%%%%%%%%%%%%%%%%%%%
\section{Hardware}

%%%%%%%%%%%%%%%%%%%%%%%%%%%%%%%%%%%%%%%%%%%%%%%%%%%%%%%%%%%%
%%%%%%%%%%%%%%%%%%%%%%%%%%%%%%%%%%%%%%%%%%%%%%%%%%%%%%%%%%%%
\subsection{Multirotor drone}

We used a fully-customised multirotor drone, which was designed for automated surveillance and detection of archaeological looting activities.
The drone includes the airframe, payload, propulsion, control, and communication systems.
The airframe is designed to be compact, lightweight, and capable of carrying payloads up to 1.5kg.
The propulsion system is composed of eight motors and eight Electronic Speed Controllers, ensuring stability and redundancy.
It has four arms that support these eight motors. 
The control system includes a high-frequency IMU, an accurate barometric altimeter, and an external GPS and compass module. 
It supports several flight modes, including manual, stabilised, heading hold, hovering, automated waypoint navigation, return to home, auto take-off and auto-landing, and provides real-time monitoring through an on-screen display. 
The communication system provides real-time control and monitoring of the drone, and its payload.

%%%%%%%%%%%%%%%%%%%%%%%%%%%%%%%%%%%%%%%%%%%%%%%%%%%%%%%%%%%%
%%%%%%%%%%%%%%%%%%%%%%%%%%%%%%%%%%%%%%%%%%%%%%%%%%%%%%%%%%%%
\subsection{Data acquisition system}

\begin{figure}[t]
    \centering
    \includegraphics[width=\columnwidth]{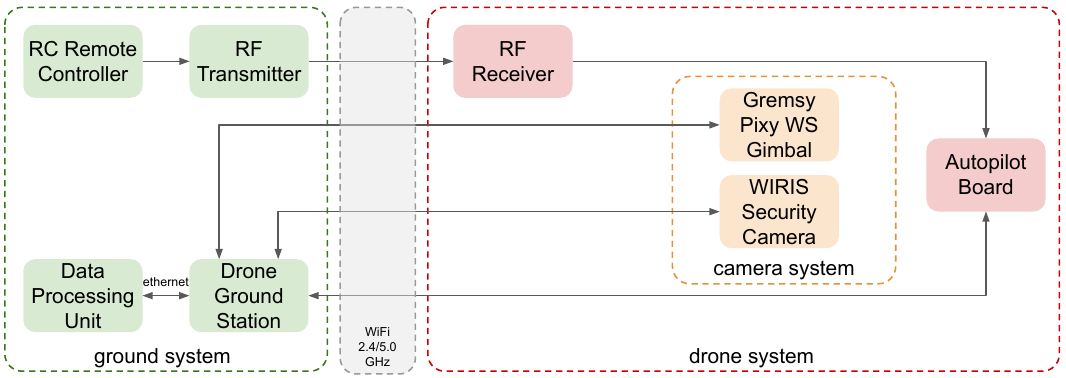}
    \vspace{-.6cm}
    \caption{Data acquisition architecture that includes ground and drone system, and WiFi communication between these two.}
    \label{fig:acq_arch}
\end{figure}

The camera acquisition system consists of 
i) the camera WIRIS Security from Workswell\footnote{\url{https://workswell-thermal-camera.com/drone-security-thermal-imaging-camera-night-vision-uav}: last access: Apr.~2023.} that features two separate sensors, i.e.~RGB and thermal, 
ii) the gimbal unit to physical hold the camera on the drone, and 
iii) the Data Processing Unit.
Fig.~\ref{fig:acq_arch} illustrates the architecture.

%-----------------------------------
%-----------------------------------

The WIRIS Security is provided with a proprietary SDK. 
In order to control the camera, we developed a customised module that implements the WIRIS Security control commands (through SDK) to the camera via Ethernet connection, which is implemented in the drone ground station using ROS~\cite{ros}. 
The RGB and thermal images are received in the Data Processing Unit as RTSP video streams, and published in ROS topics by using two independent nodes.
The RGB sensor allows up to 30x optical zoom, Full HD (1920$\times$1080) resolution, at a framerate of 30Hz. 
The thermal sensor operates in the LWIR spectrum (7.5-13.5 $\mu$m) featuring an 800$\times$600 resolution with -20$^\circ$C to 150$^\circ$C thermal sensitivity.
The camera includes an internal SSD drive with 256GB of storage space, allows communication through Ethernet, USB and HDMI.

%%%%%%%%%%%%%%%%%%%%%%%%%%%%%%%%%%%%%%%%%%%%%%%%%%%%%%%%%%%%
%%%%%%%%%%%%%%%%%%%%%%%%%%%%%%%%%%%%%%%%%%%%%%%%%%%%%%%%%%%%
%%%%%%%%%%%%%%%%%%%%%%%%%%%%%%%%%%%%%%%%%%%%%%%%%%%%%%%%%%%%
\section{Dataset}

We collected MONET in a rural area near the city of Nicosia, Cyprus, in mid-December.
The sequences were captured in the afternoon, evening and night.
One recording site is on a runway, which is property of The Cyprus Institute, the other one is on agricultural lands.
We name these sites as \textit{runway} and \textit{dirt-road}, respectively.
Fig.~\ref{fig:location} shows the two sites.

%%%%%%%%%%%%%%%%%%%%%%%%%%%%%%%%%%%%%%%%%%%%%%%%%%%%%%%%%%%%
%%%%%%%%%%%%%%%%%%%%%%%%%%%%%%%%%%%%%%%%%%%%%%%%%%%%%%%%%%%%
\subsection{Annotation procedure}
Six people contributed to the annotation of MONET by using CVAT~\cite{cvat}.
CVAT was installed on a server and utilised via web browser.
We created an account for each annotator.
A certain number of distinct sequences were assigned to each annotator.
We asked annotators to follow a set of guidelines:
i) bounding boxes should be drawn as tight as possible on the targets as long as the object is clearly distinguishable from its background;
ii) bounding box interpolation across frames is allowed as long as each frame is checked to see if the bounding boxes are correctly centred on targets;
iii) brightness, contrast and saturation can be adjusted through CVAT UI to make targets more distinguishable;
iv) if a target is partially occluded by another object or indistinguishable from the background, its bounding box should be drawn based on the best guess of the annotator and then flagged as \textit{occluded};
v) the annotation of a target should start when more than $\sim$30\% of its pixels are in the scene;
vi) the annotation of a target should terminate when more than $\sim$70\% of its pixels are outside the scene;
vii) if a target exits the scene and then re-enters a new ID should be associated to it.
Once the annotations were completed, three people double-checked them to ensure they were accurate and consistent with the guidelines.
The Supplementary Material contains annotation examples.

%%%%%%%%%%%%%%%%%%%%%%%%%%%%%%%%%%%%%%%%%%%%%%%%%%%%%%%%%%%%
%%%%%%%%%%%%%%%%%%%%%%%%%%%%%%%%%%%%%%%%%%%%%%%%%%%%%%%%%%%%
\subsection{Bounding box categories and statistics}

We annotated three types of targets: \texttt{vehicle}, \texttt{person}, and \texttt{ignore}.
The bounding boxes of \texttt{vehicle} include car-like objects, \texttt{person} is self-explanatory, and \texttt{ignore} include the hangar location next to the runway.
We decided to ignore this location because other people and vehicles are often visible next to the hangar, and we want to avoid data-driven detection algorithms to learn pattern biases (e.g.~people and vehicles often next to the hangar structure).
Therefore, we zeroed the regions of the images defined by the \texttt{ignore} bounding boxes in order to avoid this bias during training.
We provide the image data in its original form if one wants to exploit this location or additional targets for different purposes.

% ********************************
\begin{figure}[t]
\begin{subfigure}{1\columnwidth}
    \begin{tabular}{@{}c@{}c}
        \begin{overpic}[width=.585\linewidth]{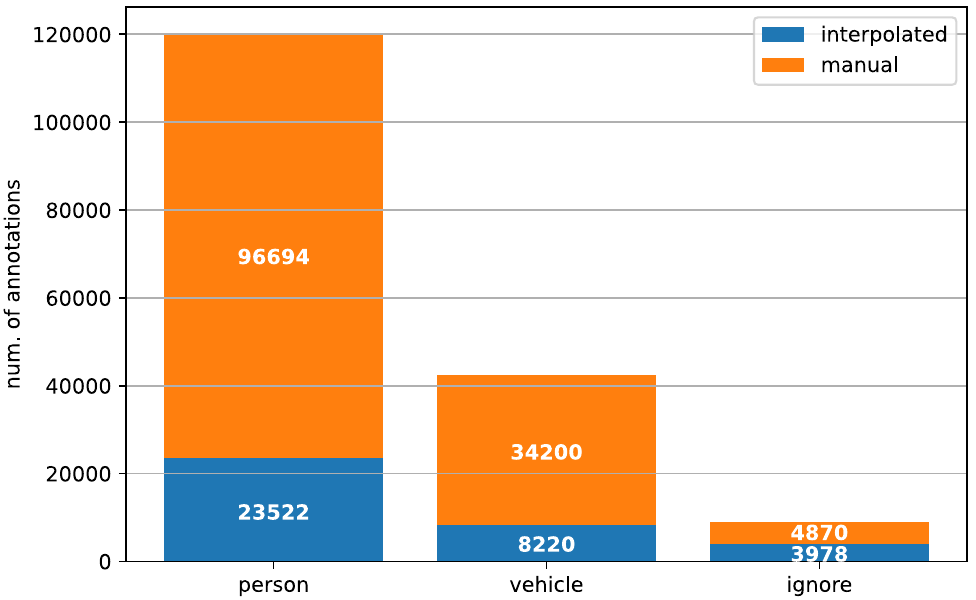}
        \end{overpic} &
        \begin{overpic}[width=.415\linewidth]{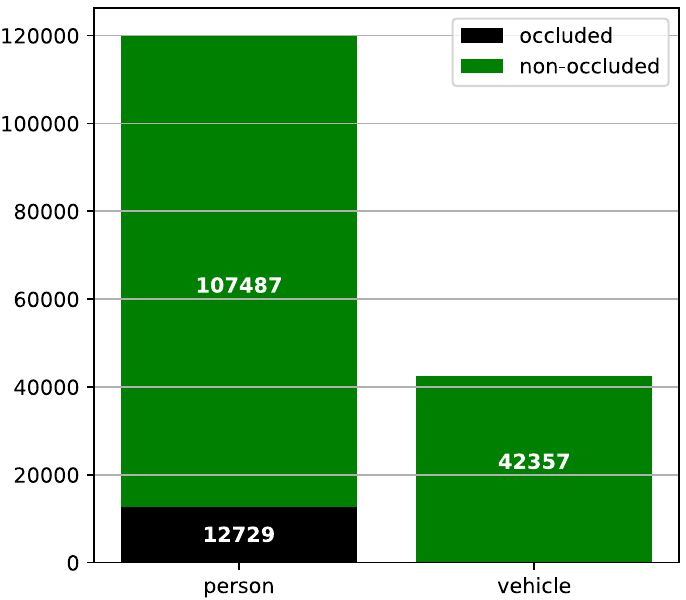}
        \end{overpic}
    \end{tabular}
\end{subfigure}
\begin{subfigure}[b]{.5\columnwidth}
\resizebox{1.1\columnwidth}{!}{
    \hspace{1cm}\begin{tabular}[b]{llll}
        \toprule
        cat. & avg. & max & min \\
        \midrule
        \texttt{person} & 2.96 & 6.00 & 1.00 \\
        \texttt{vehicle} & 1.33 & 4.00 & 1.00 \\
        \texttt{ignore} & 1.00 & 1.00 & 1.00 \\
        \bottomrule
    \end{tabular}
    }
\end{subfigure}
\hspace{5mm}
\begin{subfigure}[b]{.4\columnwidth}
    \includegraphics[width=.85\textwidth]{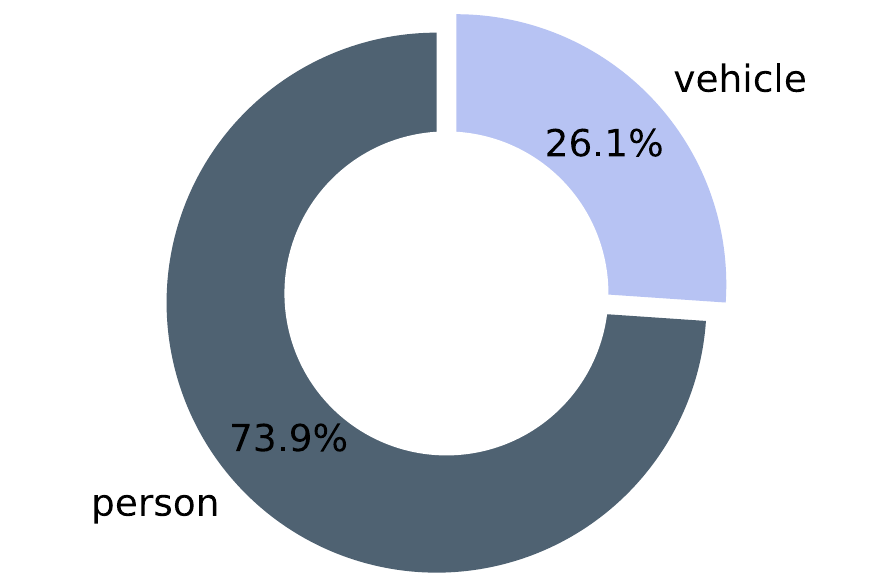}
\end{subfigure}
%
% \vspace{-.6cm}
\caption{Bounding box annotation statistics of the MONET dataset. 
The top-left graph highlights the number of interpolated and manually drawn bounding boxes.
The top-right graph highlights the number of occluded and non-occluded bounding boxes:
\texttt{vehicle} features 63 occluded bounding boxes (\texttt{ignore} is omitted because always non-occluded).
The table shows the average, maximum and minimum of number of targets that can be found.
The doughnut chart summarises the percentage of \texttt{person} and \texttt{vehicle} bounding boxes.
}
\label{fig:stats}
\end{figure}
% ********************************

Fig.~\ref{fig:stats} illustrates the bounding box annotation distributions of the whole dataset.
The left-hand side graph shows that the largest portion of bounding boxes is manually drawn, while the others are (linearly) interpolated.
Interpolation is a feature of CVAT and is applied between consecutive frames.
While interpolation can be effective when annotations are made on videos captured from static cameras, this is not the case when the camera is moving.
We only exploited interpolation occasionally because the motion of the drone plus the motion of the targets cannot be modelled with a linear motion model.
The right-hand side graph shows that the largest portion of bounding boxes is visible, while the others are occluded.
The annotators flag a bounding box as occluded when they deem the target was significantly occluded by another target, or when the target was indistinguishable from the background, e.g.~due to interfering emitted temperatures.
Although the target is indistinguishable in some frames, we purposely annotated the bounding boxes (flagging them as occluded) as it is a potential challenge if one's use case is tracking.
We investigate the effect of training detectors with and without occluded bounding box in Sec.~\ref{sec:experiments}.

%%%%%%%%%%%%%%%%%%%%%%%%%%%%%%%%%%%%%%%%%%%%%%%%%%%%%%%%%%%%
%%%%%%%%%%%%%%%%%%%%%%%%%%%%%%%%%%%%%%%%%%%%%%%%%%%%%%%%%%%%
\subsection{Drone metadata}

Together with the camera frames, MONET also includes drone metadata information.
As for the images, we logged the timestamp of each metadata during dataset acquisition.
Metadata was captured at about 40Hz on average.
Like \cite{Varga2022}, we used nearest-neighbour assignment between image and metadata timestamps.
Metadata includes the date in ISO 8601 format, the drone and gimbal attitudes (pitch, roll, yaw), latitude, longitude, altitude and speed (x, y, and z axes).
Tab.~\ref{tab:metadata} includes a detailed list of the metadata we collected along with their minimum and maximum values.

\begin{table}[t]
\centering
\tabcolsep 5pt
\caption{MONET metadata specifications indicating the units of each metadata along with its minimum and maximum value.}
\vspace{-.2cm}
\label{tab:metadata}
\resizebox{.9\columnwidth}{!}{
\begin{tabular}{llcc}
\toprule
Data & Unit & min.~value & max.~value \\
\midrule
date & ISO 8601 & - & - \\
drone pitch & degrees & -90 & 90 \\
drone roll & degrees & -90 & 90 \\
drone yaw & degrees & 0 & 360 \\
gimbal pitch & degrees & -40 & 90 \\
gimbal roll & degrees & -45 & 45 \\
gimbal yaw & degrees & -180 & 180 \\
latitude & degrees & -90 & 90 \\
longitude & degrees & -180 & 180 \\
altitude & m & 0 & user defined \\
x-axis speed & cm/s & 0 & 2800 \\
y-axis speed & cm/s & 0 & 2800 \\
z-axis speed & m/s & 0 & 10 \\
\bottomrule
\end{tabular}
 }
\end{table}

%%%%%%%%%%%%%%%%%%%%%%%%%%%%%%%%%%%%%%%%%%%%%%%%%%%%%%%%%%%%
%%%%%%%%%%%%%%%%%%%%%%%%%%%%%%%%%%%%%%%%%%%%%%%%%%%%%%%%%%%%
\subsection{Examples of images and annotations}

\begin{figure}[t]
\begin{center}
  \begin{tabular}{@{}c@{\,}c}
    \begin{overpic}[width=.5\linewidth,height=.4\linewidth]{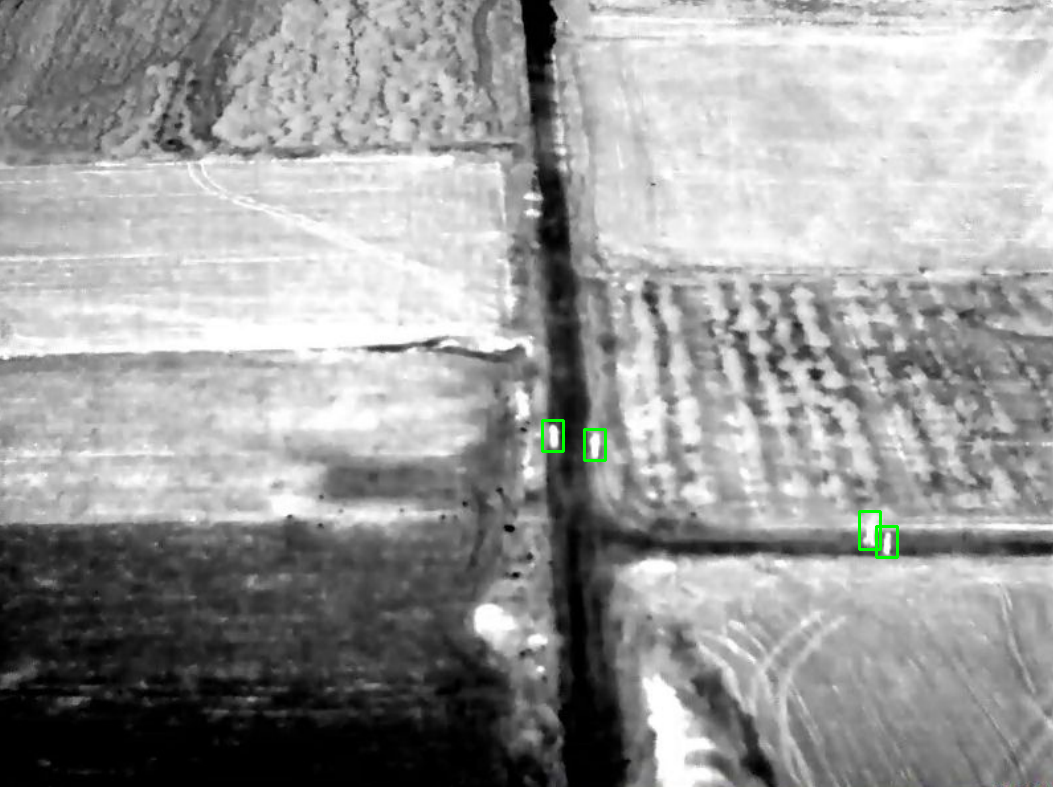}
    \put(2,3){\color{lime}\footnotesize \textbf{(a)}}
    \end{overpic} &
    \begin{overpic}[width=.5\linewidth,height=.4\linewidth]{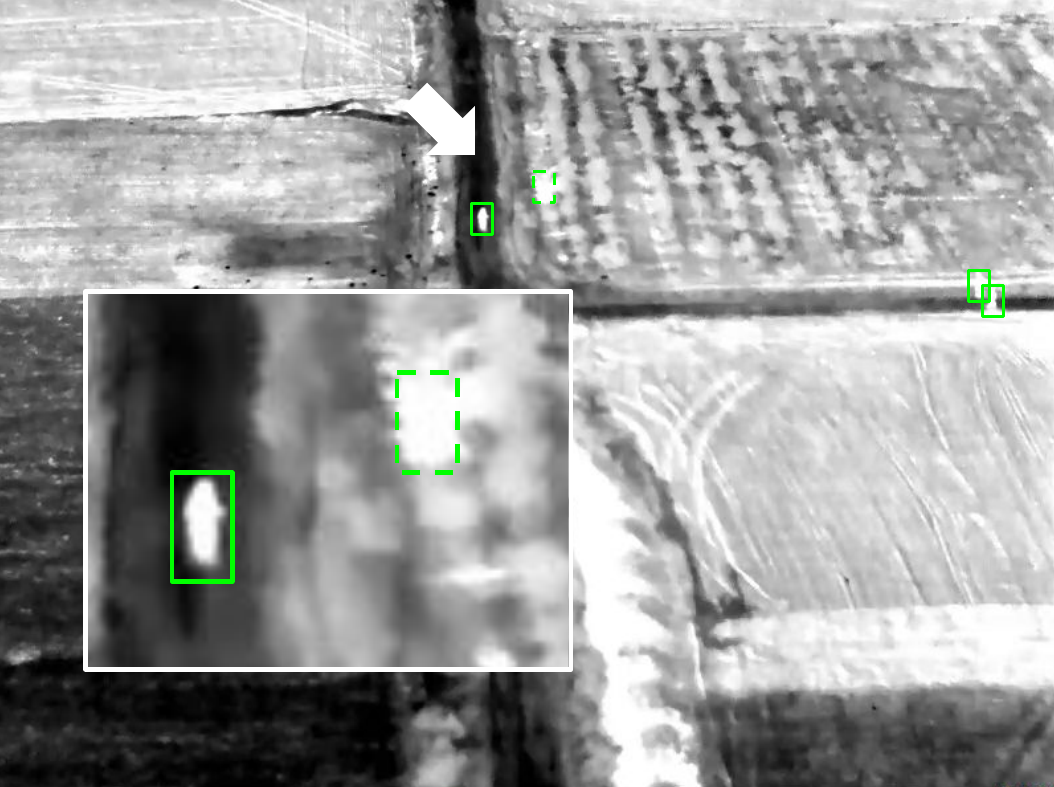}
    \put(2,3){\color{lime}\footnotesize \textbf{(b)}}
    \end{overpic}\\
    \begin{overpic}[width=.5\linewidth,height=.4\linewidth]{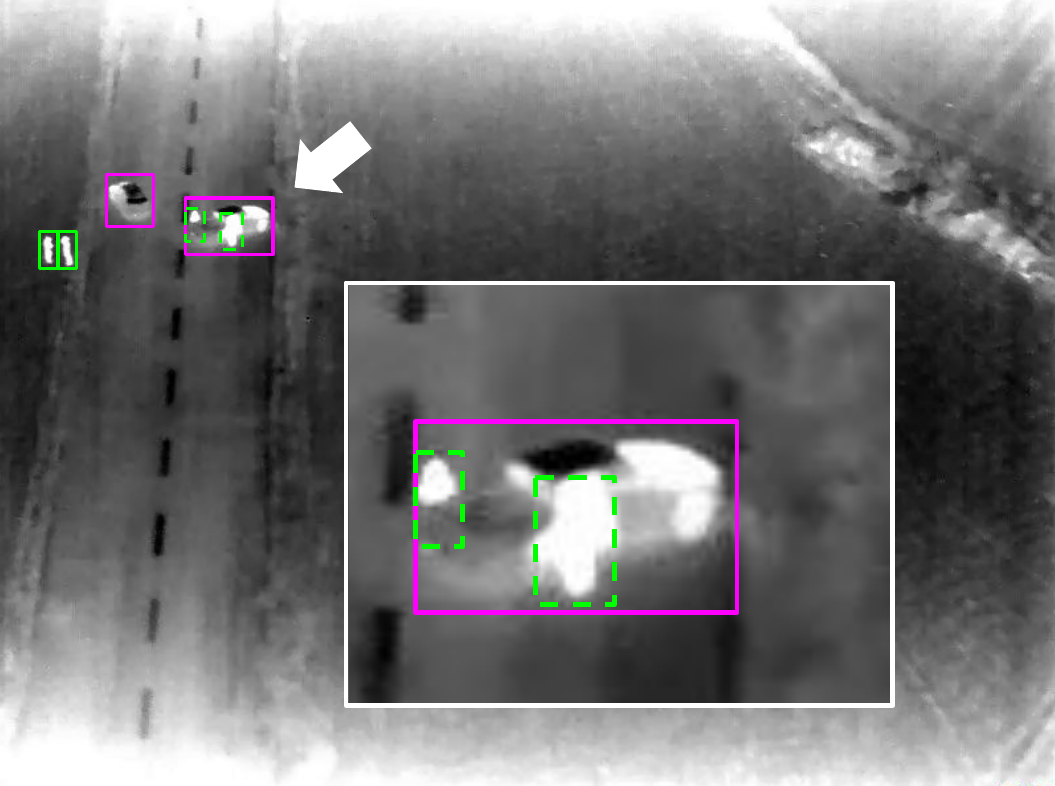}
    \put(2,3){\color{lime}\footnotesize \textbf{(c)}}
    \end{overpic} &
    \begin{overpic}[width=.5\linewidth,height=.4\linewidth]{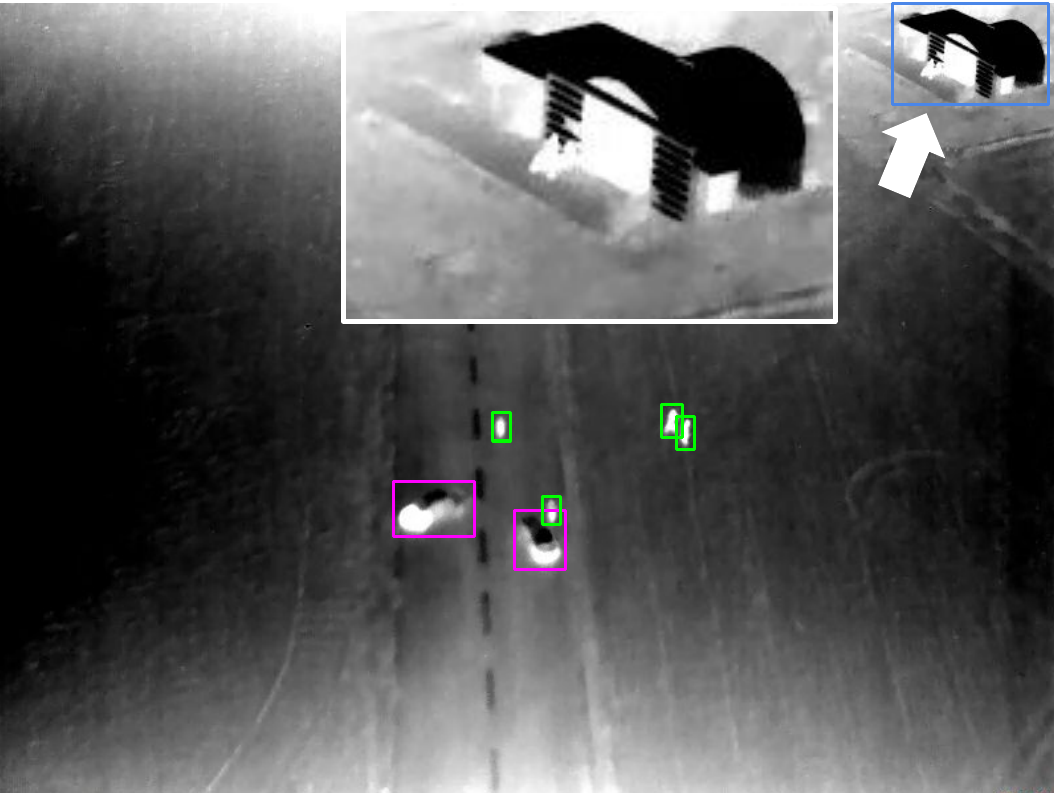}
    \put(2,3){\color{lime}\footnotesize \textbf{(d)}}
    \end{overpic}
  \end{tabular}
\end{center}
\vspace{-.6cm}
\caption{Annotation examples showing (a,b) dirt-road and (c,d) runway scenarios. 
Bounding boxes: green for \texttt{person}, magenta for \texttt{vehicle}, and blue for \texttt{ignore}.
Recording altitudes (a-c) 80m and (d) 100m.
Dotted bounding boxes indicate occlusions:
(b) Example of occlusion where the target is indistinguishable from the background.
(c) Example of occlusion where the heat from a vehicle makes the person indistinguishable.
(d) Example of bounding box labelled as \texttt{ignore}, enclosing the hangar area.}
\label{fig:qualitative_examples}
\end{figure}

Fig.~\ref{fig:qualitative_examples} shows examples of annotations in dirt-road and runway scenarios recorded from the altitudes of 80m (a-c) and 100m (d).
In particular, Fig.~\ref{fig:qualitative_examples}a shows four \texttt{person} targets in the dirt-road scenario in normal visibility conditions.
Fig.~\ref{fig:qualitative_examples}b shows the same four targets as before, but when one target is flagged as occluded: the small difference in the measured heat between target and background makes them nearly indistinguishable.
Fig.~\ref{fig:qualitative_examples}c shows a similar case where a target is difficult to distinguish from the background, but in this case this is due the vehicle's heat behind the person.
Lastly, we show an example of an \texttt{ignore} region.
This is the hangar hosting the drone operators.
We train our detection algorithms by zeroing the region of the image defined by the \texttt{ignore} bounding box.
See the project page for videos of dirt-road and runway showing the annotations along with the aligned metadata.

% ********************************
% ********************************

%%%%%%%%%%%%%%%%%%%%%%%%%%%%%%%%%%%%%%%%%%%%%%%%%%%%%%%%%%%%%%%%%%%%%%%%%%%%%%%%
%%%%%%%%%%%%%%%%%%%%%%%%%%%%%%%%%%%%%%%%%%%%%%%%%%%%%%%%%%%%%%%%%%%%%%%%%%%%%%%%
%%%%%%%%%%%%%%%%%%%%%%%%%%%%%%%%%%%%%%%%%%%%%%%%%%%%%%%%%%%%%%%%%%%%%%%%%%%%%%%%
\section{Experiments}\label{sec:experiments}

%%%%%%%%%%%%%%%%%%%%%%%%%%%%%%%%%%%%%%%%%%%%%%%%%%%%%%%%%%%%%%%%%%%%%%%%%%%%%%%%
%%%%%%%%%%%%%%%%%%%%%%%%%%%%%%%%%%%%%%%%%%%%%%%%%%%%%%%%%%%%%%%%%%%%%%%%%%%%%%%%
\subsection{Experimental setup}

%%%%%%%%%%%%%%%%%%%%%%%%%%%%%%%%%%%%%%%%%%%%%%%%%%%%%%%%%%%%%%%%%%%%%%%%%%%%%%%%
\noindent \textbf{Scenes and settings.}
We split the dataset into two scenes: \textit{dirt-road} and \textit{runway}, which include recordings of people activities nearby a runway and in an agriculture land, respectively.
Dirt-road is composed of 23.3K frames with 83.4K bounding box annotations, while runway is composed of 29.4K frames with 79.3K bounding box annotations.
Each scene is divided into disjoint splits for training, validation, and test.
In Fig.~\ref{fig:location} we can see that the structure of the environment is different between these two scenes.
So we conduct two sets of experiments to analyse the challenges of MONET in terms of transfer learning when object detection algorithms are trained and evaluated on different scenes.
Firstly, we focus on Faster R-CNN \cite{Shaoqing2015} and assess several transfer learning configurations, including the combination of MONET and HIT-UAV \cite{Suo2022} data.
Secondly, we compare the performance of nine popular object detectors, aiming to understand MONET's challenges.
We use COCO evaluation procedure and report AP, AP$_{50}$, AP$_{75}$, and AP per class \cite{Lin2014}.
Because each detector is calibrated differently, setting a comparable detection confidence threshold is impractical.
Therefore, we evaluate all the detectors by using confidence $10^{-8}$.
This aspect is further discussed in the Supplementary Material.

%%%%%%%%%%%%%%%%%%%%%%%%%%%%%%%%%%%%%%%%%%%%%%%%%%%%%%%%%%%%%%%%%%%%%%%%%%%%%%%%
%%%%%%%%%%%%%%%%%%%%%%%%%%%%%%%%%%%%%%%%%%%%%%%%%%%%%%%%%%%%%%%%%%%%%%%%%%%%%%%%
\noindent \textbf{Detectors.}
We use the MMDetection open source object detection toolbox \cite{mmdetection} for the following implementations:
Faster R-CNN (2015) \cite{Shaoqing2015},
SSD (2016) \cite{Liu2016},
CornerNet (2018) \cite{Law2018},
FCOS (2019) \cite{Tian2019},
DETR (2020) \cite{Carion2020},
Deformable DETR (2021) \cite{Zhu2021}, and
VarifocalNet (2021) \cite{Zhang2020}, while we use authors' implementation for
ObjectBox (2022) \cite{Zand2022} and
YOLOv8 (2023) \cite{Yolov8}.
We train all these methods with the same data augmentations, and, where possible, with the same backbones and optimisation parameters.
Only for ObjectBox and YOLOv8 we perform additional experiments with the data augmentations proposed by the authors to investigate if they lead to different results.
Please refer to the Supplementary Material for the training configuration of each detector.

%%%%%%%%%%%%%%%%%%%%%%%%%%%%%%%%%%%%%%%%%%%%%%%%%%%%%%%%%%%%%%%%%%%%%%%%%%%%%%%%
%%%%%%%%%%%%%%%%%%%%%%%%%%%%%%%%%%%%%%%%%%%%%%%%%%%%%%%%%%%%%%%%%%%%%%%%%%%%%%%%
\subsection{Quantitative results}

\begin{table}[t]
\centering
\tabcolsep 3pt
\caption{Evaluation of Faster R-CNN initialised with COCO trained weights. 
Values are provided in percentage.
Grey background indicates transfer learning results. Keys:
V: validation split.
T: test split.
\textdagger: additional pre-training on top of COCO initialisation. 
w/o occ.: without bounding boxes with occlusion flag.
}
\vspace{-.2cm}
\label{tab:results_faster_rcnn}
\resizebox{1\columnwidth}{!}{
\begin{tabular}{lllccccc}
\toprule
\multirow{2}{*}{Exp} & \multirow{2}{*}{Train split} & \multirow{2}{*}{Eval Split} & \multirow{2}{*}{AP} & \multirow{2}{*}{AP$_{50}$} & \multirow{2}{*}{AP$_{75}$} & \multicolumn{2}{c}{class-AP} \\
 & & & & & & person & vehicle \\
 \midrule
\multirow{4}{*}{1} & \multirow{4}{*}{\shortstack[l]{dirt-road\\+runway}}            & dirt-road V & 24.9 & 63.8 & 6.5 & 10.9 & 39.0 \\
 & & dirt-road T & 36.3 & 82.4 & 22.0 & 33.1 & 39.6 \\
\cmidrule{3-8}
 & & runway V & 44.5 & 90.2 & 38.9 & 31.8 & 57.2 \\
 & & runway T & 42.1 & 84.0 & 37.3 & 37.8 & 46.5 \\
 \midrule
\multirow{4}{*}{2} & \multirow{4}{*}{\shortstack[l]{dirt-road\\+runway\\(w/o occ.)}} & dirt-road V & 25.1 & 67.8 & 7.8 & 13.6 & 36.6 \\
 & & dirt-road T & 36.3 & 81.3 & 21.9 & 32.1 & 40.5 \\
\cmidrule{3-8}
 & & runway V & 39.1 & 82.9 & 35.7 & 18.6 & 59.6 \\
 & & runway T & 47.6 & 88.5 & 46.0 & 43.2 & 51.9 \\
\midrule
\multirow{4}{*}{3} & \multirow{4}{*}{\shortstack[l]{HIT-UAV\textsuperscript{\textdagger}\\+dirt-road\\+runway}} & dirt-road V & 28.4 & 70.1 & 9.3 & 16.3 & 40.4 \\
 & & dirt-road T & 39.1 & 85.0 & 29.1 & 32.1 & 46.0 \\
\cmidrule{3-8}
 & & runway V & 47.3 & 89.2 & 46.8 & 33.7 & 60.8 \\
 & & runway T & 46.2 & 87.7 & 43.6 & 43.5 & 48.8 \\
\midrule
\multirow{7}{*}{4} & \multirow{7}{*}{HIT-UAV} & dirt-road V & \CC{novelcolor}3.9 & \CC{novelcolor}10.4 & \CC{novelcolor}1.6 & \CC{novelcolor}5.2 & \CC{novelcolor}2.5 \\
 & & dirt-road T & \CC{novelcolor}8.9 & \CC{novelcolor}21.3 & \CC{novelcolor}5.3 & \CC{novelcolor}17.0 & \CC{novelcolor}0.8 \\
\cmidrule{3-8}
 & & runway V & \CC{novelcolor}28.7 & \CC{novelcolor}65.5 & \CC{novelcolor}19.5 & \CC{novelcolor}27.6 & \CC{novelcolor}29.9 \\
 & & runway T & \CC{novelcolor}12.7 & \CC{novelcolor}39.5 & \CC{novelcolor}4.5 & \CC{novelcolor}17.0 & \CC{novelcolor}8.4 \\
\cmidrule{3-8}
 & & HIT-UAV V & 46.6 & 82.6 & 47.1 & 39.6 & 53.5 \\
 & & HIT-UAV T & 48.0 & 84.1 & 48.9 & 41.0 & 55.1 \\
\midrule
\multirow{2}{*}{5} & \multirow{2}{*}{\shortstack[l]{dirt-road\\+runway}} & HIT-UAV V & \CC{novelcolor}6.7 & \CC{novelcolor}17.5 & \CC{novelcolor}3.4 & \CC{novelcolor}10.3 & \CC{novelcolor}3.1 \\
& & HIT-UAV T & \CC{novelcolor}7.1 & \CC{novelcolor}18.5 & \CC{novelcolor}4.3 & \CC{novelcolor}11.0 & \CC{novelcolor}3.3 \\
\midrule
\multirow{4}{*}{6} & \multirow{4}{*}{dirt-road} & dirt-road V & 22.0 & 67.5 & 4.3 & 15.2 & 28.9 \\
& & dirt-road T & 32.7 & 79.4 & 19.6 & 33.0 & 32.4 \\
\cmidrule{3-8}
 & & runway V & \CC{novelcolor}20.4 & \CC{novelcolor}42.2 & \CC{novelcolor}18.2 & \CC{novelcolor}40.5 & \CC{novelcolor}0.3 \\
 & & runway T & \CC{novelcolor}18.0 & \CC{novelcolor}50.7 & \CC{novelcolor}9.0 & \CC{novelcolor}19.2 & \CC{novelcolor}16.8 \\
\midrule
\multirow{4}{*}{7} & \multirow{4}{*}{runway} & dirt-road V & \CC{novelcolor}15.2 & \CC{novelcolor}44.5 & \CC{novelcolor}3.8 & \CC{novelcolor}1.5 & \CC{novelcolor}29.0 \\
 & & dirt-road T & \CC{novelcolor}21.3 & \CC{novelcolor}58.7 & \CC{novelcolor}5.3 & \CC{novelcolor}15.0 & \CC{novelcolor}27.7 \\
\cmidrule{3-8}
 & & runway V & 38.8 & 82.5 & 35.7 & 19.2 & 58.4 \\
 & & runway T & 46.2 & 87.5 & 43.0 & 45.6 & 46.8 \\
\bottomrule
\end{tabular}
 }
\end{table}

%%%%%%%%%%%%%%%%%%%%%%%%%%%%%%%%%%%%%%%%%%%%%%%%%%%%%%%%%%%%%%%%%%%%%%%%%%%%%%%%
\noindent \textbf{Transfer learning analysis.}
Tab.~\ref{tab:results_faster_rcnn} reports different experiments of our transfer learning analysis that we obtained by using the Faster R-CNN detector \cite{Shaoqing2015}.
We chose this detector because it is widely used in several benchmarks \cite{Bondi2020,Varga2022}.
We train models that are pre-trained on COCO \cite{Lin2014}.

In Exp.~1 we combine dirt-road and runway training data, and evaluate the performance on their respective validation and test splits.
In addition to observing that the validation split is more challenging than the test split,
the \texttt{person} category results to be the most difficult one to detect.
This is mainly due to the background heat that makes the targets difficult to distinguish.
In Exp.~2 we train without the bounding boxes that are flagged as occluded (still evaluating with the bounding boxes flagged occluded).
We can observe that the results are similar between the two experiments.
This suggests that the use of bounding boxes flagged occluded appears to marginally help the \texttt{person} class, while slightly affecting the \texttt{vehicle} class.
Because HIT-UAV contains both \texttt{person} and \texttt{vehicle}, in Exp.~3 we pre-train our detector on HIT-UAV \cite{Suo2022} (in addition to starting from the model pre-trained on COCO) and then train with dirt-road+runway.
HIT-UAV pre-training leads to improved performance compared to Exp.~1.
In Exp.~4, we evaluate the transfer learning ability of the detector from HIT-UAV to both dirt-road and runway.
Although HIT-UAV and MONET's categories and sensor modalities (thermal) are the same, experiments show that this is a rather challenging setting, i.e.~the detector poorly generalises between these scenarios.
Performances on runway are higher than those on dirt-road.
This can be due to the fact that the structure of the environment of runway is more similar to that of HIT-UAV.
We also report the upper bound on HIT-UAV in Exp.~4.
In Exp.~5, we can see that there is poor transfer learning ability when training is on dirt-road+runway and evaluation is on HIT-UAV.
Compared to HIT-UAV upper bound, we can observe that the performance gap in transfer learning is rather large.
In Exp.~6 \& 7 we report the transfer learning experiments focused on MONET's scenarios.
Despite being recorded with the same sensor, training on one scenario and testing on the other leads to lower performances than the same scenario setting.

\begin{table}[t]
\centering
\tabcolsep 2pt
\caption{Evaluation of several detectors initialised with COCO trained weights. 
Values are provided in percentage.
Grey background indicates transfer learning results. Keys:
F.~R-CNN \cite{Shaoqing2015}: Faster R-CNN. Def.~DETR \cite{Carion2020}: Deformable DETR \cite{Carion2020}. \textdagger: with data augmentations of the original paper.
Bold indicates best, underline indicates second best.}
\vspace{-.2cm}
\label{tab:results_detectors}
\resizebox{1\columnwidth}{!}{
\begin{tabular}{lllccccc}
\toprule
\multirow{2}{*}{Exp} & \multirow{2}{*}{Train/Eval} & \multirow{2}{*}{Detector} & \multirow{2}{*}{AP} & \multirow{2}{*}{AP$_{50}$} & \multirow{2}{*}{AP$_{75}$} & \multicolumn{2}{c}{class-AP} \\
 & & & & & & person & vehicle \\
\midrule
\multirow{11}{*}{1} & \multirow{11}{*}{\shortstack[l]{dirt-road/\\dirt-road}} & F.~R-CNN \cite{Shaoqing2015} & 22.0 & \underline{67.5} & 4.3 & \underline{15.2} & 28.9 \\
    &                         & SSD \cite{Liu2016}                         & 19.6                    & 64.2                    & 4.4                     & 11.3                    & 28.0                    \\
    &                         & CornerNet \cite{Law2018}                   & 10.1                    & 46.8                    & 0.2                     & 1.3                     & 18.9                    \\
    &                         & FCOS \cite{Tian2019}                        & 14.7                    & 55.2                    & 0.4                     & 3.1                     & 26.3                    \\
    &                         & DETR \cite{Carion2020}                        & 12.5                    & 44.1                    & 0.7                     & 0.7                     & 24.3                    \\
    &                         & Def.~DETR \cite{Carion2020}                   & 16.5                    & 55.2                    & 1.4                     & 3.4                     & 29.6                    \\
    &                         & VarifocalNet \cite{Zhang2020}                & 21.4                    & 61.5                    & 3.6                     & 8.2                     & 34.7                    \\
    &                         & ObjectBox \cite{Zand2022}                   & \textbf{26.4}           & \textbf{72.5}           & \underline{5.7}             & \textbf{16.4}           & \underline{36.5}            \\
    &                         & YOLOv8 \cite{Yolov8}                      & \underline{25.1}            & 64.9                    & \textbf{5.9}            & 11.0                    & \textbf{39.3}           \\
\cmidrule{3-8}
    &                         & ObjectBox\textsuperscript{\textdagger} \cite{Zand2022} & 31.4                    & 68.1                    & \textbf{19.3}                     & 15.4                     & \textbf{47.4}                    \\
    &                         & YOLOv8\textsuperscript{\textdagger} \cite{Yolov8}    & \textbf{33.3}           & \textbf{76.0}           & 15.6           & \textbf{22.2}           & 44.4           \\
\midrule
\multirow{11}{*}{2} & \multirow{11}{*}{\shortstack[l]{runway/\\dirt-road}}    & F.~R-CNN \cite{Shaoqing2015}                    & \CC{novelcolor}15.2          & \CC{novelcolor}44.5          & \CC{novelcolor}3.8           & \CC{novelcolor}\underline{1.5}   & \CC{novelcolor}29.0          \\
    &                         & SSD \cite{Liu2016}                         & \CC{novelcolor}\textbf{21.9} & \CC{novelcolor}\underline{47.0}  & \CC{novelcolor}17.6          & \CC{novelcolor}\textbf{1.6}  & \CC{novelcolor}\textbf{42.1} \\
    &                         & CornerNet \cite{Law2018}                   & \CC{novelcolor}18.9          & \CC{novelcolor}35.6          & \CC{novelcolor}\underline{22.7}  & \CC{novelcolor}0.2           & \CC{novelcolor}37.5          \\
    &                         & FCOS \cite{Tian2019}                        & \CC{novelcolor}19.2          & \CC{novelcolor}\underline{47.0}  & \CC{novelcolor}10.8          & \CC{novelcolor}0.3           & \CC{novelcolor}38.0          \\
    &                         & DETR \cite{Carion2020}                        & \CC{novelcolor}8.8           & \CC{novelcolor}26.9          & \CC{novelcolor}0.7           & \CC{novelcolor}0.0           & \CC{novelcolor} 17.6         \\
    &                         & Def.~DETR \cite{Carion2020}                   & \CC{novelcolor}8.0           & \CC{novelcolor}31.3          & \CC{novelcolor}0.3           & \CC{novelcolor}0.3           & \CC{novelcolor}15.8          \\
    &                         & VarifocalNet \cite{Zhang2020}                & \CC{novelcolor}19.3          & \CC{novelcolor}\textbf{47.1} & \CC{novelcolor}14.9          & \CC{novelcolor}1.2           & \CC{novelcolor}37.5          \\
    &                         & ObjectBox \cite{Zand2022}                   & \CC{novelcolor}14.9          & \CC{novelcolor}36.9          & \CC{novelcolor}4.8           & \CC{novelcolor}1.0           & \CC{novelcolor}28.8          \\
    &                         & YOLOv8 \cite{Yolov8}                      & \CC{novelcolor}\underline{20.8}  & \CC{novelcolor}34.0          & \CC{novelcolor}\textbf{25.8} & \CC{novelcolor}0.5           & \CC{novelcolor}\underline{41.0}  \\
\cmidrule{3-8}
    &                         & ObjectBox\textsuperscript{\textdagger} \cite{Zand2022} & \CC{novelcolor}\textbf{32.6}          & \CC{novelcolor}62.9          & \CC{novelcolor}28.4          & \CC{novelcolor}\textbf{12.1}           & \CC{novelcolor}53.0          \\
    &                         & YOLOv8\textsuperscript{\textdagger} \cite{Yolov8}    & \CC{novelcolor}\textbf{32.6} & \CC{novelcolor}\textbf{63.4} & \CC{novelcolor}\textbf{32.3} & \CC{novelcolor}11.4  & \CC{novelcolor}\textbf{53.8} \\
\midrule
\multirow{11}{*}{3} & \multirow{11}{*}{\shortstack[l]{runway/\\runway}}       & F.~R-CNN \cite{Shaoqing2015}                    & 38.8                    & 82.5                    & 35.7                    & 19.2                    & 58.4                    \\
    &                         & SSD \cite{Liu2016}                         & 42.6                    & 85.1                    & 40.9                    & 25.2                    & 60.0                    \\
    &                         & CornerNet \cite{Law2018}                   & 39.6                    & 76.4                    & 37.1                    & 20.2                    & 58.9                    \\
    &                         & FCOS \cite{Tian2019}                        & 44.5                    & 88.6                    & 41.3                    & 28.9                    & 60.2                    \\
    &                         & DETR \cite{Carion2020}                        & 31.2                    & 81.2                    & 20.2                    & 16.0                    & 46.4                    \\
    &                         & Def.~DETR \cite{Carion2020}                   & 44.1                    & \textbf{94.7}           & 39.9                    & 28.8                    & 59.5                    \\
    &                         & VarifocalNet \cite{Zhang2020}                & \underline{49.5}            & 92.4                    & 46.3                    & 37.1                    & \underline{61.9}            \\
    &                         & ObjectBox \cite{Zand2022}                   & \underline{49.5}            & \underline{93.7}            & \underline{48.3}            & \underline{39.7}            & 59.2                    \\
    &                         & YOLOv8 \cite{Yolov8}                      & \textbf{53.5}           & 93.0                    & \textbf{58.8}           & \textbf{42.0}           & \textbf{65.1}           \\
\cmidrule{3-8}
    &                         & ObjectBox\textsuperscript{\textdagger} \cite{Zand2022} & 51.4                   & 95.0                    & 49.9                    & 41.3                    & \textbf{61.6}                    \\
    &                         & YOLOv8\textsuperscript{\textdagger} \cite{Yolov8}    & \textbf{52.4}           & \textbf{95.8}           & \textbf{54.9}           & \textbf{45.0}           & 59.8           \\
\midrule
\multirow{11}{*}{4} & \multirow{11}{*}{\shortstack[l]{dirt-road/\\runway}}     & F.~R-CNN \cite{Shaoqing2015}                    & \CC{novelcolor}20.4          & \CC{novelcolor}42.2          & \CC{novelcolor}18.2          & \CC{novelcolor}\textbf{40.5} & \CC{novelcolor}0.3 \\
    &                         & SSD \cite{Liu2016}                         & \CC{novelcolor}18.6          & \CC{novelcolor}43.1          & \CC{novelcolor}13.6          & \CC{novelcolor}37.0          & \CC{novelcolor}0.3           \\
    &                         & CornerNet \cite{Law2018}                   & \CC{novelcolor}22.7          & \CC{novelcolor}51.9          & \CC{novelcolor}18.0          & \CC{novelcolor}20.2          & \CC{novelcolor}25.2          \\
    &                         & FCOS \cite{Tian2019}                        & \CC{novelcolor}19.0          & \CC{novelcolor}53.4          & \CC{novelcolor}9.4           & \CC{novelcolor}32.1          & \CC{novelcolor}5.9           \\
    &                         & DETR \cite{Carion2020}                        & \CC{novelcolor}13.9          & \CC{novelcolor}48.0          & \CC{novelcolor}3.5           & \CC{novelcolor}19.0          & \CC{novelcolor}8.8           \\
    &                         & Def.~DETR \cite{Carion2020}                   & \CC{novelcolor}\textbf{32.9} & \CC{novelcolor}\textbf{73.0} & \CC{novelcolor}\underline{26.0}  & \CC{novelcolor}33.8          & \CC{novelcolor}\textbf{32.0} \\
    &                         & VarifocalNet \cite{Zhang2020}                & \CC{novelcolor}\underline{30.7}  & \CC{novelcolor}\textbf{73.0} & \CC{novelcolor}21.8          & \CC{novelcolor}35.3          & \CC{novelcolor}\underline{26.1}  \\
    &                         & ObjectBox \cite{Zand2022}                   & \CC{novelcolor}27.2          & \CC{novelcolor}59.7          & \CC{novelcolor}19.8          & \CC{novelcolor}\underline{39.0} & \CC{novelcolor}15.4          \\
    &                         & YOLOv8 \cite{Yolov8}                      & \CC{novelcolor}\underline{30.7}  & \CC{novelcolor}\underline{60.2}  & \CC{novelcolor}\textbf{28.3} & \CC{novelcolor}38.6  & \CC{novelcolor}22.7          \\
\cmidrule{3-8}
    &                         & ObjectBox\textsuperscript{\textdagger} \cite{Zand2022} & \CC{novelcolor}\textbf{32.4} & \CC{novelcolor}\textbf{63.2} & \CC{novelcolor}\textbf{29.7} & \CC{novelcolor}34.7 & \CC{novelcolor}\textbf{30.1} \\
    &                         & YOLOv8\textsuperscript{\textdagger} \cite{Yolov8}    & \CC{novelcolor}27.8          & \CC{novelcolor}55.4          & \CC{novelcolor}24.8          & \CC{novelcolor}\textbf{34.9}          & \CC{novelcolor}20.7 \\
\bottomrule
\end{tabular}
}
\end{table}

%%%%%%%%%%%%%%%%%%%%%%%%%%%%%%%%%%%%%%%%%%%%%%%%%%%%%%%%%%%%%%%%%%%%%%%%%%%%%%%%
\noindent \textbf{Detector comparisons.}
Tab.~\ref{tab:results_detectors} reports the comparisons amongst the different detectors.
Experiments were executed by including the bounding boxes flagged occluded and by zeroing the areas marked with the \texttt{ignore} label.

In the same-scenario setting, YOLOv8 and ObjectBox are the best performing ones.
In the transfer learning setting, SSD and Def.~DETR are the best performing ones, while YOLOv8 and VarifocalNet consistently perform second best.
YOLOv8 consistently outperforms the other detectors in terms of AP$_{75}$, which indicates its superior ability in estimating the correct bounding box sizes.
In terms of AP$_{50}$, results are mixed.
Although the two settings were captured with the same camera and in similar locations, we can observe that object detection is generally rather challenging for all the detectors, especially in the transfer learning setting.
All detectors perform poorly in dirt-road, in particular with the person category.
Dirt-road is more challenging because it was captured during daytime, where ground heat makes human targets more difficult to distinguish than the vehicle targets.
Except Def.~DETR, all the detectors perform poorly on the vehicle class in dirt-road/runway.
We believe that this is because the vehicles were captured from much fewer viewpoints in dirt-road than in runway, thus affecting generalisation when tested in runway.
YOLOv8's original data augmentation strategies appears to be effective in dirt-road but less effective in runway, suggesting that a more tailored design of data augmentation for this problem could help improving the detection accuracy.

% ********************************
\begin{figure*}[t]
\begin{center}
  \begin{tabular}{@{}c@{\,}c@{\,}c@{\,}c}
    \begin{overpic}[width=.25\linewidth]{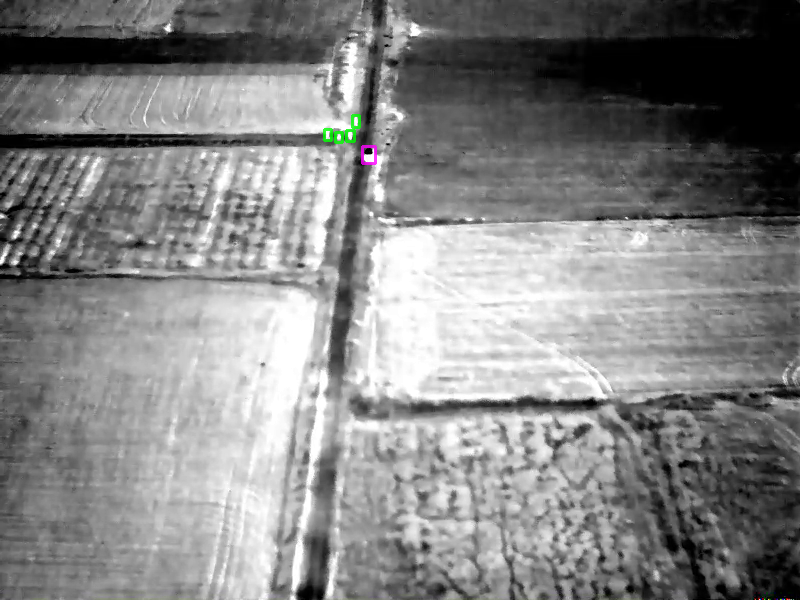}
    \put(2,3){\color{lime}\footnotesize \textbf{ground truth}}
    \end{overpic} &
    \begin{overpic}[width=.25\linewidth]{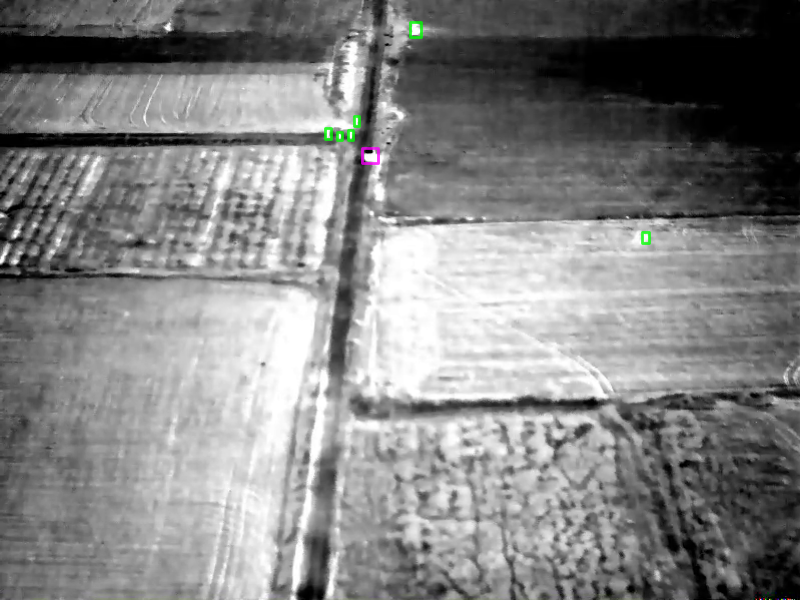}
    \put(2,3){\color{lime}\footnotesize \textbf{trained on dirt-road}}
    \end{overpic} &
    \begin{overpic}[width=.25\linewidth]{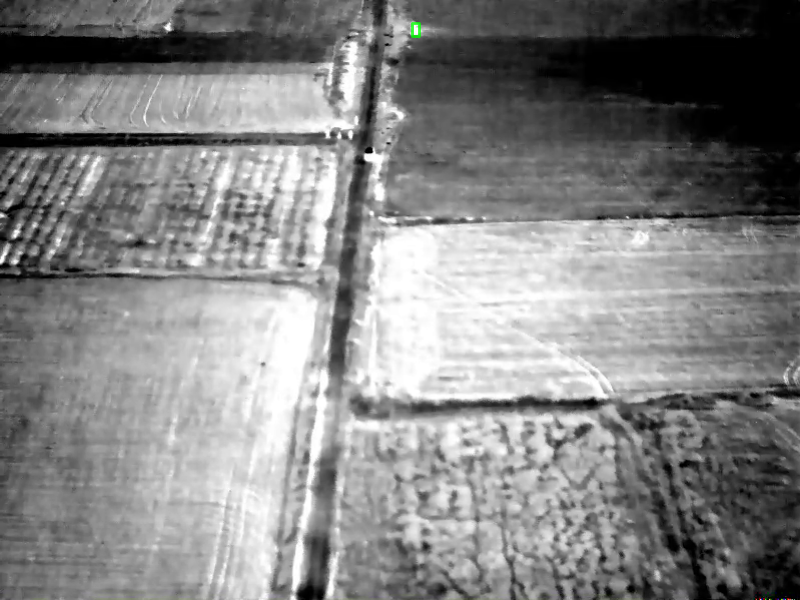}
    \put(2,3){\color{lime}\footnotesize \textbf{trained on runway}}
    \end{overpic} &
    \begin{overpic}[width=.25\linewidth]{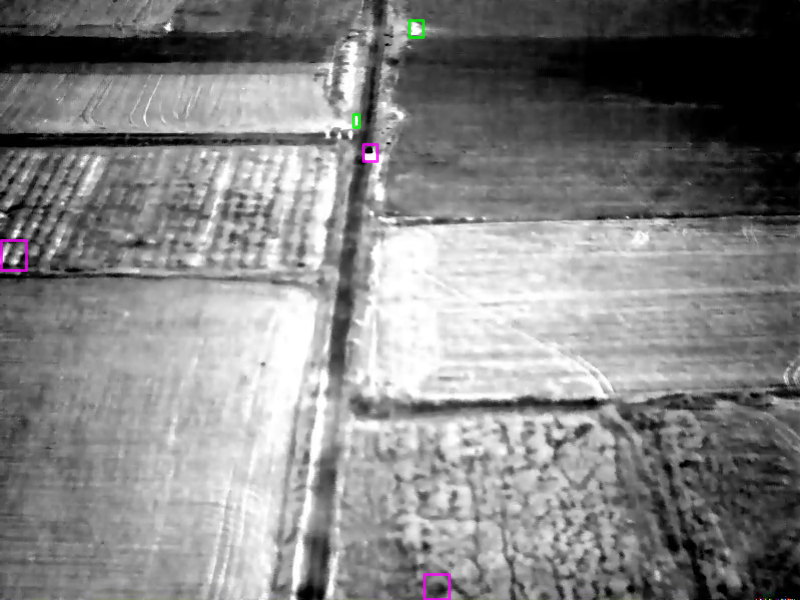}
    \put(2,3){\color{lime}\footnotesize \textbf{trained on HIT-UAV}}
    \end{overpic} \\
    \begin{overpic}[width=.25\linewidth]{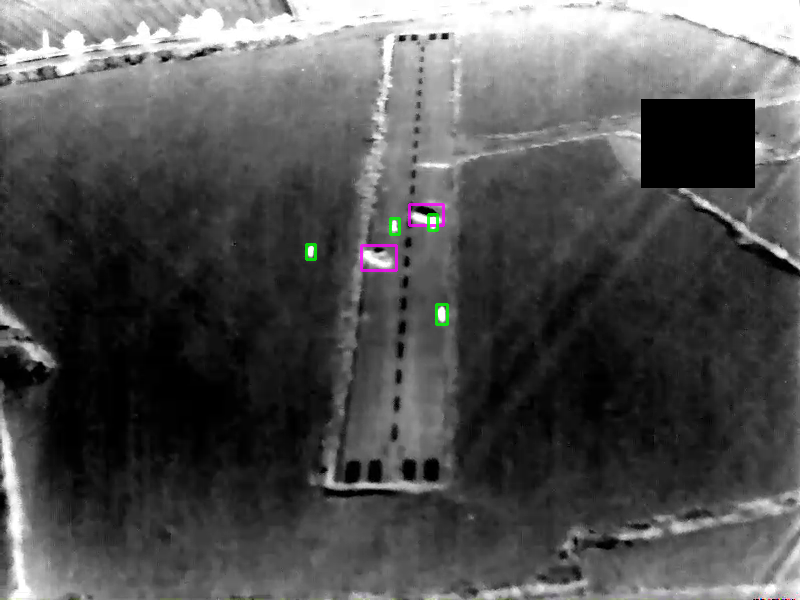}
    \put(2,3){\color{lime}\footnotesize \textbf{ground truth}}
    \end{overpic} &
    \begin{overpic}[width=.25\linewidth]{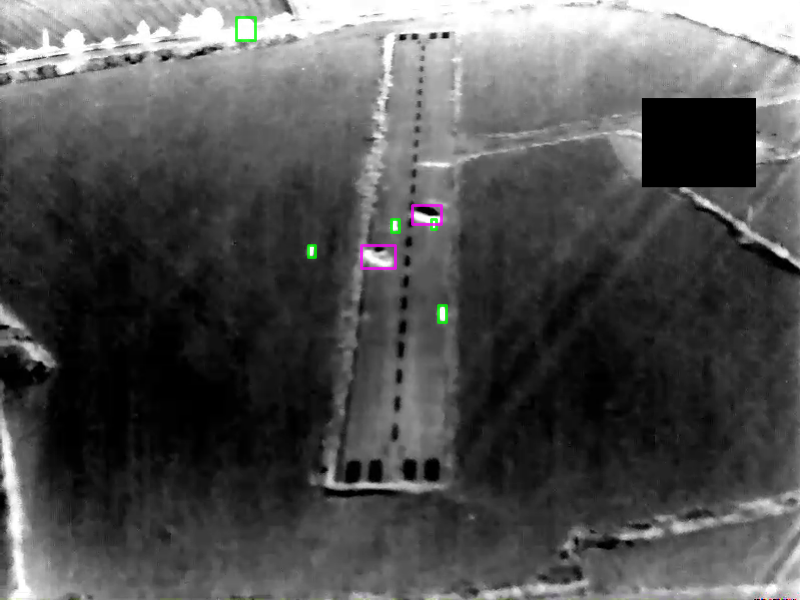}
    \put(2,3){\color{lime}\footnotesize \textbf{trained on dirt-road}}
    \end{overpic} &
    \begin{overpic}[width=.25\linewidth]{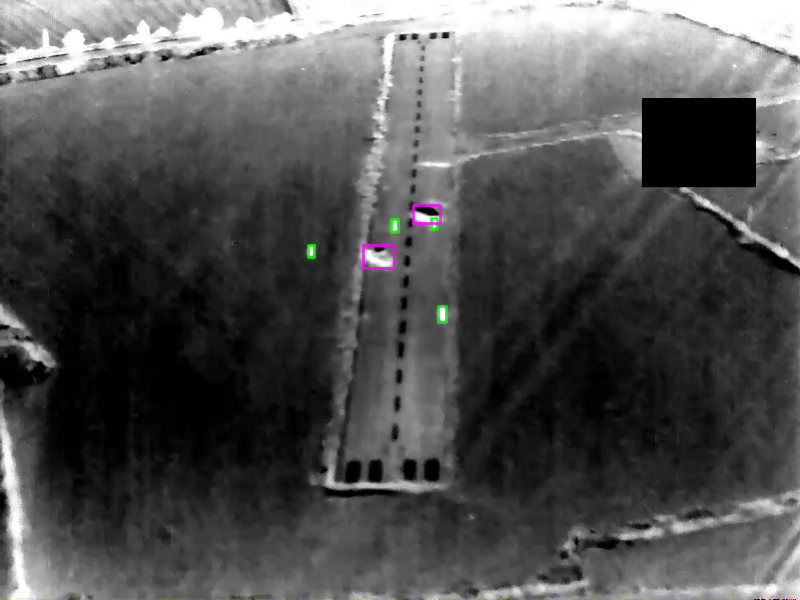}
    \put(2,3){\color{lime}\footnotesize \textbf{trained on runway}}
    \end{overpic} &
    \begin{overpic}[width=.25\linewidth]{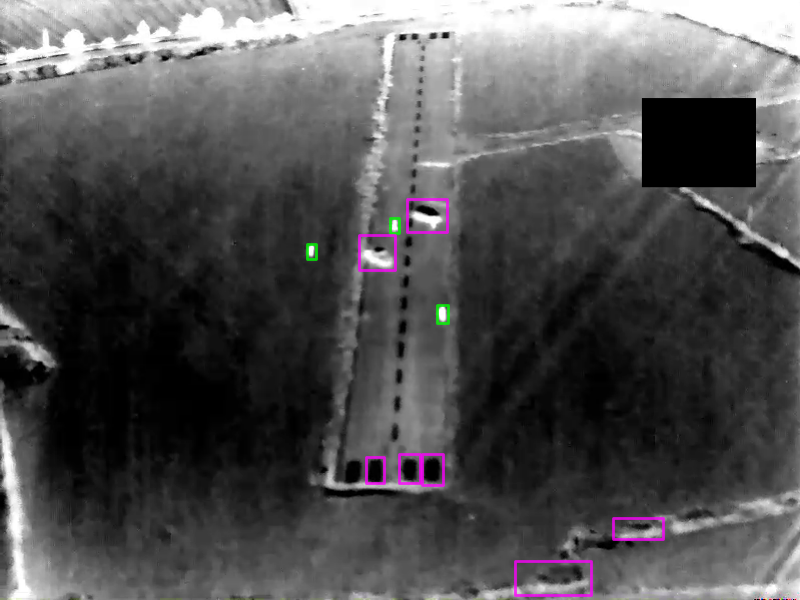}
    \put(2,3){\color{lime}\footnotesize \textbf{trained on HIT-UAV}}
    \end{overpic} \\
  \end{tabular}
\end{center}
\vspace{-.6cm}
\caption{Qualitative results showing the performances of Faster R-CNN in the dirt-road (first row) and runway (second row) scenarios with different training datasets.
Bounding boxes: green for \texttt{person}, magenta for \texttt{vehicle}.}
\label{fig:qualitative_results_monet}
\end{figure*}
% ********************************
% ********************************
\begin{figure}[t]
\begin{center}
  \begin{tabular}{@{}c@{\,}c}
    \begin{overpic}[width=.5\linewidth]{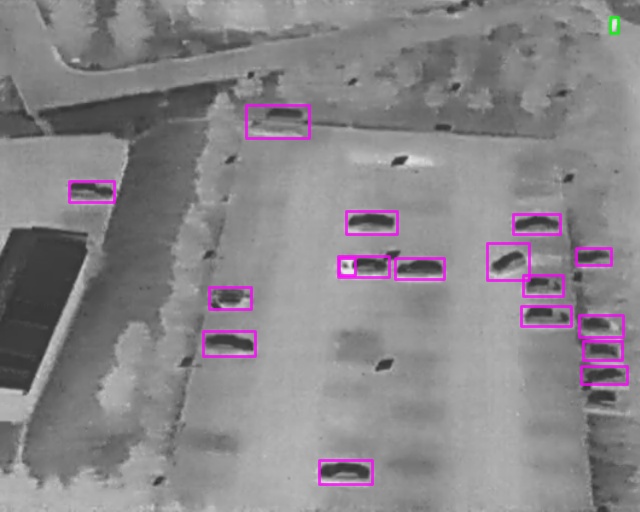}
    \put(2,3){\color{lime}\footnotesize \textbf{trained on HIT-UAV}}
    \end{overpic} &
    \begin{overpic}[width=.5\linewidth]{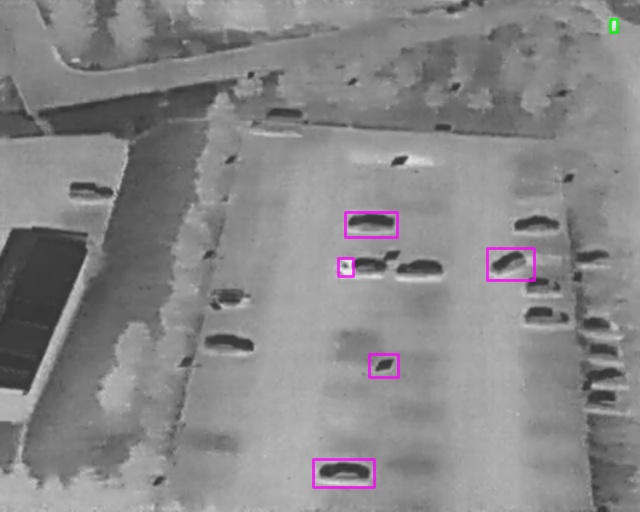}
    \put(2,3){\color{lime}\footnotesize \textbf{trained on dirt-road+runway}}
    \end{overpic} \\
  \end{tabular}
\end{center}
\vspace{-.6cm}
\caption{Qualitative results showing the performances of Faster R-CNN in a scene from HIT-UAV with different training datasets.}
\label{fig:qualitative_results_hit_uav}
\end{figure}
% ********************************

%%%%%%%%%%%%%%%%%%%%%%%%%%%%%%%%%%%%%%%%%%%%%%%%%%%%%%%%%%%%%%%%%%%%%%%%%%%%%%%%
%%%%%%%%%%%%%%%%%%%%%%%%%%%%%%%%%%%%%%%%%%%%%%%%%%%%%%%%%%%%%%%%%%%%%%%%%%%%%%%%
\subsection{Qualitative results}

Fig.~\ref{fig:qualitative_results_monet} shows examples of Faster R-CNN's detections on dirt-road (first row) and on runway (second row) produced with different training data.
In the first case, training on dirt-road and evaluating on dirt-road leads to satisfactory results, when we train on runway the detector misses all the targets, and when we train on HIT-UAV the detector detects some targets plus some false positives.
Conversely, in the second case, training on dirt-road and evaluating on runway leads to good results, same as when we train on runway, however when we train on HIT-UAV the detector produces several false positive detections.
We observed that one of the reasons for this is that HIT-UAV contains several annotations of parked vehicles, which emits a low temperature, see example in Fig.~\ref{fig:qualitative_results_hit_uav}.
Hence, the detector relates dark patterns to vehicles, which are similar to the false positive detections in runway in Fig.~\ref{fig:qualitative_results_monet}.
Fig.~\ref{fig:qualitative_results_hit_uav} also shows that training on MONET and evaluating on HIT-UAV leads to poor results.
The Supplementary Material contains additional qualitative results from the nine detectors evaluated.

%%%%%%%%%%%%%%%%%%%%%%%%%%%%%%%%%%%%%%%%%%%%%%%%%%%%%%%%%%%%%%%%%%%%%%%%%%%%%%%%
%%%%%%%%%%%%%%%%%%%%%%%%%%%%%%%%%%%%%%%%%%%%%%%%%%%%%%%%%%%%%%%%%%%%%%%%%%%%%%%%
%%%%%%%%%%%%%%%%%%%%%%%%%%%%%%%%%%%%%%%%%%%%%%%%%%%%%%%%%%%%%%%%%%%%%%%%%%%%%%%%
\section{Conclusions}

We introduce MONET, a novel and challenging multimodal dataset for vision-based object localisation in the thermal spectrum from drones. 
MONET was collected with a drone that flew over rural areas and captured human and vehicle activities.
MONET comprises two scenarios in agricultural lands, namely \textit{dirt-road} and \textit{runway}. 
We benchmarked nine state-of-the-art object detection algorithms on MONET and found that they performed poorly due to the large scale variation of the targets, and the background clutter caused by the ground heat. 
The \textit{dirt-road} scenario is more difficult than the \textit{runway} scenario as it contains several more of the above-mentioned challenges than runway. 
To our knowledge, MONET is one of the few datasets in the thermal spectrum that provides a large number of manually annotated frames with timestamp-aligned metadata. 
Moreover, MONET includes bounding boxes with identities for each target, making it suitable for multi-object tracking research. 
We hope that MONET will foster further research, especially in the development of multimodal solutions for object localisation that exploits knowledge from metadata.

\vspace{1mm}
\noindent\textbf{Limitations.}
Although we collected both RGB and thermal images and annotated both of them, for now we will not include the RGB images in this release of MONET due to still unaddressed privacy concerns. 
It is worth noting that only a portion of the dataset contains RGB images captured in light, while the others were captured in darkness.

{\small
\bibliographystyle{ieee_fullname}
\bibliography{refs}
}

\clearpage

% --- repeat the title (AT: haven't found a more elegant way to do this...)
\twocolumn[
\centering
\Large
\textbf{Supplementary Material} \\
\vspace{1.0em}
] %< twocolumn
\appendix

\section{Introduction}

We provide some additional material in support of the main paper. 
The content is organised as follows:
\setlist{nolistsep}
\begin{itemize}
    \item In Sec.~\ref{sec:detector_setup} we provide details about the training configuration of each detector and describe each data augmentation we used.
    \item In Sec.~\ref{sec:dataset_stats} we analyse the statistics of each dataset split for the dirt-road and runway scenarios.
    \item In Sec.~\ref{sec:additional_qualitative_results} we provide examples of qualitative results taken from each detector that we evaluated.
\end{itemize}

%%%%%%%%%%%%%%%%%%%%%%%%%%%%%%%%%%%%%%%%%%%%%%%%%%%%%%%%%%%%%%%%%
%%%%%%%%%%%%%%%%%%%%%%%%%%%%%%%%%%%%%%%%%%%%%%%%%%%%%%%%%%%%%%%%%
%%%%%%%%%%%%%%%%%%%%%%%%%%%%%%%%%%%%%%%%%%%%%%%%%%%%%%%%%%%%%%%%%
\section{Detector setups}\label{sec:detector_setup}

In the main paper we have two sets of results that compare state-of-the-art detectors.
For both sets we use non-maximum suppression with IoU threshold 0.5 and we evaluate all the bounding boxes with confidence above $10^{-8}$.
The reason behind this choice is that the models are calibrated differently, hence their output confidences can not be directly compared.
Setting a low confidence threshold is to minimally filter detector predictions, resulting in a fairer comparison of their performance.

The first set compares nine detectors with training configurations that we set as similar as possible.
Tab.~\ref{tab:detectors_setup} reports the details of the chosen configurations.

We train all the detectors with the same data augmentation strategy.
Data augmentations are applied in the following order:
i) RandomCrop: This crops a portion of the image with a size determined by randomly sampling two independent values within the interval [0.8, 1.0] and by multiplying them by the height and width of the original image;
ii) Resize: This randomly resizes the eventually cropped image between (600, 800) and (300, 400) while keeping its original aspect ratio.
iii) RandomHorizontalFlip: This randomly flips the image horizontally with a probability of 0.7.
iv) Padding: This is applied to make all the images of the same size, i.e.~(600, 800).
v) Normalisation: This involves normalising the image pixels with a mean of 126.225 and a standard deviation of 73.338. Note that these normalisation factors differ from the standard ones computed on ImageNet and were explicitly calculated for the MONET dataset.

\begin{table}[t]
\centering
\tabcolsep 2pt
\caption{Detectors setup. 
Keys: 
BS: Batch Size.
LR: Learning Rate.
CosAnn.: Cosine Annealing.}
\vspace{-.2cm}
\label{tab:detectors_setup}
\resizebox{1\columnwidth}{!}{
\begin{tabular}{lccccccc}
\toprule
Detector & Epochs & BS & Backbone & LR & Schedule & Optimiser\\
\midrule
F.~R-CNN \cite{Shaoqing2015} & 20 & 24 & ResNet-50 & 1e-3 & CosAnn. & AdamW\\
SSD \cite{Liu2016} & 20 & 24 & VGG-16 & 1e-3 & CosAnn. & AdamW\\
CornerNet \cite{Law2018} & 20 & 9 & HourglassNet-104 & 1e-4 & CosAnn & AdamW\\
FCOS \cite{Tian2019} & 20 & 24 & ResNet-50 & 1e-4 & CosAnn & AdamW\\
DETR \cite{Carion2020} & 50 & 24 & ResNet-50 & 5e-5 & CosAnn & AdamW\\
Def.~DETR \cite{Zhu2021} & 20 & 9 & ResNet-50 & 5e-5 & CosAnn & AdamW\\
VarifocalNet \cite{Zhang2020} & 20 & 24 & ResNet-50 & 1e-4 & CosAnn & AdamW\\
\multirow{2}{*}{ObjectBox \cite{Zand2022}} & \multirow{2}{*}{20} & \multirow{2}{*}{24} & \multirow{2}{*}{YOLOv5 v6.0} & \multirow{2}{*}{1e-2} & Warmup & \multirow{2}{*}{SGD}\\
& & & & & CosAnn &\\
\multirow{2}{*}{YOLOv8 \cite{Yolov8}} & \multirow{2}{*}{20} & \multirow{2}{*}{24} & \multirow{2}{*}{YOLOv8.0x} & \multirow{2}{*}{1e-2} & Warmup & \multirow{2}{*}{SGD}\\
& & & & & CosAnn &\\
\midrule
\multirow{2}{*}{ObjectBox\textsuperscript{\textdagger}\cite{Zand2022}} & \multirow{2}{*}{40} & \multirow{2}{*}{24} & \multirow{2}{*}{YOLOv5 v6.0} & \multirow{2}{*}{1e-2} & Warmup & \multirow{2}{*}{SGD}\\
 & & & & & CosAnn &\\
\multirow{2}{*}{YOLOv8\textsuperscript{\textdagger}\cite{Yolov8}} & \multirow{2}{*}{40} & \multirow{2}{*}{24} & \multirow{2}{*}{YOLOv8.0x} & \multirow{2}{*}{1e-2} & Warmup & \multirow{2}{*}{SGD} &\\
& & & & & CosAnn &\\
\bottomrule
\end{tabular}
 }
\end{table}

The second set compares ObjectBox~\cite{Zand2022} and YOLOv8~\cite{Yolov8} with their original data augmentation.
We use the $^\dagger$ in the main paper to represent these setups.

The data augmentations we use for ObjectBox are applied in the following order:
i) Mosaic: This combines 4 images (600, 800) into a single image (1200, 1600). Padding is then applied to produce a squared image of size (1600, 1600);
ii) RandomAffine: This applies translation and scale operations. The scale factor is randomly sampled from [0.5, 1.5] while the independent vertical and horizontals shifts are randomly sampled in the interval [-160, 160], i.e.~ using a maximum absolute fraction of 0.1. The image is then resized to (800, 800);
iii) Blur: This blurs the image using a random kernel size sampled in the interval [3, 7] with a probability of 0.1;
iv) MedianBlur: This blurs the image using a median filter with random aperture linear size sampled in the interval [3, 7] with a probability of 0.1;
v) RandomHSV: This firstly converts the image to HSV colorspace. 
Then, three scalars are sampled with the intervals [0.985, 1.015], [0.3, 1.7], and [0.6, 1.4], which are used to multiply the original values of Hue, Saturation and Value, respectively.
Lastly, the image is converted back to RGB colorspace;
vi) RandomHorizontalFlip: This randomly flips the image horizontally with a probability of 0.5.

The data augmentations we use for YOLOv8 are applied in the following order:
i) Mosaic: This combines 4 images (600, 800) into a single image (1200, 1600). 
Padding is then applied to produce a squared image of size (1600, 1600);
ii) MixUp: This averages two mosaic images with a probability of 0.15; 
iii) RandomAffine: This applies translation and scale operations. The scale factor is randomly sampled from [0.1, 1.9] while vertical and horizontals shifts are independently randomly sampled in the interval [-160, 160], i.e.~ using a maximum absolute fraction of 0.1. The image is then resized to (800, 800);
iv) Blur: This blurs the image using a random kernel size sampled in the interval [3, 7] with a probability of 0.01;
v) MedianBlur: This blurs the image using a median filter with a random aperture linear size sampled in the interval [3, 7] with a probability of 0.01;
vi) CLAHE: This applies Contrast Limited Adaptive Histogram Equalisation with probability 0.01;
vii) RandomHSV: This firstly converts the image to HSV colorspace. 
Then, three scalars are sampled in the intervals [0.985, 1.015], [0.3, 1.7], and [0.6, 1.4], which are used to multiply the original values of Hue, Saturation and Value, respectively.
Lastly, the image is converted back to RGB colorspace;
viii) RandomHorizontalFlip: This randomly flips the image horizontally with a probability of 0.5.

%%%%%%%%%%%%%%%%%%%%%%%%%%%%%%%%%%%%%%%%%%%%%%%%%%%%%%%%%%%%%%%%%
%%%%%%%%%%%%%%%%%%%%%%%%%%%%%%%%%%%%%%%%%%%%%%%%%%%%%%%%%%%%%%%%%
%%%%%%%%%%%%%%%%%%%%%%%%%%%%%%%%%%%%%%%%%%%%%%%%%%%%%%%%%%%%%%%%%
% \section{Drone metadata specifications}\label{sec:metadata_specs}

% Tab.~\ref{tab:metadata} reports a detailed list of the metadata we collected along with their minimum and maximum values.
% The unit values included in the table indicate how that specific metadata is represented in the dataset files.
% \input{tabels/metadata.tex}

%%%%%%%%%%%%%%%%%%%%%%%%%%%%%%%%%%%%%%%%%%%%%%%%%%%%%%%%%%%%%%%%%
%%%%%%%%%%%%%%%%%%%%%%%%%%%%%%%%%%%%%%%%%%%%%%%%%%%%%%%%%%%%%%%%%
%%%%%%%%%%%%%%%%%%%%%%%%%%%%%%%%%%%%%%%%%%%%%%%%%%%%%%%%%%%%%%%%%
\section{Additional dataset statistics}\label{sec:dataset_stats}

\begin{figure}[t]
    \centering
    \includegraphics[width=\columnwidth]{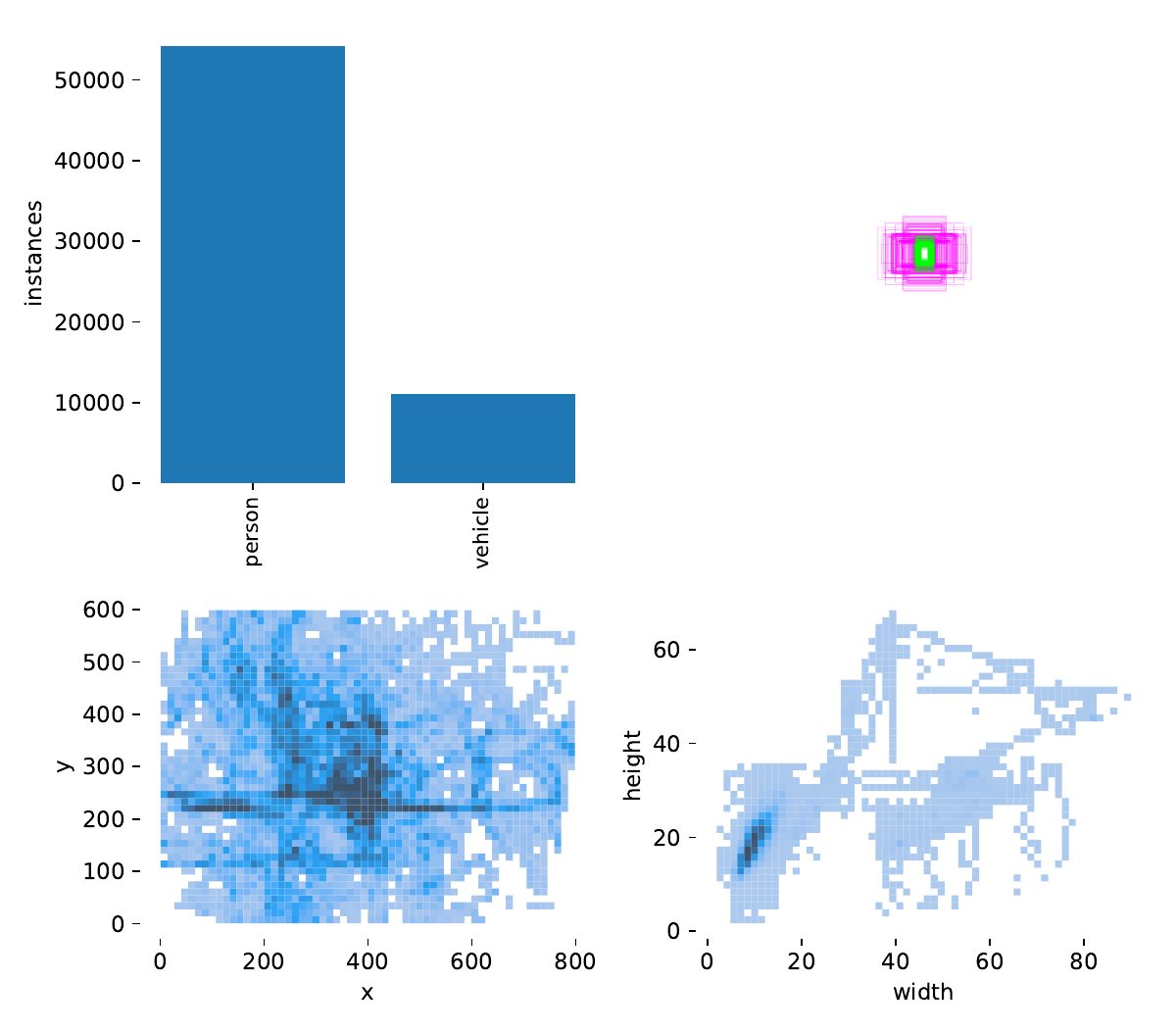}
    \caption{Bounding box ground-truth statistics of dirt-road train split.
    The top-left figure shows the histogram of bounding box instances.
    The top-right figure shows 1K examples of randomly sampled bounding boxes.
    The bottom-left figure shows the distribution of bounding box locations over the image plane.
    The bottom-right figure shows the distribution of bounding box sizes.}
    \label{fig:stats_train_dirtroad}
\end{figure}

\begin{figure}[t]
    \centering
    \includegraphics[width=\columnwidth]{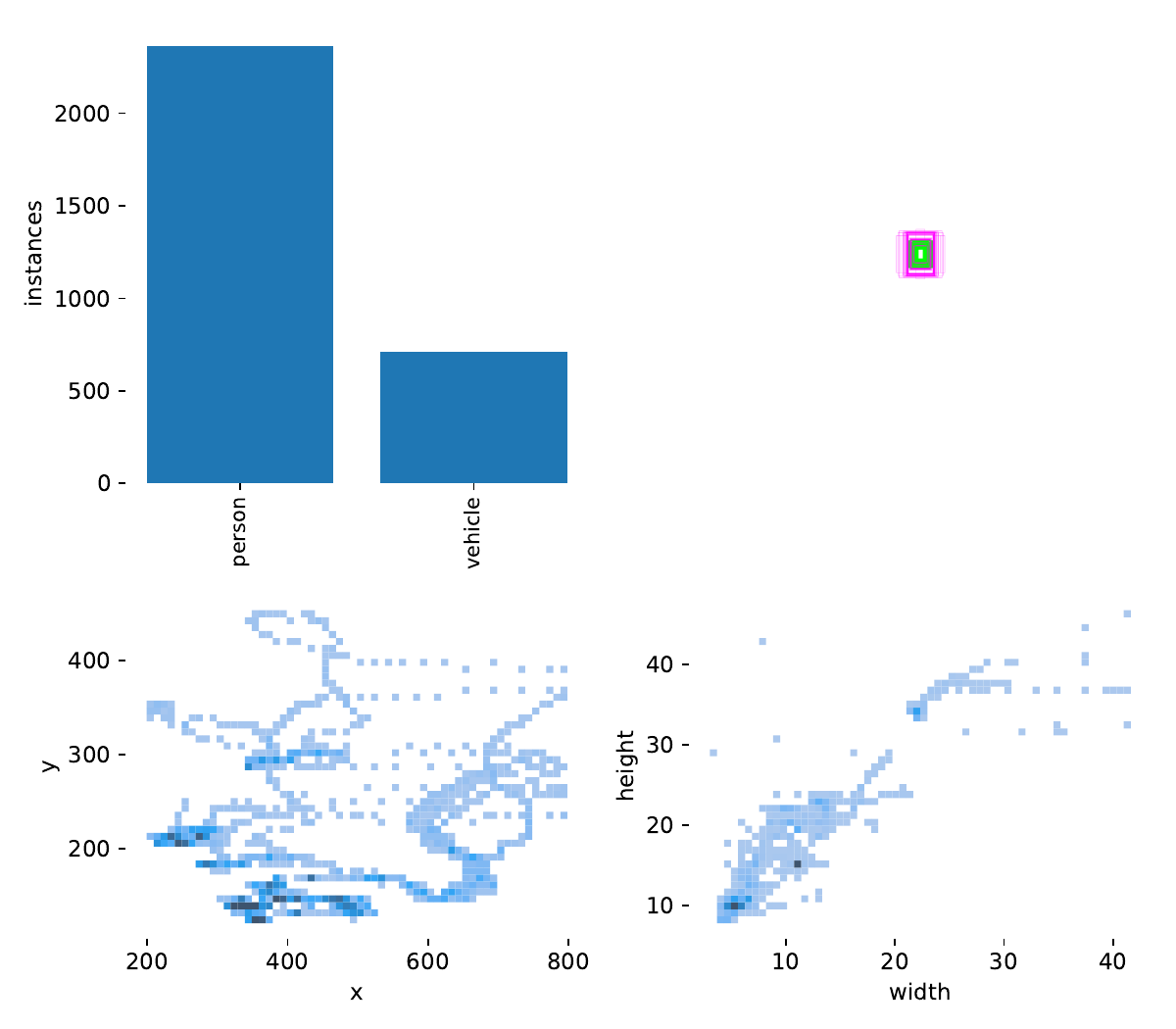}
    \caption{Bounding box ground-truth statistics of dirt-road validation split.
    The top-left figure shows the histogram of bounding box instances.
    The top-right figure shows 1K examples of randomly sampled bounding boxes.
    The bottom-left figure shows the distribution of bounding box locations over the image plane.
    The bottom-right figure shows the distribution of bounding box sizes.}
    \label{fig:stats_val_dirtroad}
\end{figure}

\begin{figure}[t]
    \centering
    \includegraphics[width=\columnwidth]{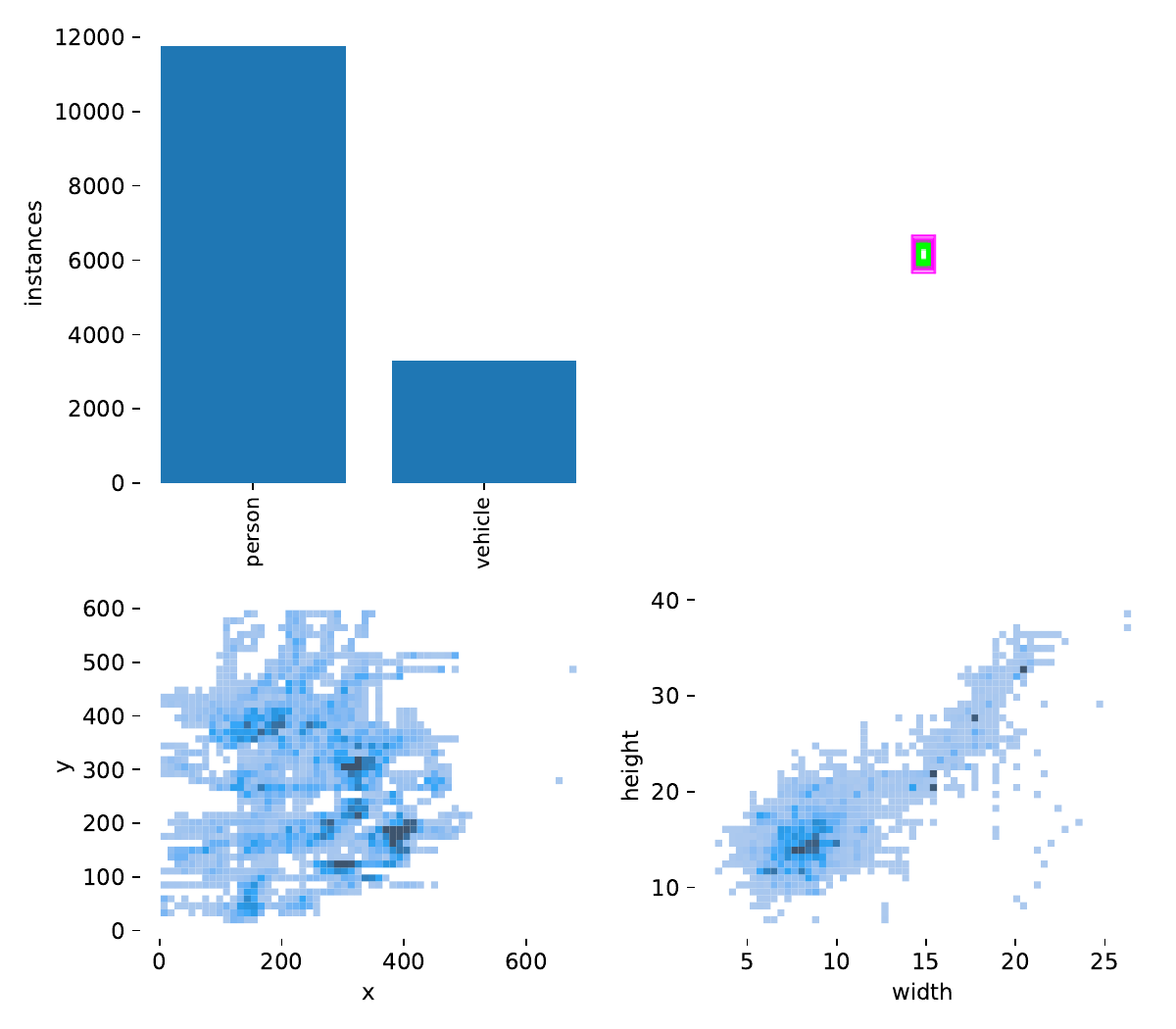}
    \caption{Bounding box ground-truth statistics of dirt-road test split.
    The top-left figure shows the histogram of bounding box instances.
    The top-right figure shows 1K examples of randomly sampled bounding boxes.
    The bottom-left figure shows the distribution of bounding box locations over the image plane.
    The bottom-right figure shows the distribution of bounding box sizes.}
    \label{fig:stats_test_dirtroad}
\end{figure}

\begin{figure}[t]
    \centering
    \includegraphics[width=\columnwidth]{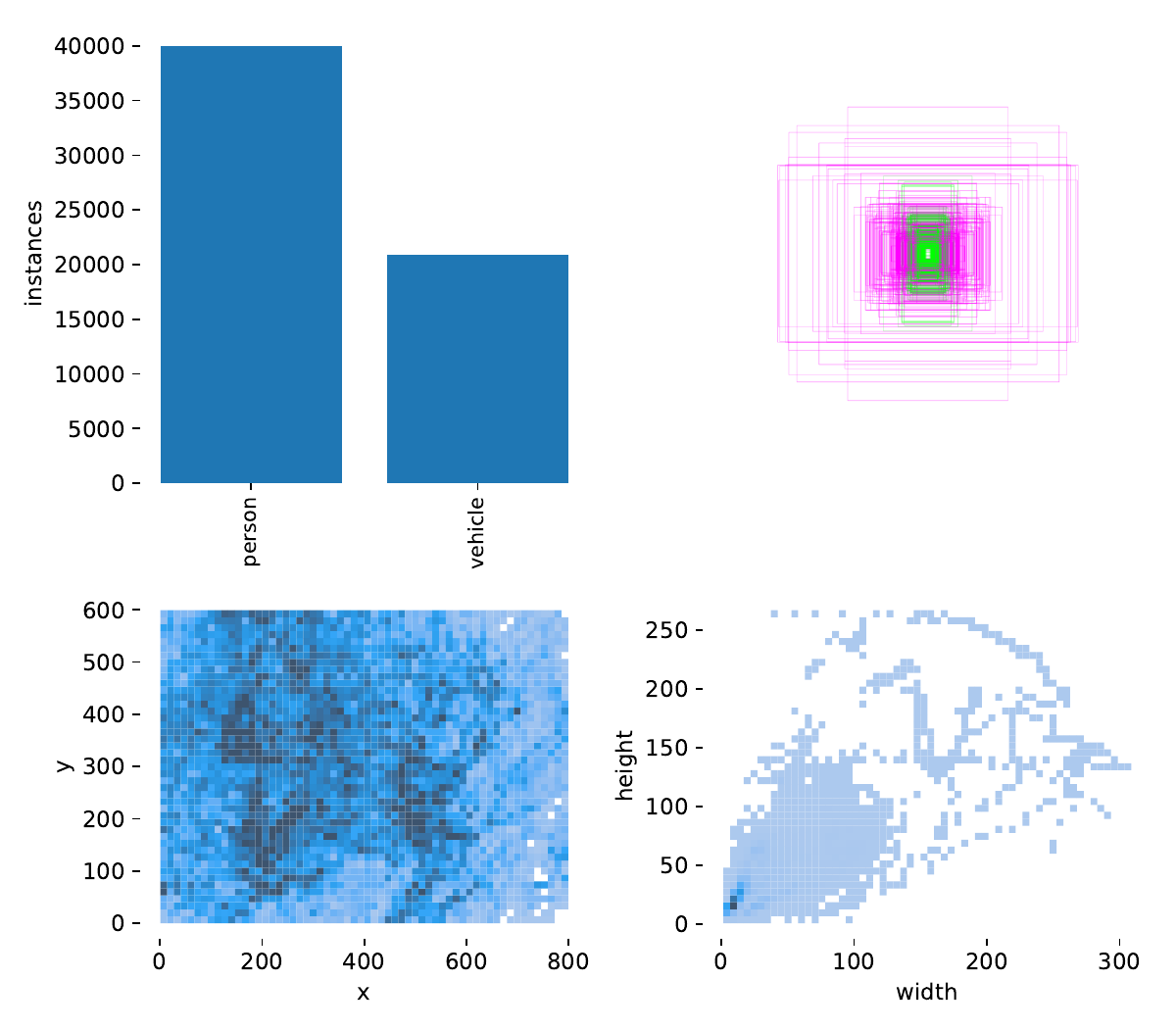}
    \caption{Bounding box ground-truth statistics of runway train split.
    The top-left figure shows the histogram of bounding box instances.
    The top-right figure shows 1K examples of randomly sampled bounding boxes.
    The bottom-left figure shows the distribution of bounding box locations over the image plane.
    The bottom-right figure shows the distribution of bounding box sizes.}
    \label{fig:stats_train_runway}
\end{figure}

\begin{figure}[t]
    \centering
    \includegraphics[width=\columnwidth]{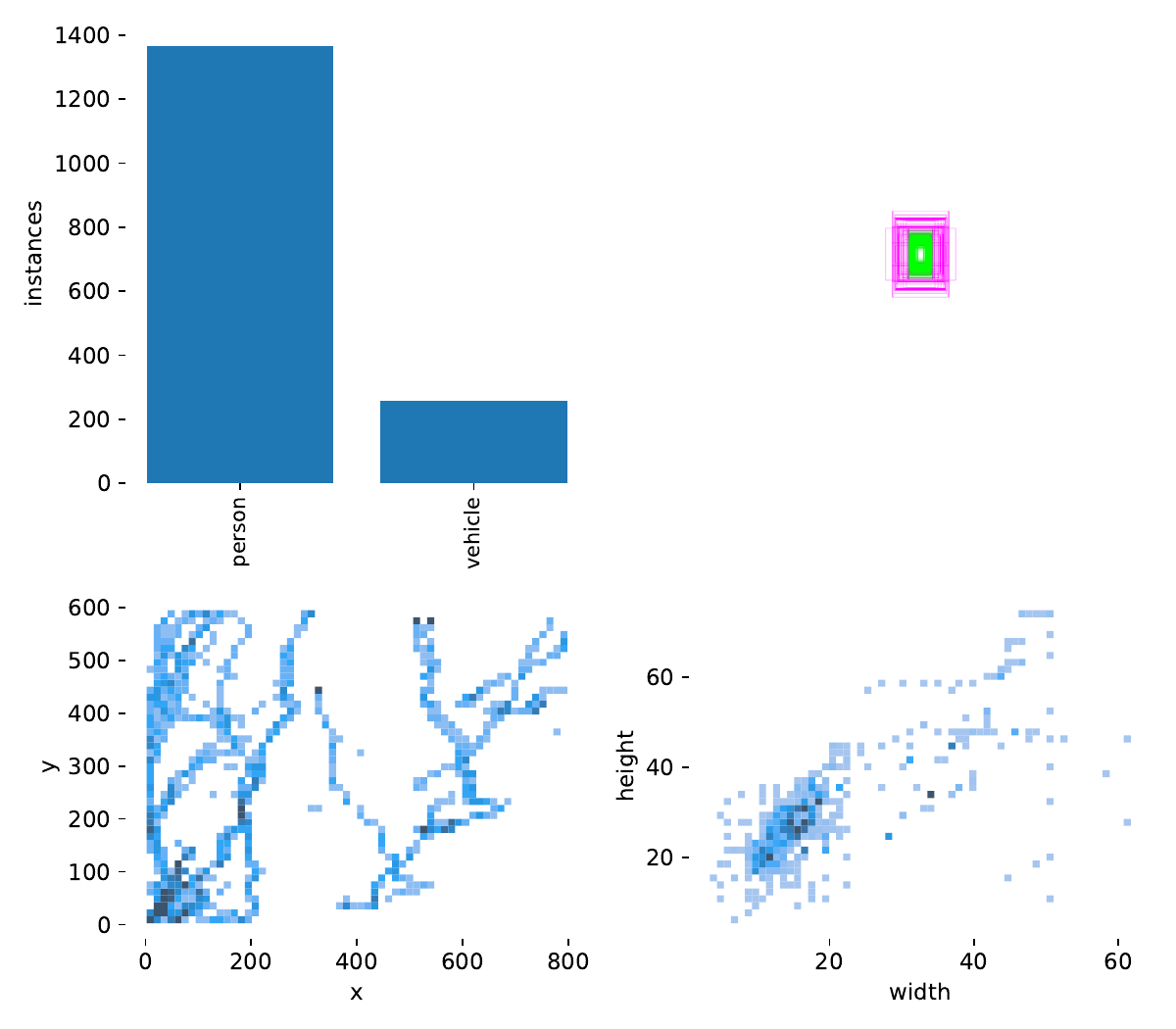}
    \caption{Bounding box ground-truth statistics of runway validation split.
    The top-left figure shows the histogram of bounding box instances.
    The top-right figure shows 1K examples of randomly sampled bounding boxes.
    The bottom-left figure shows the distribution of bounding box locations over the image plane.
    The bottom-right figure shows the distribution of bounding box sizes.}
    \label{fig:stats_val_runway}
\end{figure}

\begin{figure}[t]
    \centering
    \includegraphics[width=\columnwidth]{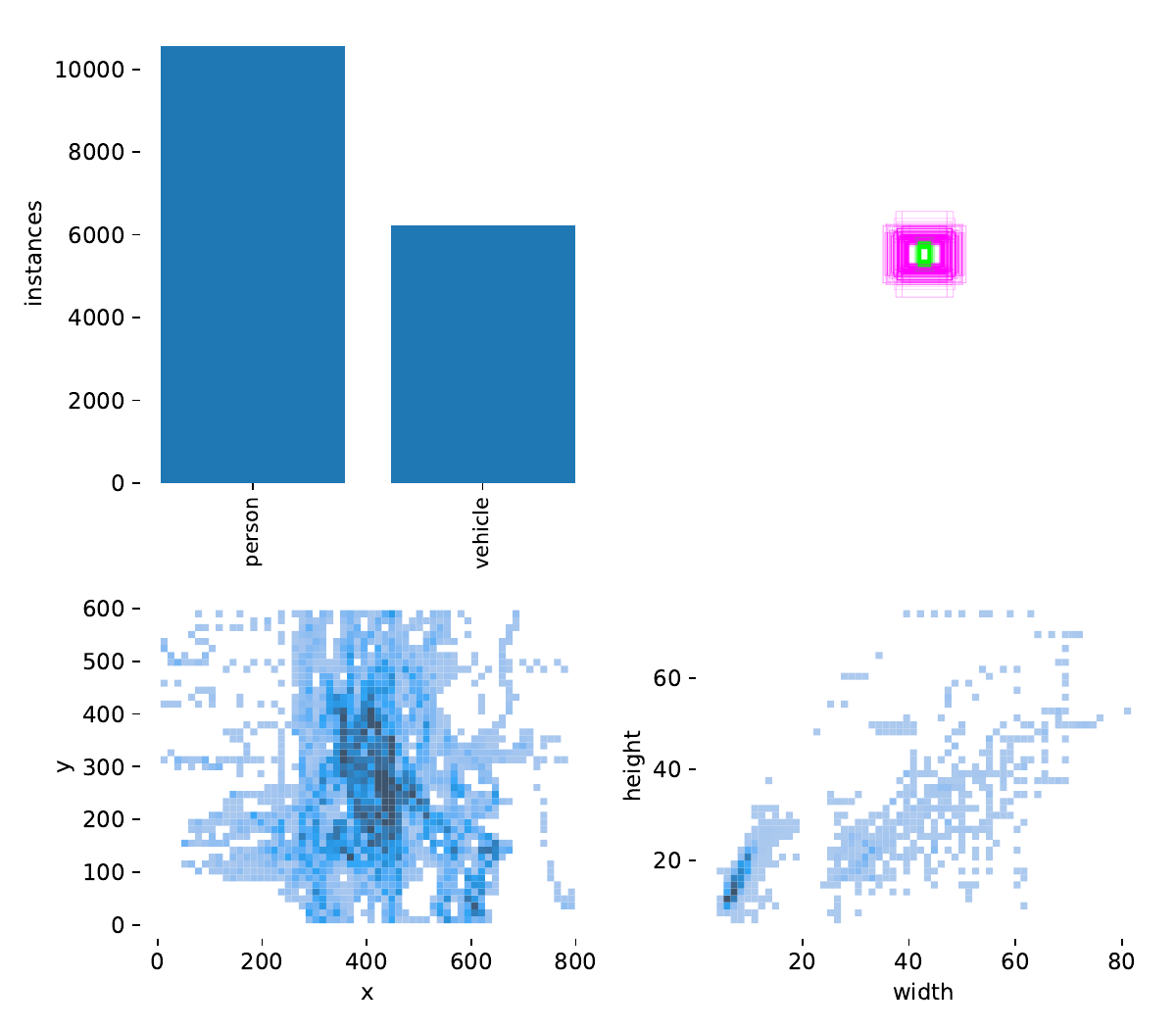}
    \caption{Bounding box ground-truth statistics of runway test split.
    The top-left figure shows the histogram of bounding box instances.
    The top-right figure shows 1K examples of randomly sampled bounding boxes.
    The bottom-left figure shows the distribution of bounding box locations over the image plane.
    The bottom-right figure shows the distribution of bounding box sizes.}
    \label{fig:stats_test_runway}
\end{figure}

Figs.~\ref{fig:stats_train_dirtroad}, 
\ref{fig:stats_val_dirtroad},
and \ref{fig:stats_test_dirtroad},
show the statistics of train, validation, and test splits of dirt-road, respectively,
while
Figs.~\ref{fig:stats_train_runway},
\ref{fig:stats_val_runway},
and \ref{fig:stats_test_runway},
show the statistics of train, validation, and test splits of runway, respectively.
The statistics include i) the histogram of the bounding box instances, ii) examples of bounding boxes randomly sampled from the ground truth, iii) the distribution of the bounding box locations over the image plane, and iv) the distribution of the bounding box sizes as a function of the width and height.
These figures are generated with the software provided with YOLOv5~\cite{Jocher_YOLOv5_by_Ultralytics_2020} and ObjectBox~\cite{Zand2022}.
It is interesting to observe the difference in bounding box sizes between dirt-road and runway splits.

%%%%%%%%%%%%%%%%%%%%%%%%%%%%%%%%%%%%%%%%%%%%%%%%%%%%%%%%%%%%%%%%%
%%%%%%%%%%%%%%%%%%%%%%%%%%%%%%%%%%%%%%%%%%%%%%%%%%%%%%%%%%%%%%%%%
%%%%%%%%%%%%%%%%%%%%%%%%%%%%%%%%%%%%%%%%%%%%%%%%%%%%%%%%%%%%%%%%%
\section{Additional qualitative results}\label{sec:additional_qualitative_results}

Because the detectors are calibrated differently, it is unfair to apply the same confidence threshold to visualise the results.
Therefore, we choose a different threshold for each detector that corresponds to the maximum between $\gamma$ and 0.10, where $\gamma$ is the confidence value of the true positive detection with lowest confidence value in a given frame.
This approach allows us to visualise all detected targets, but it may result in more false alarms.

\begin{figure*}[t]
\begin{center}
  \begin{tabular}{@{}c@{\,}c@{\,}c@{\,}c}
    \begin{overpic}[width=.24\linewidth]{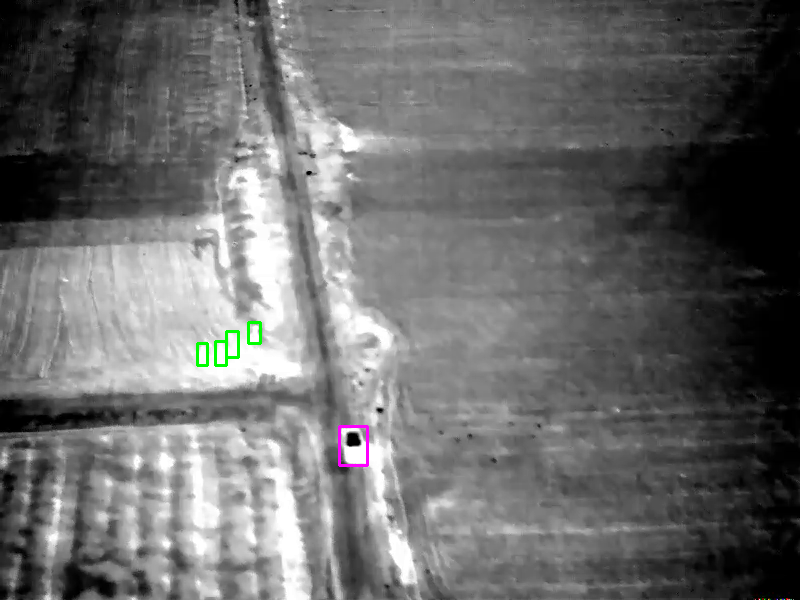}
    \put(2,3){\color{lime}\footnotesize \textbf{ground truth}}
    % \put(85,3){\color{lime}\footnotesize \textbf{0.10}}
    \end{overpic} &
    \begin{overpic}[width=.24\linewidth]{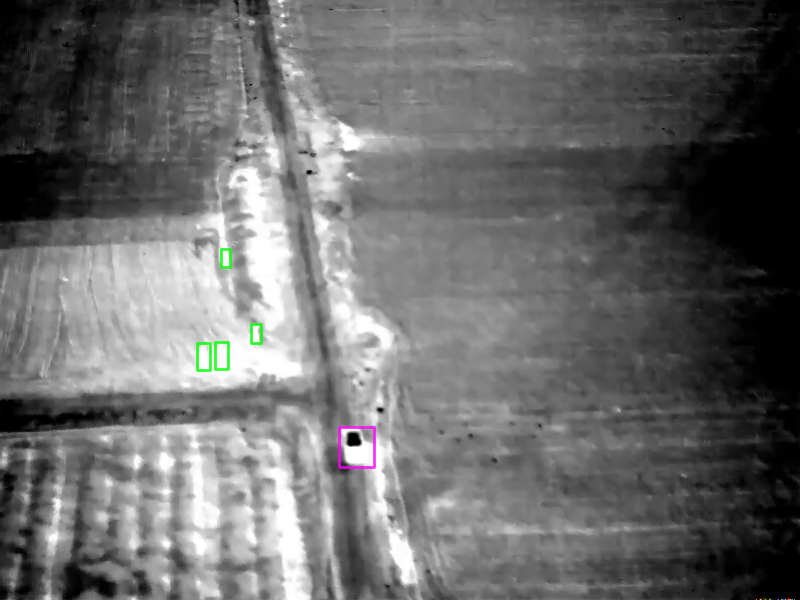}
    \put(2,3){\color{lime}\footnotesize \textbf{Faster R-CNN}}
    \put(85,3){\color{lime}\footnotesize \textbf{0.10}}
    \end{overpic} &
    \begin{overpic}[width=.24\linewidth]{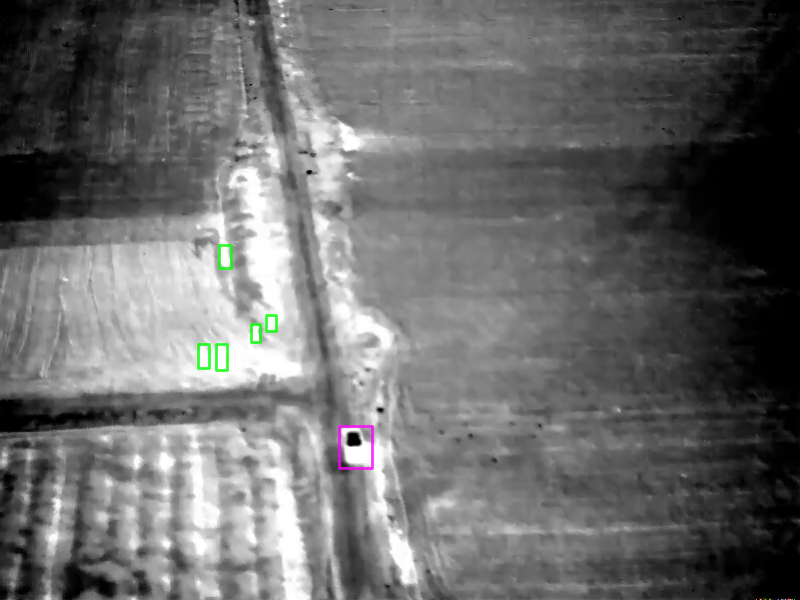}
    \put(2,3){\color{lime}\footnotesize \textbf{SSD}}
    \put(85,3){\color{lime}\footnotesize \textbf{0.10}}
    \end{overpic} &
    \begin{overpic}[width=.24\linewidth]{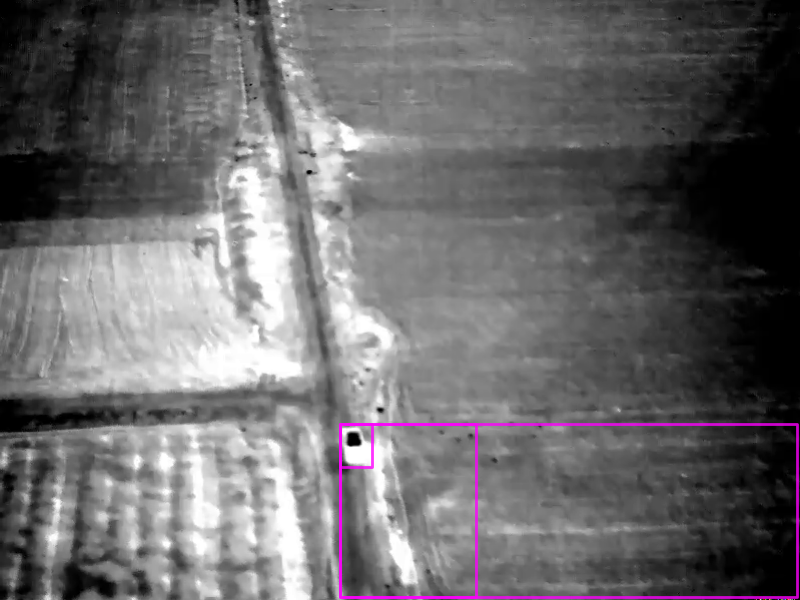}
    \put(2,3){\color{lime}\footnotesize \textbf{CornerNet}}
    \put(85,3){\color{lime}\footnotesize \textbf{0.10}}
    \end{overpic}\\
    \begin{overpic}[width=.24\linewidth]{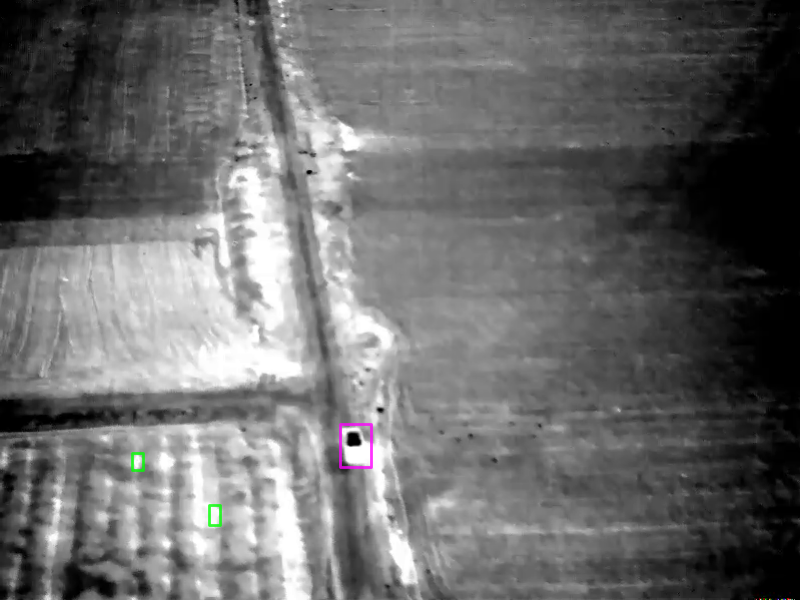}
    \put(2,3){\color{lime}\footnotesize \textbf{FCOS}}
    \put(85,3){\color{lime}\footnotesize \textbf{0.10}}
    \end{overpic} &
    \begin{overpic}[width=.24\linewidth]{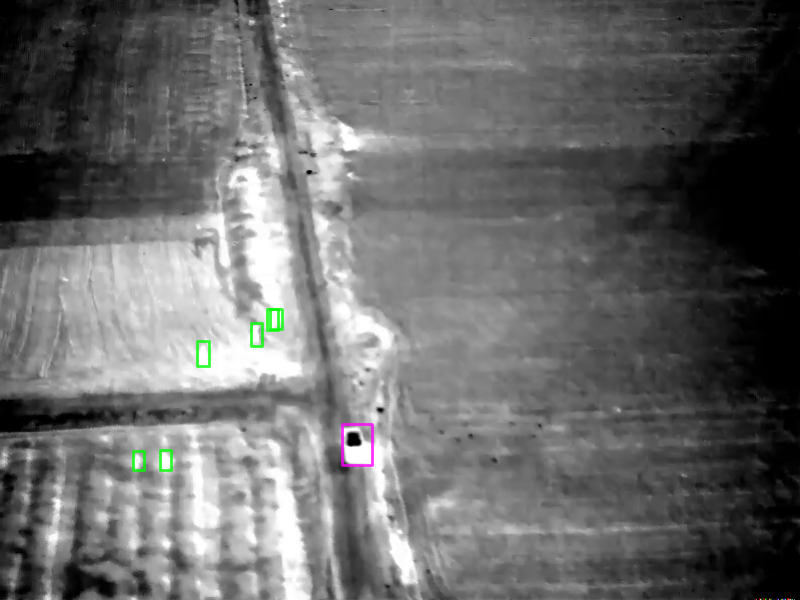}
    \put(2,3){\color{lime}\footnotesize \textbf{DETR}}
    \put(85,3){\color{lime}\footnotesize \textbf{0.10}}
    \end{overpic} &
    \begin{overpic}[width=.24\linewidth]{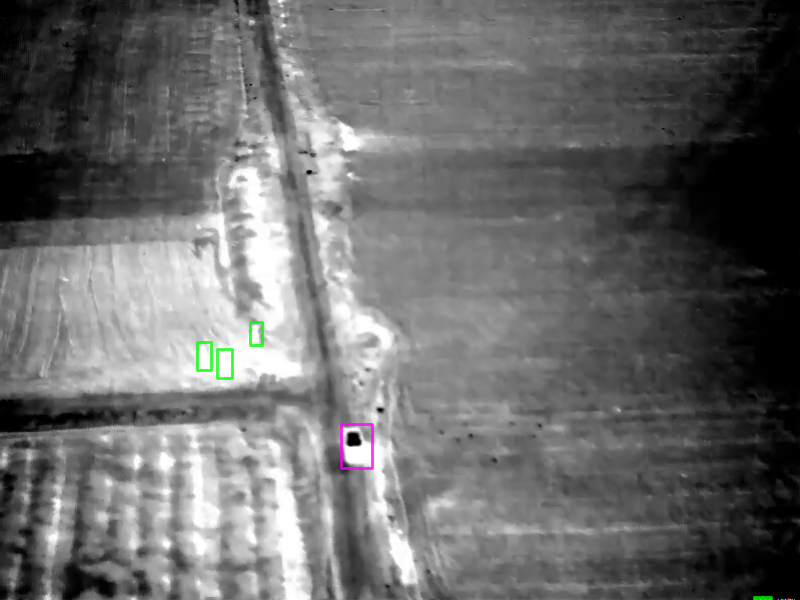}
    \put(2,3){\color{lime}\footnotesize \textbf{Deformable DETR}}
    \put(85,3){\color{lime}\footnotesize \textbf{0.10}}
    \end{overpic} &
    \begin{overpic}[width=.24\linewidth]{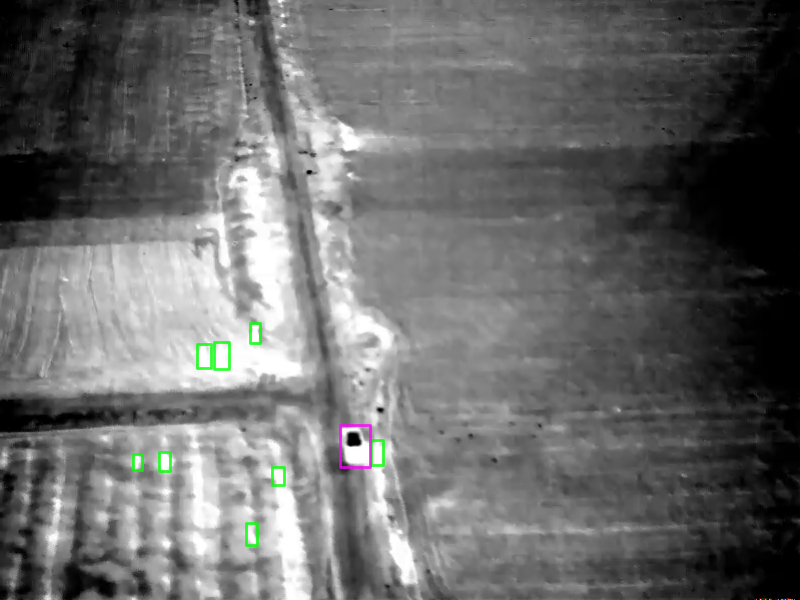}
    \put(2,3){\color{lime}\footnotesize \textbf{VarifocalNet}}
    \put(85,3){\color{lime}\footnotesize \textbf{0.10}}
    \end{overpic}\\
    \begin{overpic}[width=.24\linewidth]{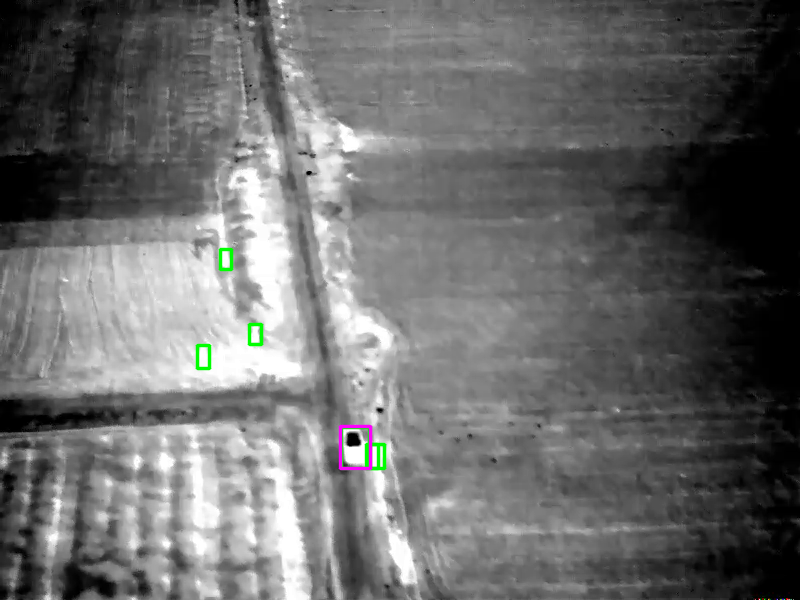}
    \put(2,3){\color{lime}\footnotesize \textbf{ObjectBox}}
    \put(85,3){\color{lime}\footnotesize \textbf{0.10}}
    \end{overpic} &
    \begin{overpic}[width=.24\linewidth]{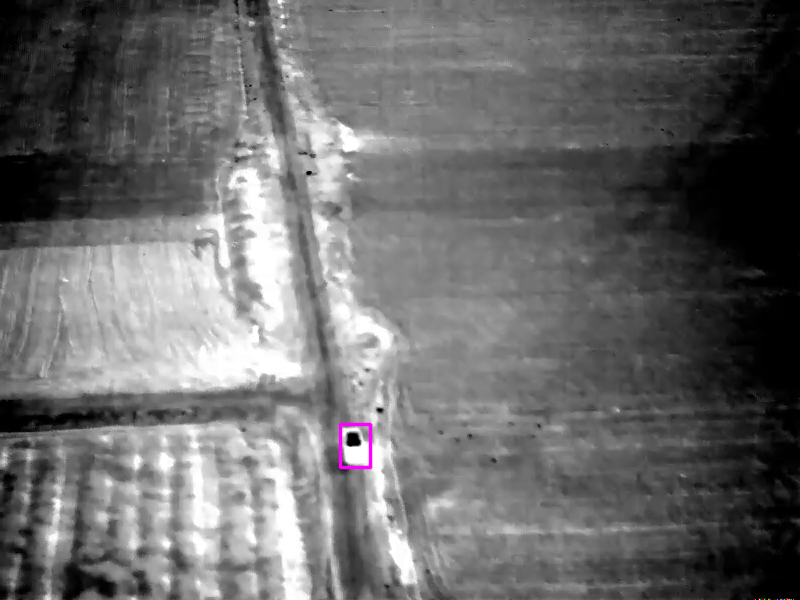}
    \put(2,3){\color{lime}\footnotesize \textbf{YOLOv8}}
    \put(85,3){\color{lime}\footnotesize \textbf{0.10}}
    \end{overpic} &
    \begin{overpic}[width=.24\linewidth]{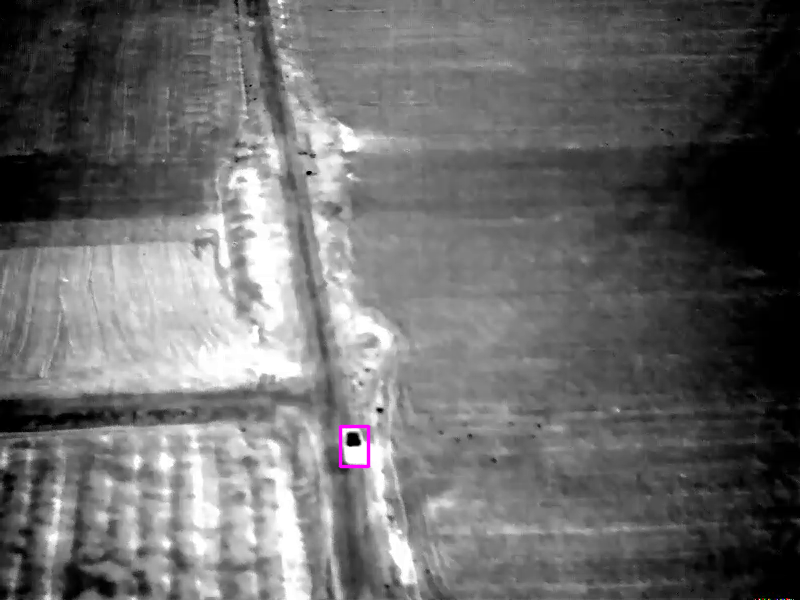}
    \put(2,3){\color{lime}\footnotesize \textbf{ObjectBox\textsuperscript{\textdagger}}}
    \put(85,3){\color{lime}\footnotesize \textbf{0.10}}
    \end{overpic} &
    \begin{overpic}[width=.24\linewidth]{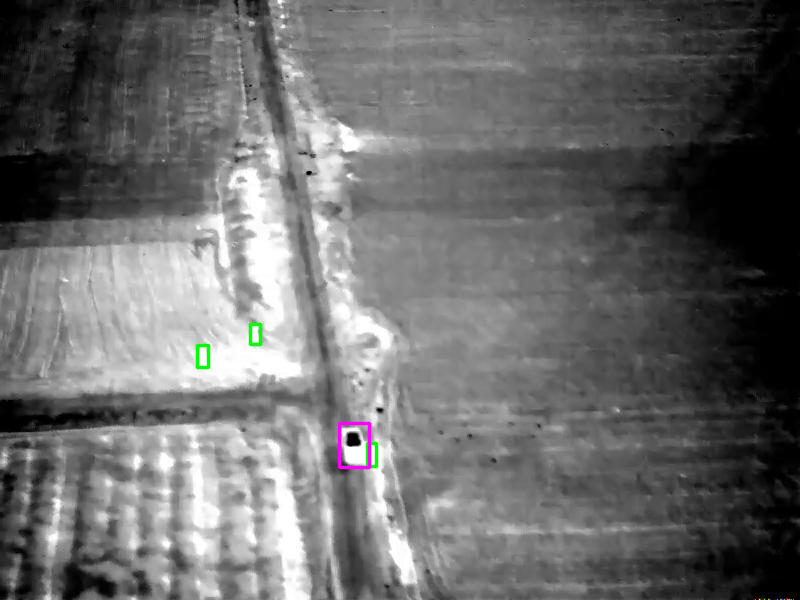}
    \put(2,3){\color{lime}\footnotesize \textbf{YOLOv8\textsuperscript{\textdagger}}}
    \put(85,3){\color{lime}\footnotesize \textbf{0.10}}
    \end{overpic}\\
  \end{tabular}
\end{center}
\vspace{-.6cm}
\caption{Qualitative results of the different detectors trained on dirt-road and evaluated on dirt-road.
Bottom right value in each image represents the conference threshold adopted at inference time.
Bounding boxes: green for \texttt{person}, magenta for \texttt{vehicle}.
Recording altitude: 80m.}
\label{fig:qualitative_results_d2d}
\end{figure*}

Fig.~\ref{fig:qualitative_results_d2d} shows the qualitative results of the different detectors when their models are trained on dirt-road and evaluated on dirt-road.
We can observe that this scenario is very challenging because all the detectors fail to detect all the person targets.
The vehicle is accurately detected by all the detectors except for CornerNet.

\begin{figure*}[t]
\begin{center}
  \begin{tabular}{@{}c@{\,}c@{\,}c@{\,}c}
    \begin{overpic}[width=.24\linewidth]{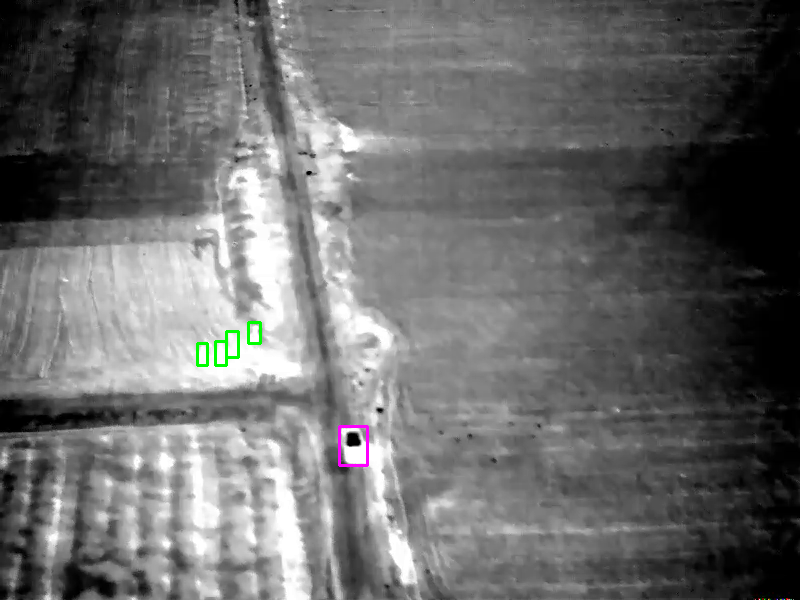}
    \put(2,3){\color{lime}\footnotesize \textbf{ground truth}}
    \put(85,3){\color{lime}\footnotesize \textbf{0.10}}
    \end{overpic} &
    \begin{overpic}[width=.24\linewidth]{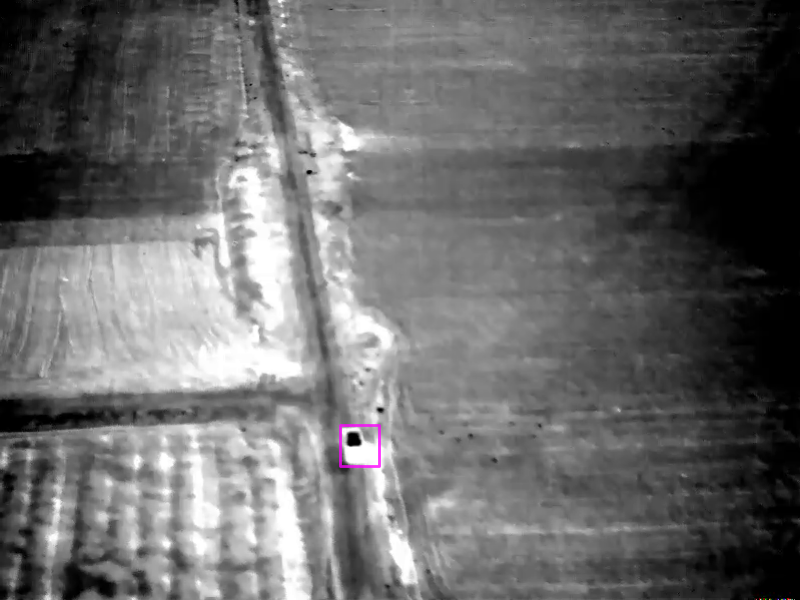}
    \put(2,3){\color{lime}\footnotesize \textbf{Faster R-CNN}}
    \put(85,3){\color{lime}\footnotesize \textbf{0.10}}
    \end{overpic} &
    \begin{overpic}[width=.24\linewidth]{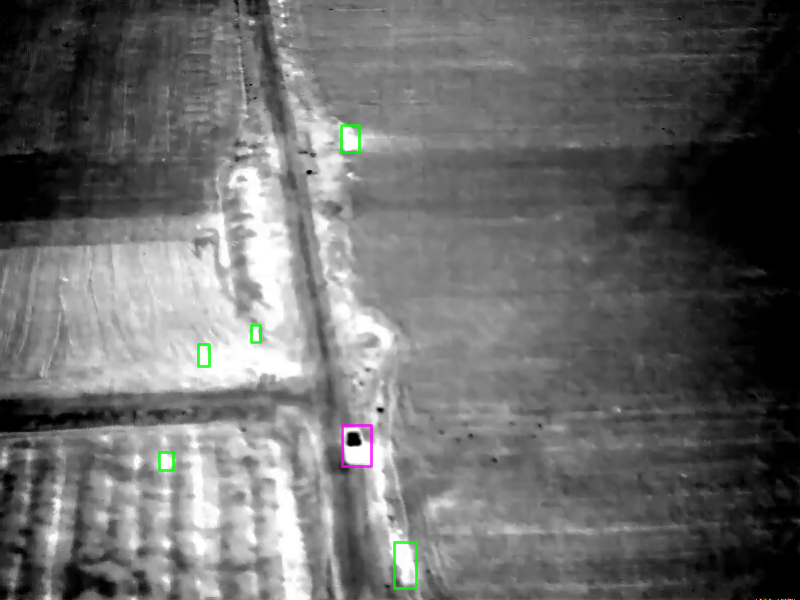}
    \put(2,3){\color{lime}\footnotesize \textbf{SSD}}
    \put(85,3){\color{lime}\footnotesize \textbf{0.10}}
    \end{overpic} &
    \begin{overpic}[width=.24\linewidth]{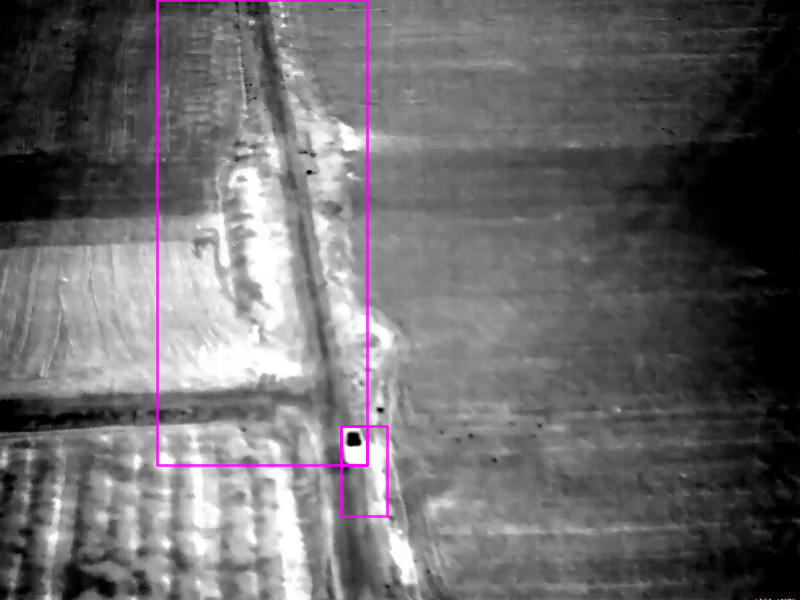}
    \put(2,3){\color{lime}\footnotesize \textbf{CornerNet}}
    \put(85,3){\color{lime}\footnotesize \textbf{0.10}}
    \end{overpic}\\
    \begin{overpic}[width=.24\linewidth]{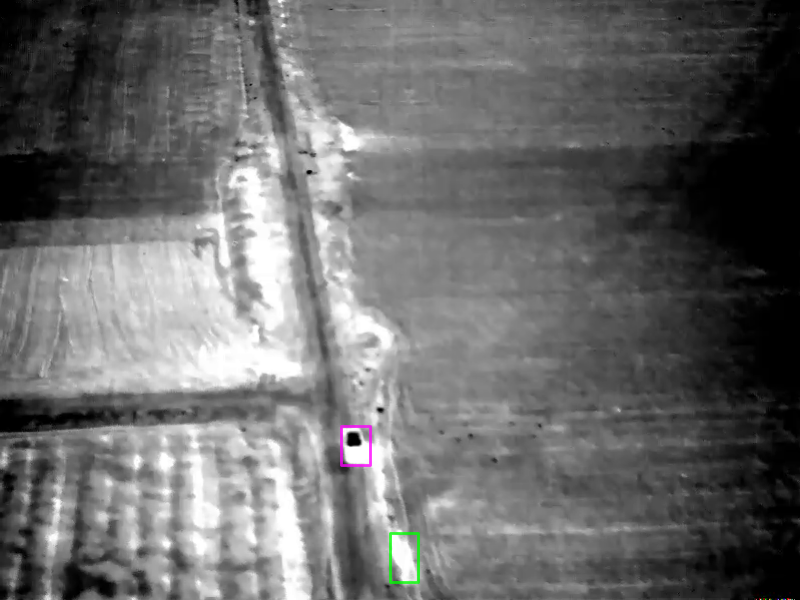}
    \put(2,3){\color{lime}\footnotesize \textbf{FCOS}}
    \put(85,3){\color{lime}\footnotesize \textbf{0.10}}
    \end{overpic} &
    \begin{overpic}[width=.24\linewidth]{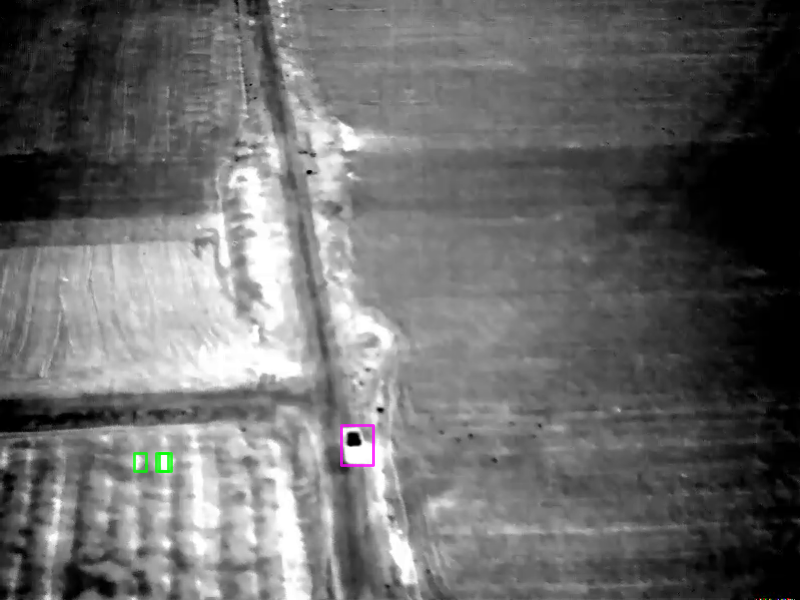}
    \put(2,3){\color{lime}\footnotesize \textbf{DETR}}
    \put(85,3){\color{lime}\footnotesize \textbf{0.10}}
    \end{overpic} &
    \begin{overpic}[width=.24\linewidth]{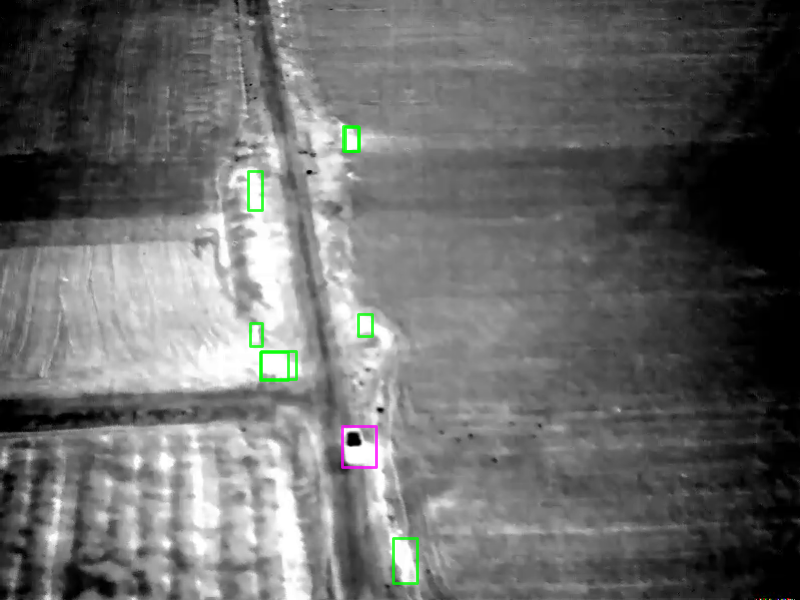}
    \put(2,3){\color{lime}\footnotesize \textbf{Deformable DETR}}
    \put(85,3){\color{lime}\footnotesize \textbf{0.10}}
    \end{overpic} &
    \begin{overpic}[width=.24\linewidth]{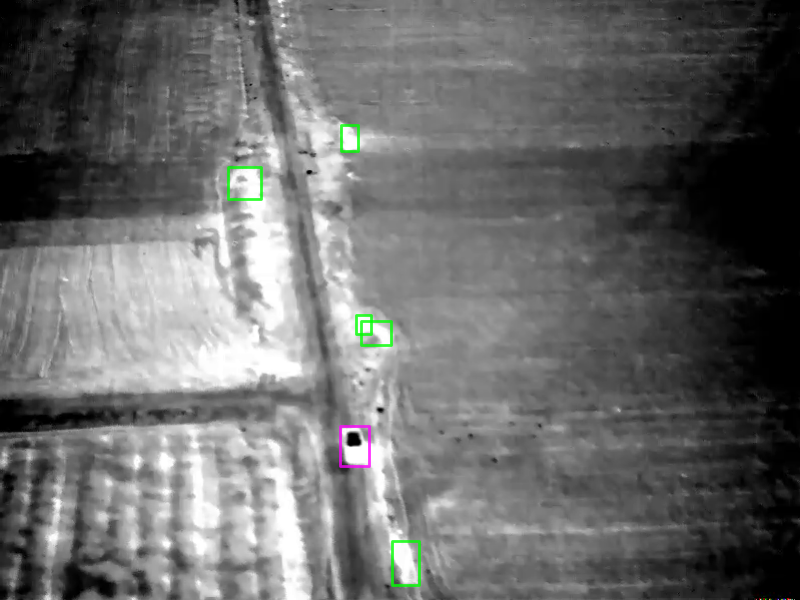}
    \put(2,3){\color{lime}\footnotesize \textbf{VarifocalNet}}
    \put(85,3){\color{lime}\footnotesize \textbf{0.10}}
    \end{overpic}\\
    \begin{overpic}[width=.24\linewidth]{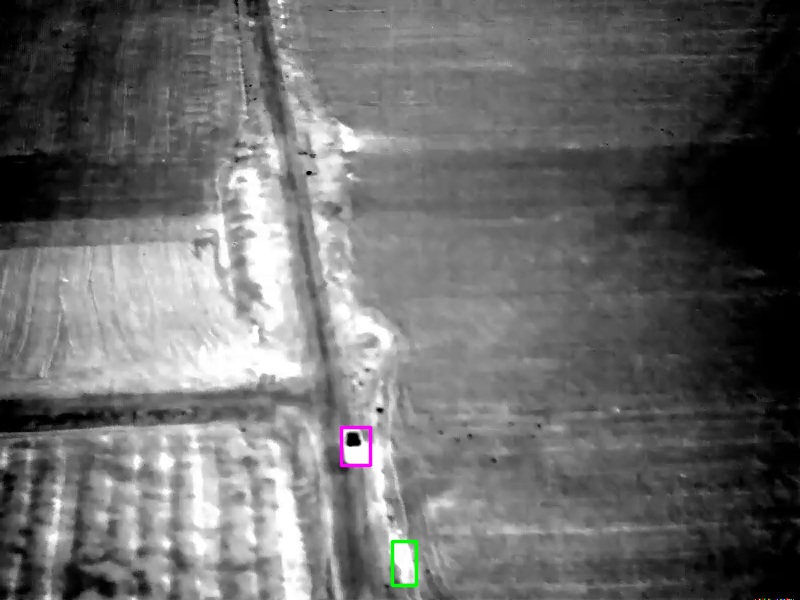}
    \put(2,3){\color{lime}\footnotesize \textbf{ObjectBox}}
    \put(85,3){\color{lime}\footnotesize \textbf{0.10}}
    \end{overpic} &
    \begin{overpic}[width=.24\linewidth]{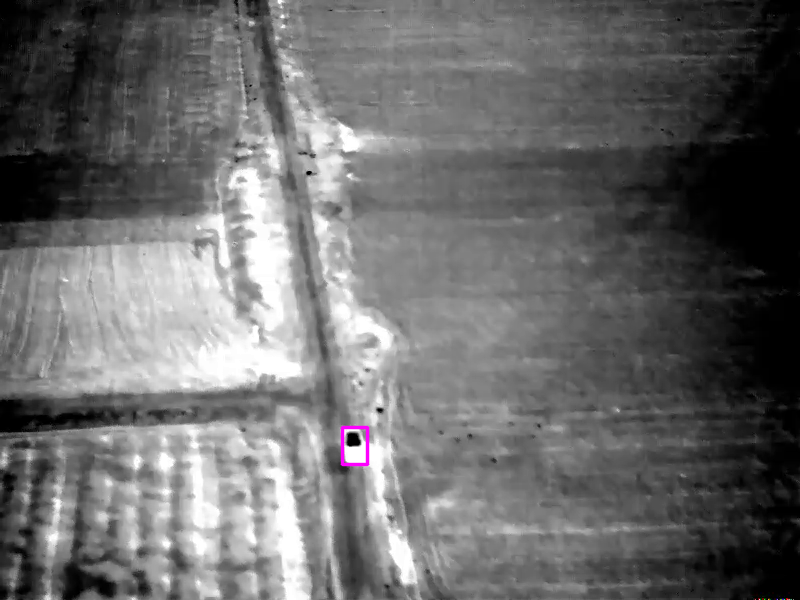}
    \put(2,3){\color{lime}\footnotesize \textbf{YOLOv8}}
    \put(85,3){\color{lime}\footnotesize \textbf{0.10}}
    \end{overpic} &
    \begin{overpic}[width=.24\linewidth]{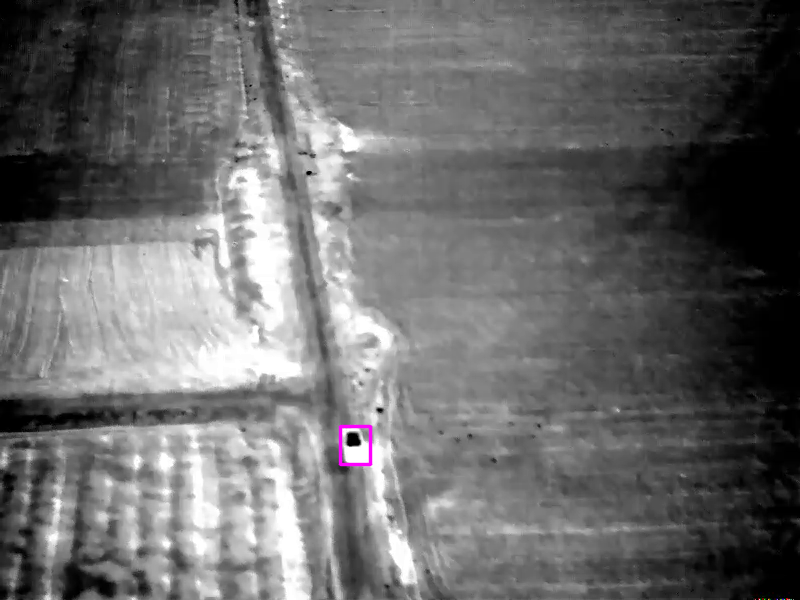}
    \put(2,3){\color{lime}\footnotesize \textbf{ObjectBox\textsuperscript{\textdagger}}}
    \put(85,3){\color{lime}\footnotesize \textbf{0.10}}
    \end{overpic} &
    \begin{overpic}[width=.24\linewidth]{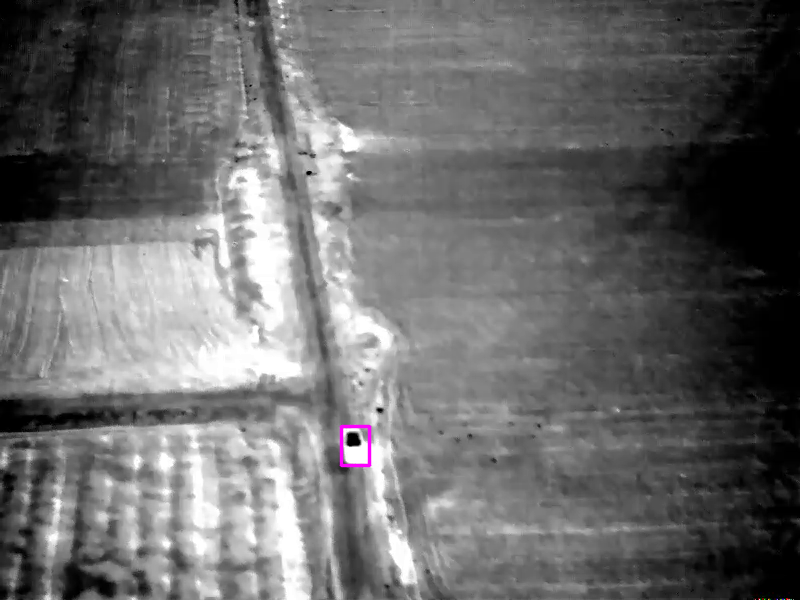}
    \put(2,3){\color{lime}\footnotesize \textbf{YOLOv8\textsuperscript{\textdagger}}}
    \put(85,3){\color{lime}\footnotesize \textbf{0.10}}
    \end{overpic}\\
  \end{tabular}
\end{center}
\vspace{-.6cm}
\caption{Qualitative results of the different detectors trained on runway and evaluated on dirt-road.
Bottom right value in each image represents the conference threshold adopted at inference time.
Bounding boxes: green for \texttt{person}, magenta for \texttt{vehicle}.
Recording altitude: 80m.}
\label{fig:qualitative_results_r2d}
\end{figure*}

Fig.~\ref{fig:qualitative_results_r2d} shows the qualitative results of the different detectors when their models are trained on runway and evaluated on dirt-road.
We can observe that SSD is the only detector that can detect some person targets. 
All the others either produce false alarms or do not detect any person targets.
Like before, the vehicle is accurately detected by all the detectors except for CornerNet.

\begin{figure*}[t]
\begin{center}
  \begin{tabular}{@{}c@{\,}c@{\,}c@{\,}c}
    \begin{overpic}[width=.24\linewidth]{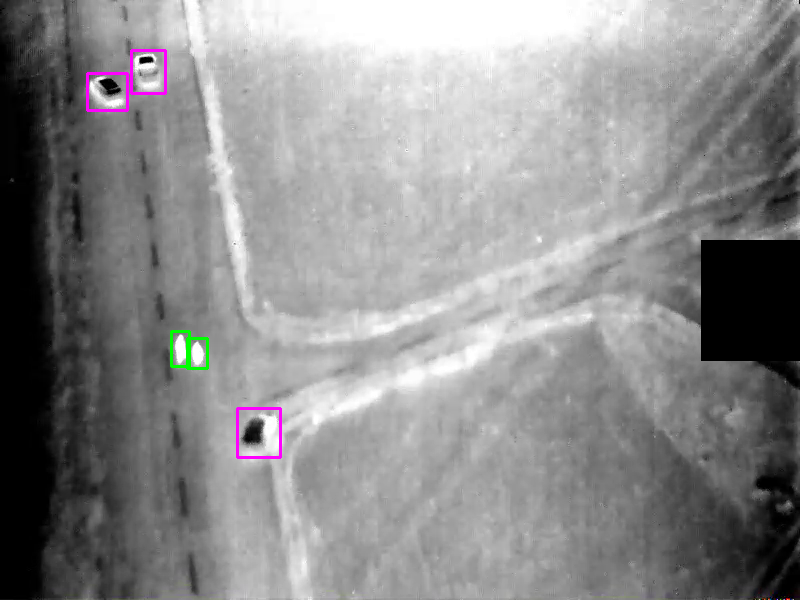}
    \put(2,3){\color{lime}\footnotesize \textbf{ground truth}}
    % \put(85,3){\color{lime}\footnotesize \textbf{0.10}}
    \end{overpic} &
    \begin{overpic}[width=.24\linewidth]{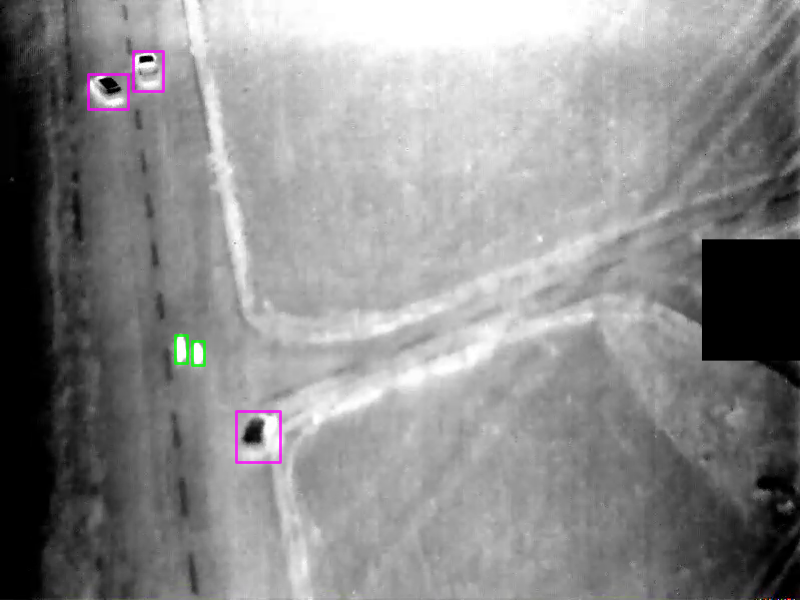}
    \put(2,3){\color{lime}\footnotesize \textbf{Faster R-CNN}}
    \put(85,3){\color{lime}\footnotesize \textbf{0.99}}
    \end{overpic} &
    \begin{overpic}[width=.24\linewidth]{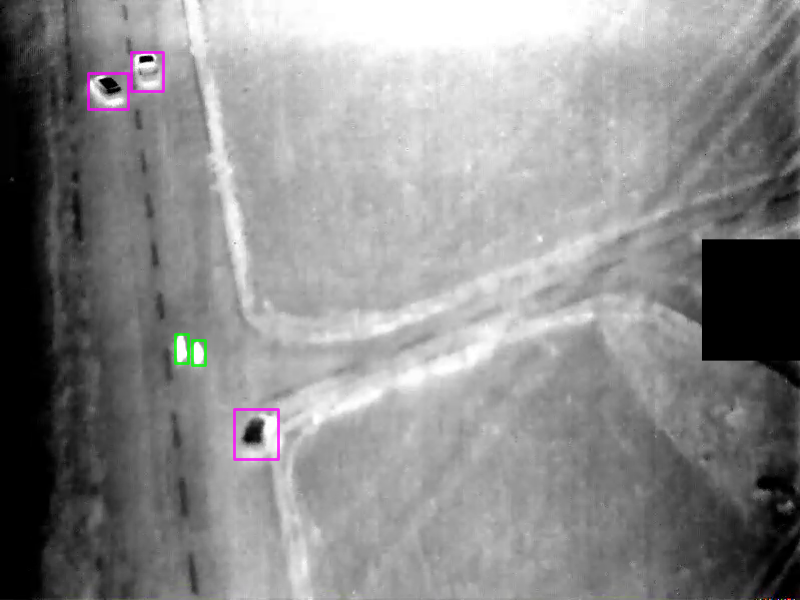}
    \put(2,3){\color{lime}\footnotesize \textbf{SSD}}
    \put(85,3){\color{lime}\footnotesize \textbf{0.96}}
    \end{overpic} &
    \begin{overpic}[width=.24\linewidth]{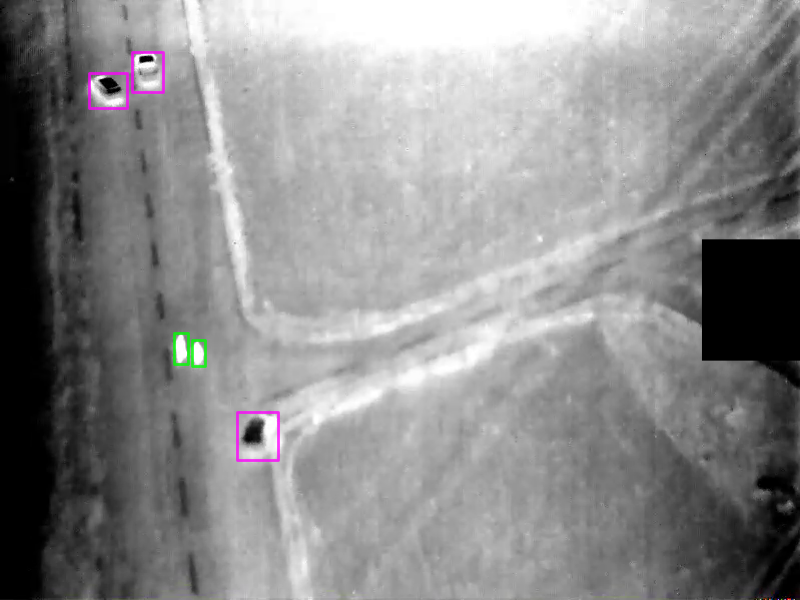}
    \put(2,3){\color{lime}\footnotesize \textbf{CornerNet}}
    \put(85,3){\color{lime}\footnotesize \textbf{0.56}}
    \end{overpic}\\
    \begin{overpic}[width=.24\linewidth]{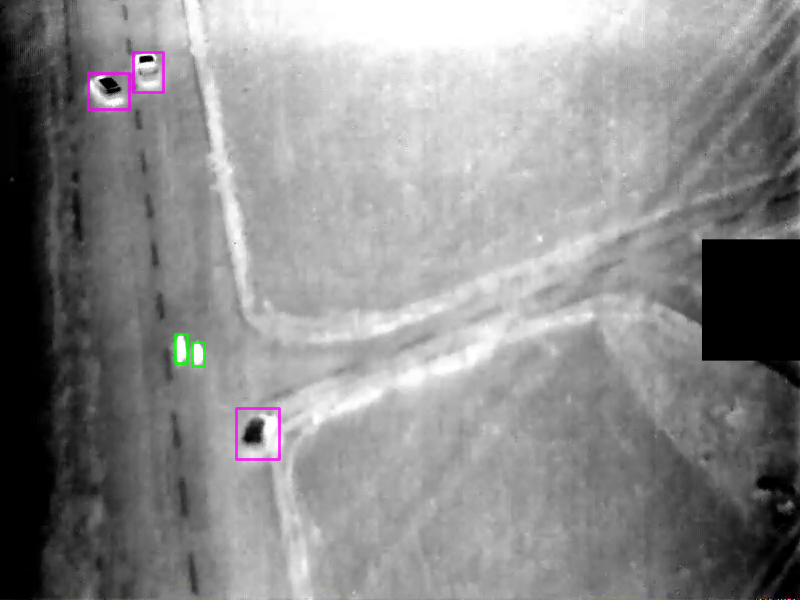}
    \put(2,3){\color{lime}\footnotesize \textbf{FCOS}}
    \put(85,3){\color{lime}\footnotesize \textbf{0.61}}
    \end{overpic} &
    \begin{overpic}[width=.24\linewidth]{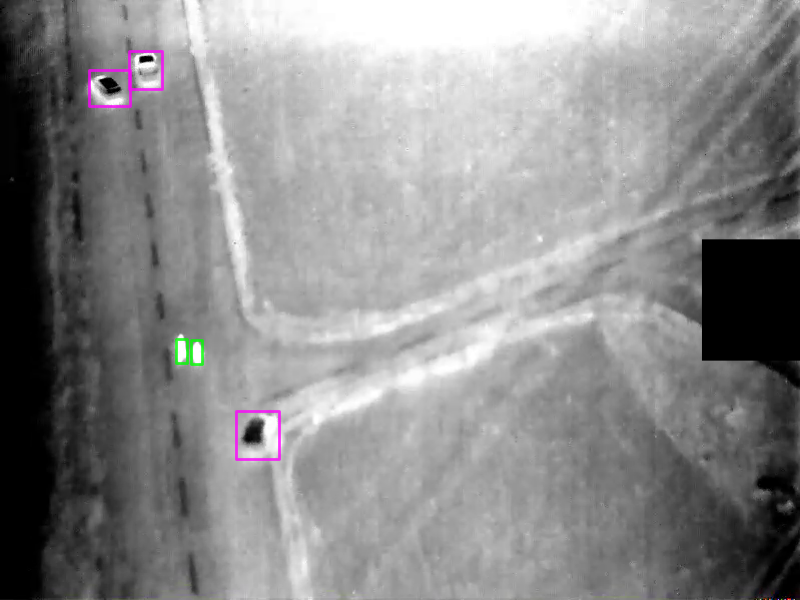}
    \put(2,3){\color{lime}\footnotesize \textbf{DETR}}
    \put(85,3){\color{lime}\footnotesize \textbf{0.97}}
    \end{overpic} &
    \begin{overpic}[width=.24\linewidth]{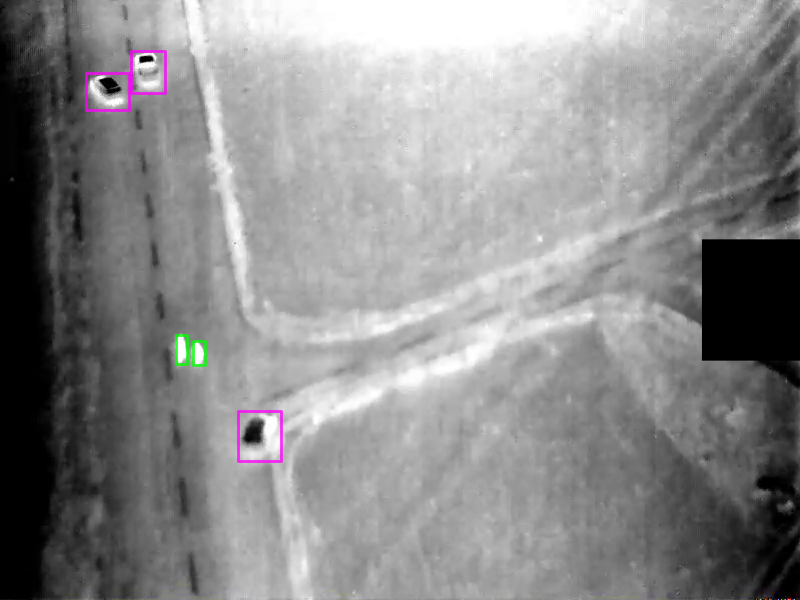}
    \put(2,3){\color{lime}\footnotesize \textbf{Deformable DETR}}
    \put(85,3){\color{lime}\footnotesize \textbf{0.93}}
    \end{overpic} &
    \begin{overpic}[width=.24\linewidth]{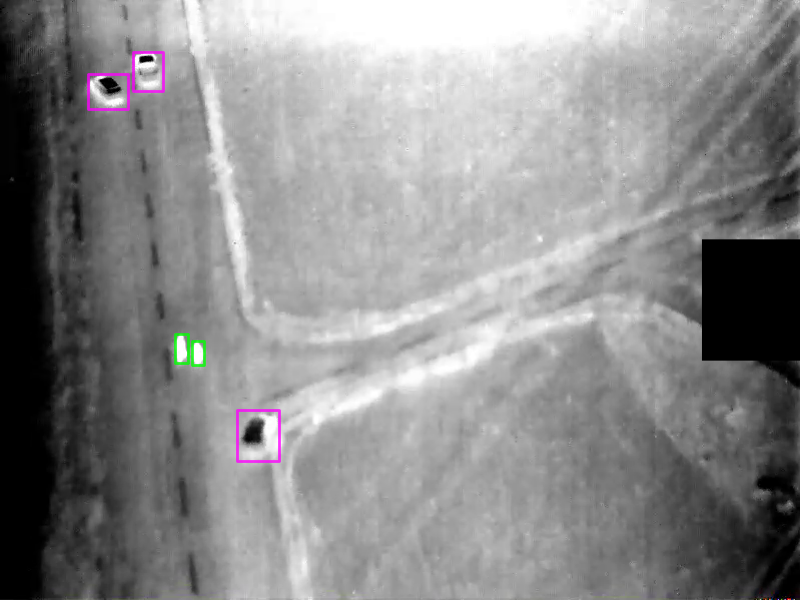}
    \put(2,3){\color{lime}\footnotesize \textbf{VarifocalNet}}
    \put(85,3){\color{lime}\footnotesize \textbf{0.85}}
    \end{overpic}\\
    \begin{overpic}[width=.24\linewidth]{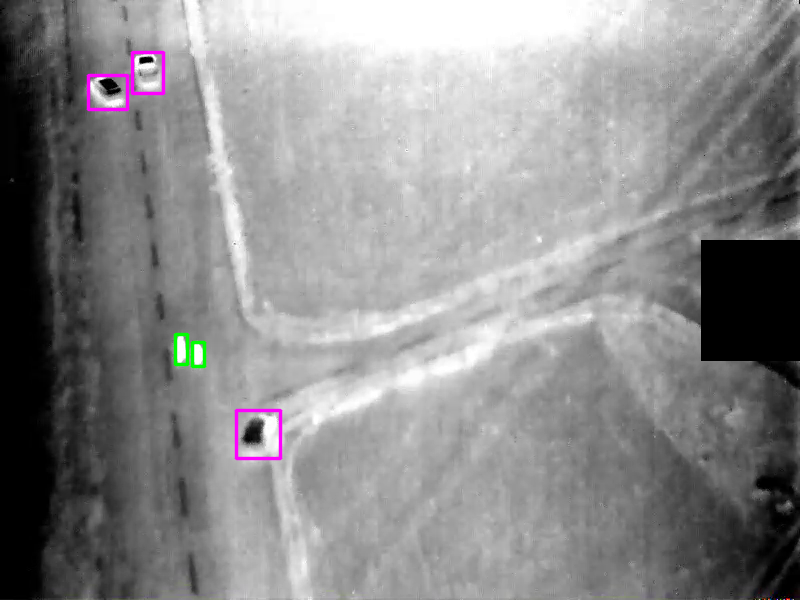}
    \put(2,3){\color{lime}\footnotesize \textbf{ObjectBox}}
    \put(85,3){\color{lime}\footnotesize \textbf{0.88}}
    \end{overpic} &
    \begin{overpic}[width=.24\linewidth]{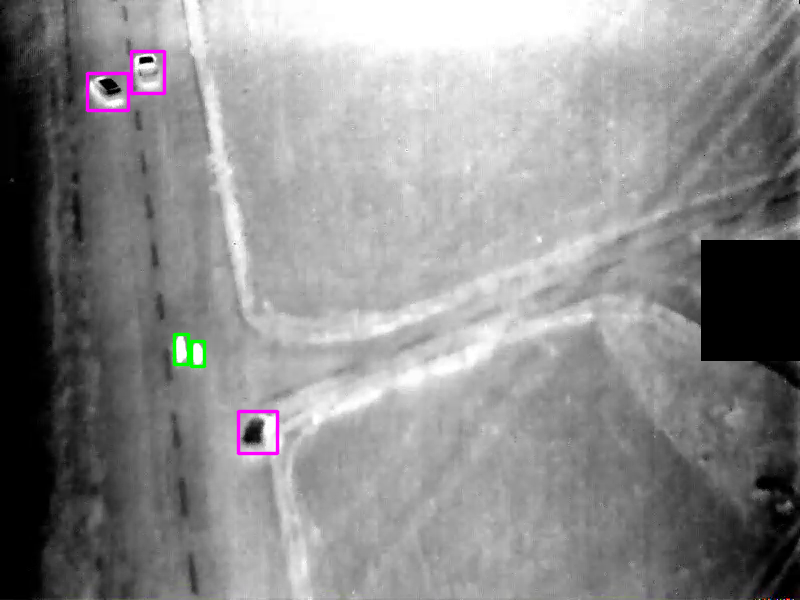}
    \put(2,3){\color{lime}\footnotesize \textbf{YOLOv8}}
    \put(85,3){\color{lime}\footnotesize \textbf{0.78}}
    \end{overpic} &
    \begin{overpic}[width=.24\linewidth]{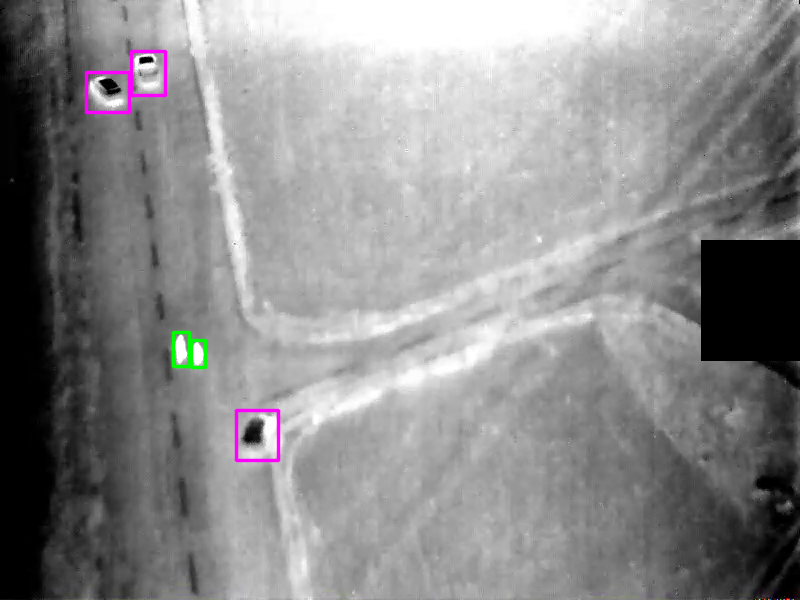}
    \put(2,3){\color{lime}\footnotesize \textbf{ObjectBox\textsuperscript{\textdagger}}}
    \put(85,3){\color{lime}\footnotesize \textbf{0.86}}
    \end{overpic} &
    \begin{overpic}[width=.24\linewidth]{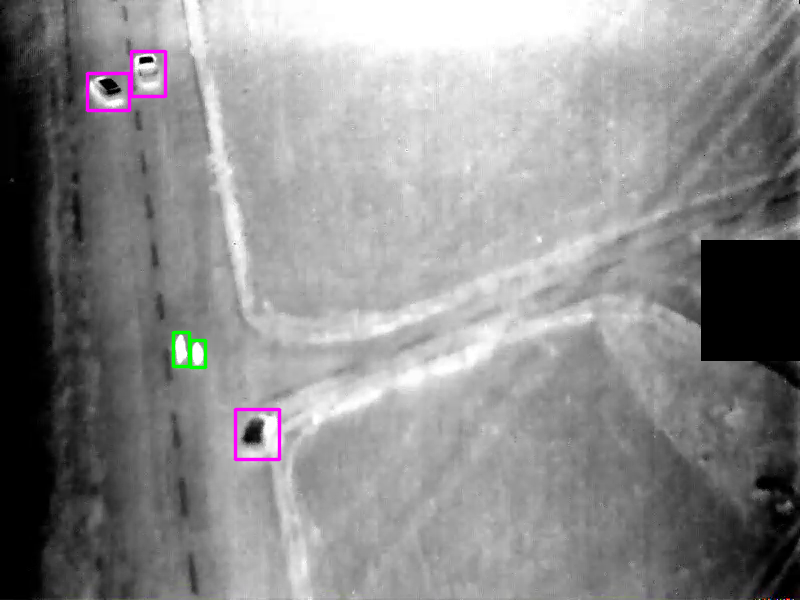}
    \put(2,3){\color{lime}\footnotesize \textbf{YOLOv8\textsuperscript{\textdagger}}}
    \put(85,3){\color{lime}\footnotesize \textbf{0.82}}
    \end{overpic}\\
  \end{tabular}
\end{center}
\vspace{-.6cm}
\caption{Qualitative results of the different detectors trained on runway and evaluated on runway.
Bottom right value in each image represents the conference threshold adopted at inference time.
Bounding boxes: green for \texttt{person}, magenta for \texttt{vehicle}.
Recording altitude: 82m.}
\label{fig:qualitative_results_r2r}
\end{figure*}

Fig.~\ref{fig:qualitative_results_r2r} shows the qualitative results of the different detectors when their models are trained on runway and evaluated on runway.
Unlike before, all the targets are correctly detected in this setting.
Moreover, we can observe that the confidence value of each detector is rather different from each other.

\begin{figure*}[t]
\begin{center}
  \begin{tabular}{@{}c@{\,}c@{\,}c@{\,}c}
    \begin{overpic}[width=.24\linewidth]{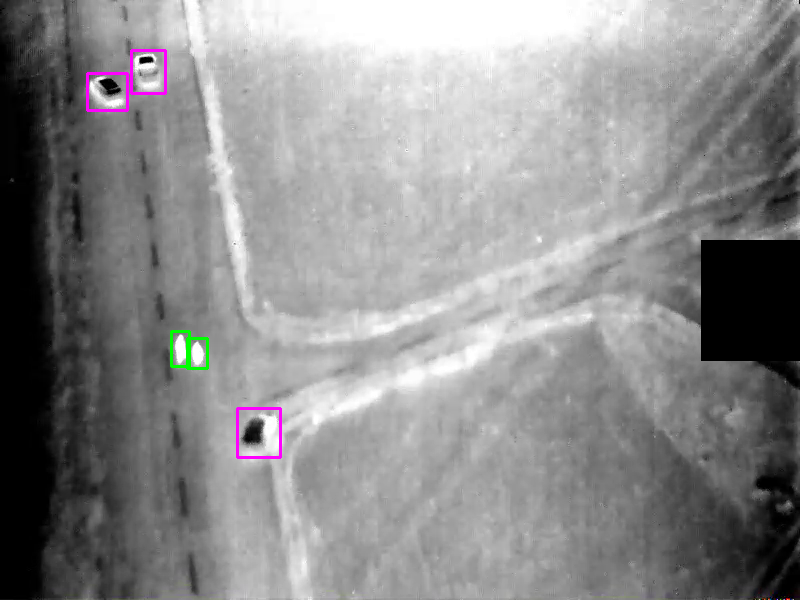}
    \put(2,3){\color{lime}\footnotesize \textbf{ground truth}}
    \put(85,3){\color{lime}\footnotesize \textbf{0.10}}
    \end{overpic} &
    \begin{overpic}[width=.24\linewidth]{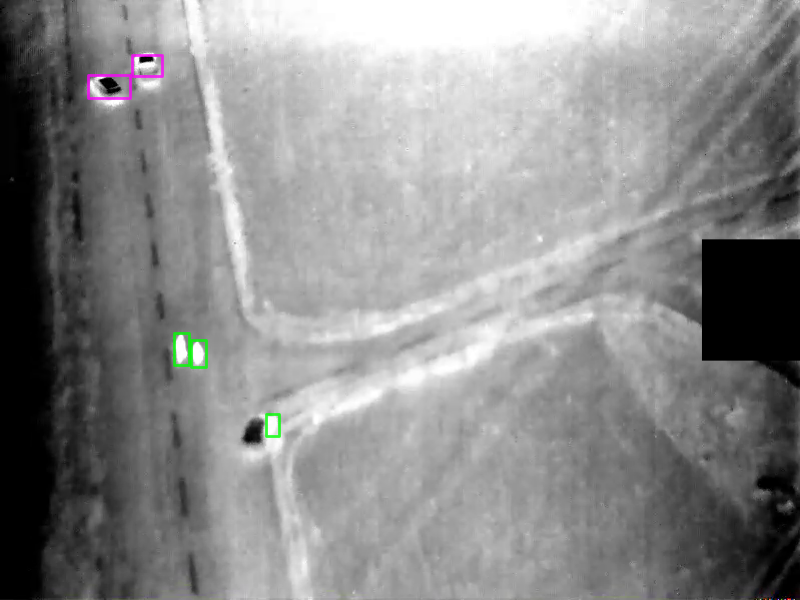}
    \put(2,3){\color{lime}\footnotesize \textbf{Faster R-CNN}}
    \put(85,3){\color{lime}\footnotesize \textbf{0.10}}
    \end{overpic} &
    \begin{overpic}[width=.24\linewidth]{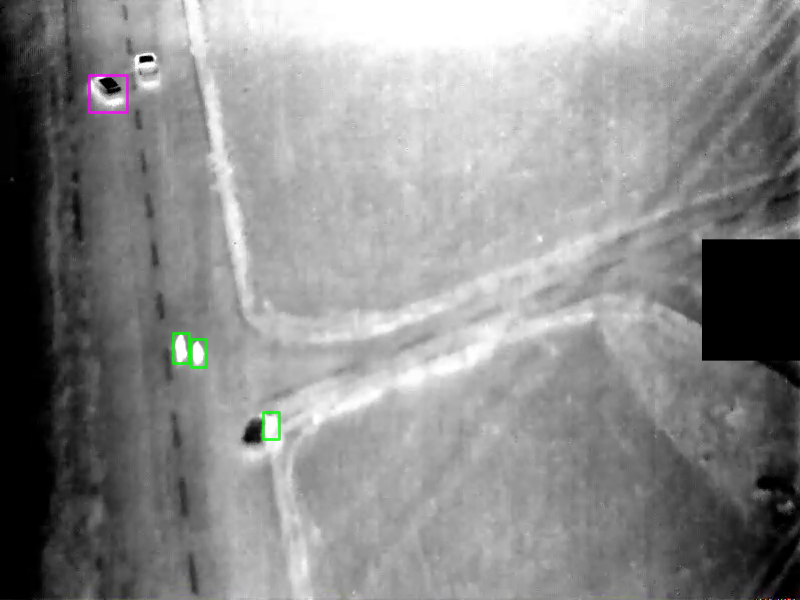}
    \put(2,3){\color{lime}\footnotesize \textbf{SSD}}
    \put(85,3){\color{lime}\footnotesize \textbf{0.10}}
    \end{overpic} &
    \begin{overpic}[width=.24\linewidth]{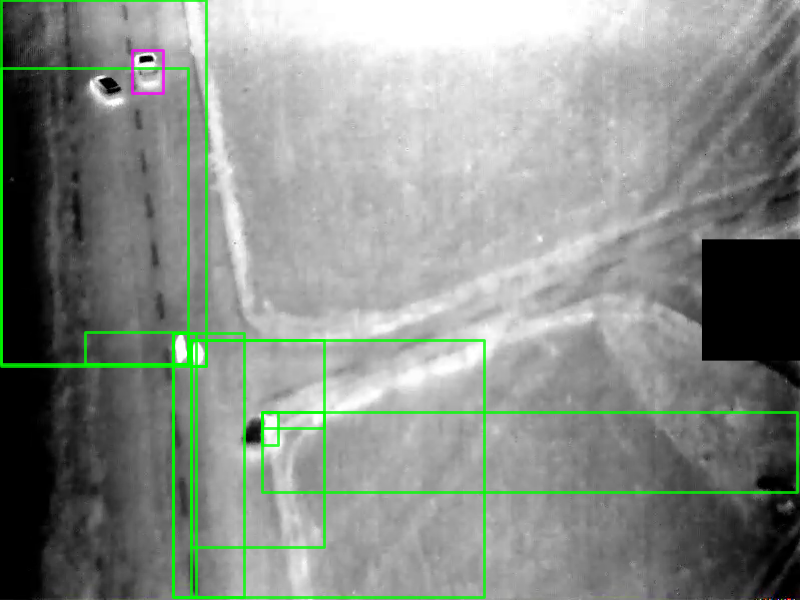}
    \put(2,3){\color{lime}\footnotesize \textbf{CornerNet}}
    \put(85,3){\color{lime}\footnotesize \textbf{0.10}}
    \end{overpic}\\
    \begin{overpic}[width=.24\linewidth]{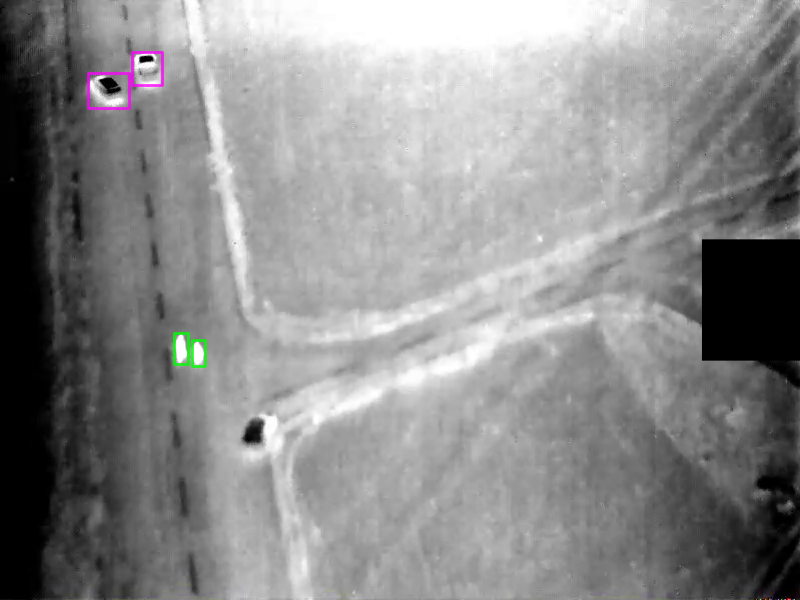}
    \put(2,3){\color{lime}\footnotesize \textbf{FCOS}}
    \put(85,3){\color{lime}\footnotesize \textbf{0.10}}
    \end{overpic} &
    \begin{overpic}[width=.24\linewidth]{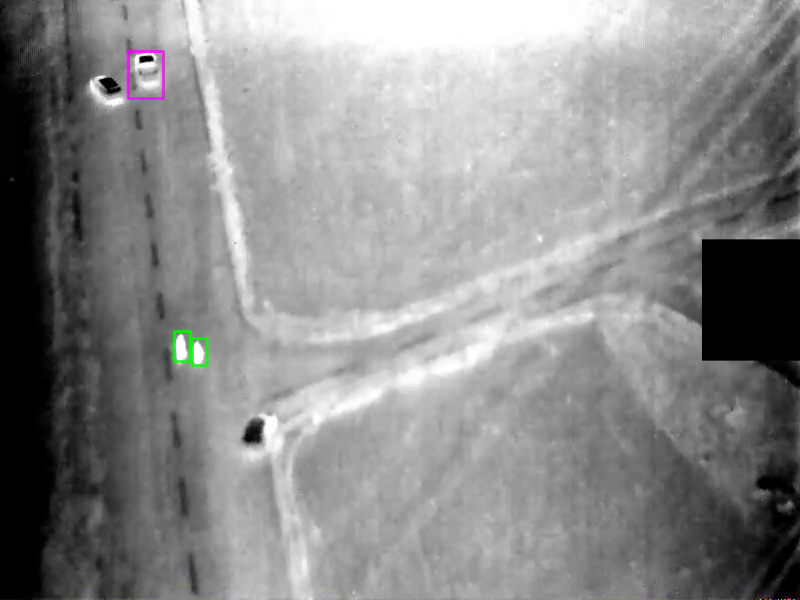}
    \put(2,3){\color{lime}\footnotesize \textbf{DETR}}
    \put(85,3){\color{lime}\footnotesize \textbf{0.10}}
    \end{overpic} &
    \begin{overpic}[width=.24\linewidth]{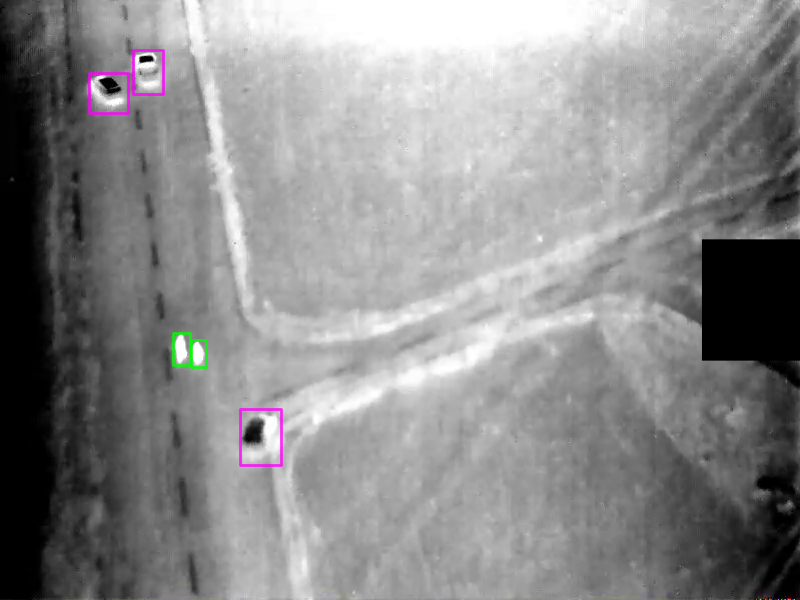}
    \put(2,3){\color{lime}\footnotesize \textbf{Deformable DETR}}
    \put(85,3){\color{lime}\footnotesize \textbf{0.10}}
    \end{overpic} &
    \begin{overpic}[width=.24\linewidth]{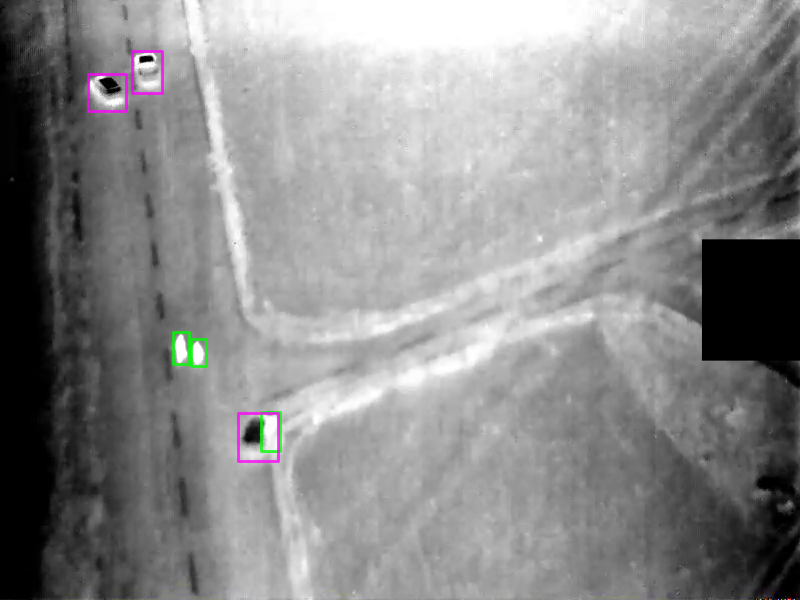}
    \put(2,3){\color{lime}\footnotesize \textbf{VarifocalNet}}
    \put(85,3){\color{lime}\footnotesize \textbf{0.46}}
    \end{overpic}\\
    \begin{overpic}[width=.24\linewidth]{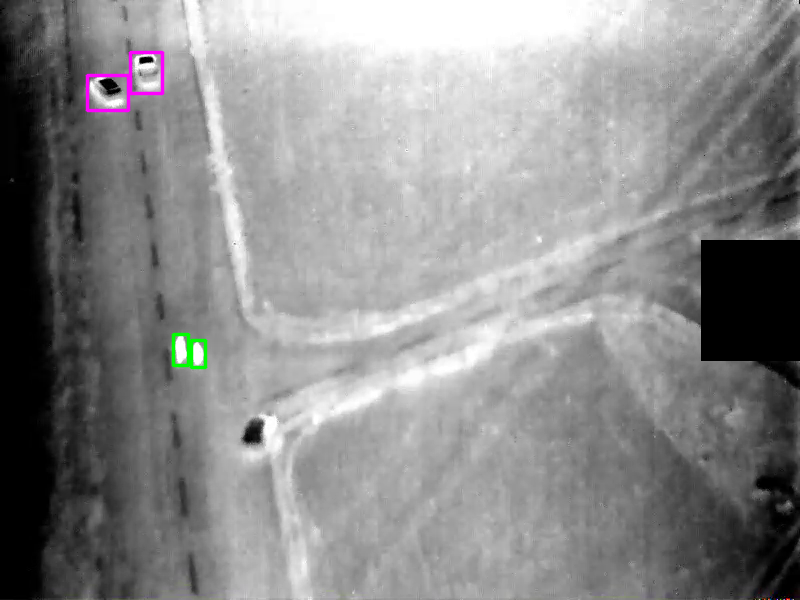}
    \put(2,3){\color{lime}\footnotesize \textbf{ObjectBox}}
    \put(85,3){\color{lime}\footnotesize \textbf{0.10}}
    \end{overpic} &
    \begin{overpic}[width=.24\linewidth]{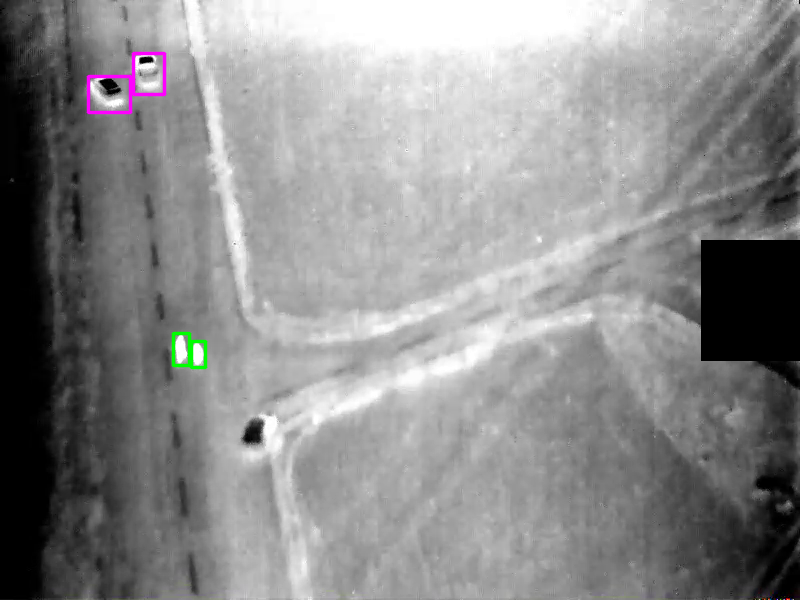}
    \put(2,3){\color{lime}\footnotesize \textbf{YOLOv8}}
    \put(85,3){\color{lime}\footnotesize \textbf{0.10}}
    \end{overpic} &
    \begin{overpic}[width=.24\linewidth]{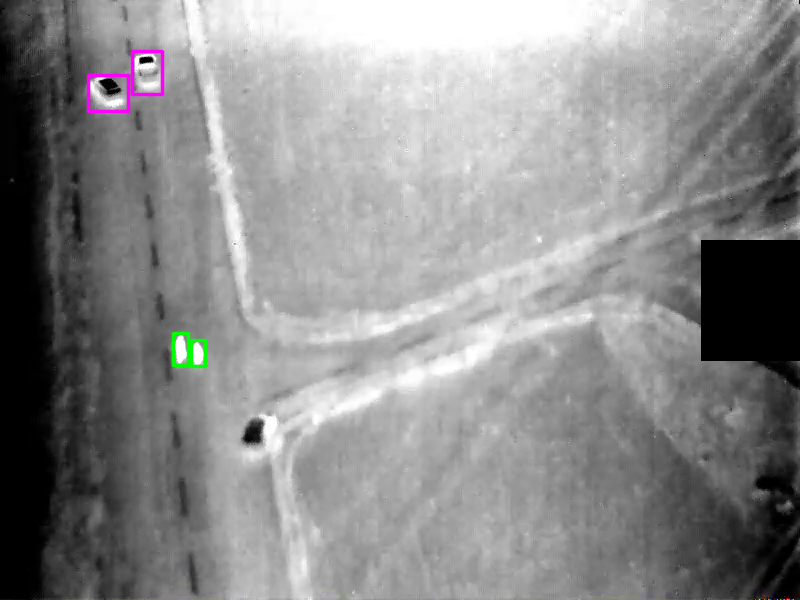}
    \put(2,3){\color{lime}\footnotesize \textbf{ObjectBox\textsuperscript{\textdagger}}}
    \put(85,3){\color{lime}\footnotesize \textbf{0.10}}
    \end{overpic} &
    \begin{overpic}[width=.24\linewidth]{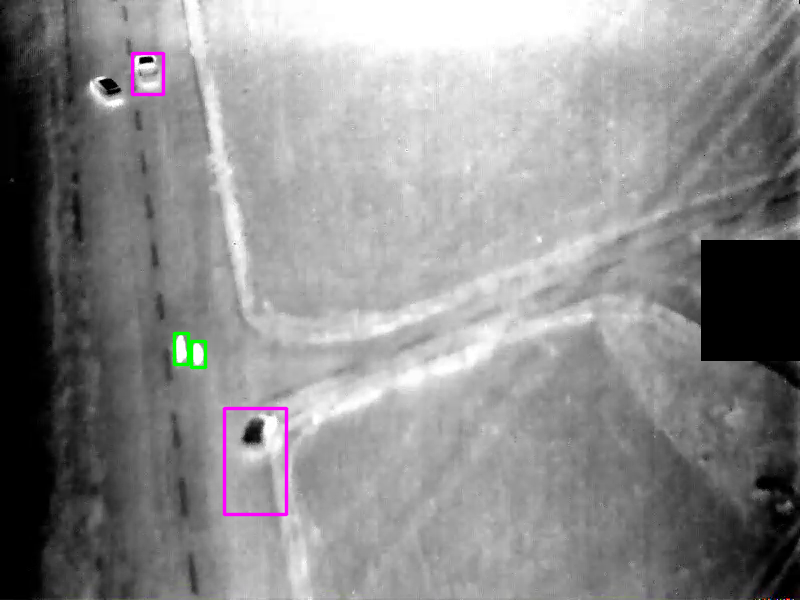}
    \put(2,3){\color{lime}\footnotesize \textbf{YOLOv8\textsuperscript{\textdagger}}}
    \put(85,3){\color{lime}\footnotesize \textbf{0.10}}
    \end{overpic}\\
  \end{tabular}
\end{center}
\vspace{-.6cm}
\caption{Qualitative results of the different detectors trained on dirt-road and evaluated on runway.
Bottom right value in each image represents the conference threshold adopted at inference time.
Bounding boxes: green for \texttt{person}, magenta for \texttt{vehicle}.
Recording altitude: 82m.}
\label{fig:qualitative_results_d2r}
\end{figure*}

Fig.~\ref{fig:qualitative_results_d2r} shows the qualitative results of the different detectors when their models are trained on dirt-road and evaluated on runway.
We can observe that Deformable DETR is the best performing one, followed by VarifocalNet.
The most noisy one resulted to be CornerNet.

\vfill

%-----------------------------------
%-----------------------------------

%-----------------------------------
%-----------------------------------

%-----------------------------------
%-----------------------------------

%-----------------------------------
%-----------------------------------

%-----------------------------------
%-----------------------------------

%-----------------------------------
%-----------------------------------

% %%%%%%%%% REFERENCES
% \onecolumn
% \clearpage
% \twocolumn

% {\small
% \bibliographystyle{ieee_fullname}
% \bibliography{refs}
% }

% ********************************
% ********************************
% ********************************
% ********************************
% ********************************
% ********************************
% ********************************
% ********************************

% \end{document}

\end{document}